\documentclass[lettersize,journal]{IEEEtran}
\usepackage{amsmath,amsfonts}
\usepackage[table]{xcolor}

\usepackage{algorithmic}
\usepackage{array}
\usepackage[caption=false,font=normalsize]{subfig}
\usepackage{textcomp}
\usepackage{stfloats}
\usepackage{url}
\usepackage{verbatim}
\usepackage{graphicx}
\usepackage{cite}
\usepackage{tabularx}
\usepackage{nicematrix}
\usepackage{empheq}    
\usepackage{booktabs,subcaption,amsfonts,dcolumn}
\usepackage{float}
\usepackage[export]{adjustbox}
\usepackage{dcolumn}

\usepackage{titlesec}
\usepackage{placeins}
\usepackage{xcolor}
\usepackage{times}  
\usepackage{helvet}  
\usepackage{courier}  
\usepackage{pifont}
\usepackage{caption} 
\usepackage[linesnumbered,ruled,vlined]{algorithm2e}

\usepackage{booktabs}
\usepackage{multirow}
\usepackage{enumitem}
\usepackage{amssymb}

\usepackage[bb=dsserif]{mathalpha}
\usepackage{newfloat}
\usepackage{makecell}
\usepackage{listings}

\usepackage{xcolor}

\definecolor{cvprblue}{rgb}{0.21, 0.49, 0.74}

\usepackage[colorlinks=true,
            linkcolor=black,
            citecolor=black,
            urlcolor=cvprblue]{hyperref}
\DeclareCaptionStyle{ruled}{labelfont=normalfont,labelsep=colon,strut=off} 
\begin{document}

\title{Exploring Dynamic Properties of Backdoor Training Through \\ Information Bottleneck}

\author{Xinyu Liu,~\IEEEmembership{Student Member,~IEEE,} Xu Zhang,~\IEEEmembership{Student Member,~IEEE,} Can Chen,~\IEEEmembership{Member,~IEEE,} Ren Wang$^\star$,~\IEEEmembership{Senior Member,~IEEE,} 
\thanks{Xinyu Liu is with Michigan State University, East Lansing, MI 48824 USA (email: liuxin73@msu.edu).
Xu Zhang is with Illinois Institute of Technology, Chicago, IL 60616 USA (email: xzhang156@hawk.iit.edu).
Can Chen is with The University of North Carolina at Chapel Hill, Chapel Hill, NC 27514 USA (email: canc@unc.edu).
Ren Wang is with Illinois Institute of Technology, Chicago, IL 60616 USA (email: rwang74@illinoistech.edu).}
\thanks{$^\star$Corresponding Authors}
\thanks{This work was supported in part by the National Science Foundation under grants IIS-2246157 and FMitF-2319243, and by the Department of Energy under grant DE-CR0000042.}}



\maketitle

\begin{abstract}
Understanding how backdoor data influences neural network training dynamics remains a complex and underexplored challenge. In this paper, we present a rigorous analysis of the impact of backdoor data on the learning process, with a particular focus on the distinct behaviors between the target class and other clean classes. Leveraging the Information Bottleneck (IB) principle connected with clustering of internal representation, We find that backdoor attacks create unique mutual information (MI) signatures, which evolve across training phases and differ based on the attack mechanism. Our analysis uncovers a surprising trade-off: visually conspicuous attacks like BadNets can achieve high stealthiness from an information-theoretic perspective, integrating more seamlessly into the model than many visually imperceptible attacks. Building on these insights, we propose a novel, dynamics-based stealthiness metric that quantifies an attack's integration at the model level. We validate our findings and the proposed metric across multiple datasets and diverse attack types, offering a new dimension for understanding and evaluating backdoor threats. Our code is available in \url{https://github.com/XinyuLiu71/Information_Bottleneck_Backdoor.git}.

\end{abstract}


\section{Introduction}
\label{sec:intro}
\IEEEPARstart{T}{th} rise of deep neural networks (DNNs) has revolutionized artificial intelligence, driving breakthroughs in computer vision~\cite{krizhevsky2012imagenet}, natural language processing~\cite{torfi2020natural}, and autonomous systems~\cite{wen2022deep}. However, the effectiveness and trustworthiness of these networks depend heavily on the integrity of their training data. Compromised data, particularly through backdoor attacks, can covertly embed malicious behaviors into models, jeopardizing their reliability and security~\cite{goldblum2022dataset, li2022backdoor}.

Backdoor attacks inject stealthy triggers into the training data, enabling adversaries to manipulate a model's predictions at will by presenting the corresponding trigger during inference. While appearing normal on benign inputs, these compromised models are poised for exploitation, posing a significant threat to their operational integrity~\cite{gu2017badnets, turner2019label, nguyen2021wanet, zhao2020clean, bai2023physics}. Defending against such attacks has spurred extensive research, leading to strategies such as backdoor trigger recovery, model detection, and network reconstruction~\cite{borgnia2021strong, huang2022backdoor, pal2023towards, wang2020practical}. However, most of these defenses focus on post-compromise detection, with limited attention to understanding the training dynamics of networks affected by backdoor data. This gap underscores the need for deeper insights into the fundamental mechanisms through which backdoor data alters learning processes.

A few recent efforts delve into training dynamics using coarse metrics such as loss functions or gradients~\cite{li2021anti,mo2024robust}.  However, they can only provide limited insight into the amount of information the network encodes about certain features or how features related to backdoor triggers evolve over time. In contrast, analyzing training from a Mutual Information (MI) perspective offers a richer, more granular view of how data representations evolve during training.
To this end, we leverage the Information Bottleneck (IB) principle, a framework originally devised to study the trade-off between compression and task-relevant feature preservation in learning representations~\cite{tishby2000information, tishby2015deep}. The IB method has proven invaluable in understanding neural networks' learning dynamics, including representation acquisition~\cite{amjad2019learning, federici2020learning}, architectural effects~\cite{simon2022towards, namekawa2021evolutionary}, and robustness properties~\cite{wang2021revisiting}.



Our work presents a novel IB-based framework to analyze how different backdoor patterns are learned and retained during neural network training. We reveal that backdoor samples exhibit unique MI trajectories, which existing metrics fail to capture. By examining six representative backdoor attacks, we identify critical properties of backdoor training, such as different MI dynamics from other clean classes at specific phases and different MI dynamics between different attacks, providing new insights into how these attacks disrupt learning processes. Visualizations and clustering analyses further demonstrate that backdoor triggers form distinct or augmented sub-clusters within the learned representation space. Building on these findings, we propose an IB-based stealth metric that leverages the distinctive MI dynamics of backdoor data. Notably, we observe the counterintuitive yet significant result that BadNets, often considered a simple and conspicuous attack, consistently achieves high stealthiness among the studied attacks through IB perspective.




\section{Related Work}
\label{sec: related_work}




\paragraph{Backdoor Attacks}
Backdoor attacks embed hidden malicious behaviors into models, activated by a specific trigger at inference time~\cite{li2022backdoor, goldblum2022dataset}. The design of these attacks has evolved significantly, from early methods like that of~\cite{gu2017badnets}, which used conspicuous patterns, to highly sophisticated, imperceptible triggers~\cite{liu2018trojaning}. To improve stealth, subsequent strategies diversified:~\cite{chen2017targeted} proposed blending triggers globally with the image, while others explored subtle spatial warping~\cite{nguyen2021wanet} or more adaptive blending techniques~\cite{qi2023revisiting}. A major leap in stealth was the development of clean-label attacks, which introduce small, adversarially crafted perturbations without altering the original label~\cite{turner2019label}, a concept later extended to more adaptive attack like~\cite{qi2023revisiting} which blended triggers into target class inputs and regularized variants to disrupt latent clustering without label flipping. Even greater imperceptibility has been achieved by operating in alternate domains, such as injecting triggers into the frequency space~\cite{wang2022invisible} or drew inspiration from the physical world like~\cite{liu2020reflection} using mathematically modeled natural reflections as a highly stealthy and robust clean-label backdoor trigger. This continuous evolution towards more covert and diverse attack strategies necessitates a deeper analysis of their impact on training dynamics, which is the focus of our work.



\paragraph{Backdoor Defenses.} 
In defense against these attacks, researchers have proposed numerous strategies. These include backdoor trigger recovery~\cite{wang2019neural, guo2019tabor,liu2022complex,xiang2022post,hu2021trigger}, which attempts to reverse-engineer the adversary's trigger, and backdoor model reconstruction~\cite{borgnia2021strong,huang2022backdoor,pal2023towards}, aiming to cleanse the backdoor model of its malicious behaviors. Another line of works is model (data) detection methods~\cite{kolouri2020universal,wang2020practical,shen2021backdoor,xu2021detecting,palBackdoorSecretsUnveiled2024,downer2024securing}, which seek to determine whether backdoor triggers have compromised a model (data sample). 
While these methods have shown effectiveness, they often overlook the dynamic characteristics of model training, which are crucial to understanding backdoor behavior. Some recent works have begun exploring training dynamics: Li et al.~\cite{li2021anti} analyze cross-entropy loss trajectories and observe that backdoor samples are learned faster, though mainly in early training and without considering different attack types. Mo et al.~\cite{mo2024robust} examine topological changes in latent space but do not model temporal evolution. In contrast, our work provides a phase-aware, class-specific analysis of training dynamics, uncovering how various attacks differentially affect information flow across classes over time.


\paragraph{Information Bottleneck in Deep Learning.} 
The IB principle provides a theoretical framework for understanding the learning process in deep neural networks by balancing the trade-off between compressing input information and preserving predictive information about the output~\cite{tishby2000information, shwartz2017opening}. This principle has been used to study learning dynamics~\cite{saxe2019information}, network depth~\cite{tishby2015deep}, internal representations~\cite{shwartz2017opening}, and generalization using IB-inspired objectives~\cite{vera2018role}. While IB has offered insights into standard training processes~\cite{goldfeld2018estimating}, its application to backdoor learning remains limited. Existing works do not examine how triggers affect information flow or how backdoor signals persist across layers. Our work bridges this gap by using IB analysis to reveal how different backdoor attacks alter class-wise information dynamics throughout training, providing a foundation for phase-aware backdoor detection and mitigation strategies.


\section{Preliminaries}
\label{sec: preliminaries}
Driven by the need to explore how neural networks train on both clean and backdoor-affected data, this section introduces IB formulations, presents the basic backdoor attack framework, and outlines the threat model for our study.

\paragraph{Information Bottleneck Principle.} 
MI between two random discrete variables $X$ and $Y$ quantifies their statistical dependence and is defined as:
\begin{equation}
    I(X;Y) = \sum _{x,y} p(x,y)\text{log}\frac{p(x,y)}{p(x)p(y)}
\end{equation}
where $p(x,y)$ is the joint probability mass function, $p(x)$ and $p(y)$ are the marginal probability distributions of $X$ and $Y$, respectively. In the continuous case, the summation is replaced by an integral and the probability mass functions by probability density functions. 

In a DNN with \( N \) layers, the output of any hidden layer can be modeled as a random variable \( T \), with the input and labels denoted as \( X \) and \( Y \), respectively. \( T \) represents the intermediate representation that extracts relevant information from \( X \) necessary for predicting \( Y \), while discarding irrelevant or redundant features.
We focus on two MI terms: \( I(X;T) \), measuring the compression of input into representation, and \( I(T;Y) \), capturing the predictive power of \( T \) with respect to the labels.
The core idea of IB is to compress the input \( X \) into a representation \( T \) that retains only the information necessary for predicting \( Y \). Mathematically, the IB objective is expressed as: $
\min_{T} I(X;T) - \beta I(T;Y)
$, where the parameter \( \beta \) balances the trade-off between compression and prediction. In deep learning, the IB principle is inherently applied as each layer progressively compresses \( X \) into a more task-relevant representation. Tracking the dynamics of \( I(X;T) \) and \( I(T;Y) \) across training provides insights into how representations evolve over time. We hypothesize that, from an IB perspective the MI dynamics of the target class (containing backdoor samples) differ significantly from those of other clean classes due to the presence of backdoor triggers. By analyzing and comparing the training dynamics of these MI terms across clean and backdoor classes, we aim to uncover the distinct training patterns introduced by backdoor attacks and evaluate their impact on information flow throughout the learning process.

\paragraph{Backdoor Attacks.} 
Given a classification task with training data \(\mathcal{D} = \{(\textbf{x}_i, y_i)\}_{i=1}^N\), where \(\textbf{x}_i \in \mathcal{X} = [0,1]^{C \times W \times H}\) represents input samples (e.g., images) and \(y_i \in \mathcal{Y} = \{0,1,\ldots,K-1\}\) represents the class labels, an attacker selects a poison ratio \(\alpha = \frac{|\mathcal{D}_s|}{|\mathcal{D}|}\), where \(\mathcal{D}_s\) is the subset of samples to be poisoned, and typically \(\alpha \ll 1\). The attacker modifies the training dataset \(\mathcal{D}\) to generate a poisoned dataset \(\mathcal{D}_m = \mathcal{D}_b \cup \mathcal{D}_c\), where \(\mathcal{D}_b = \{(\textbf{x}', y_t) \mid \textbf{x}' = g(\textbf{x}), (\textbf{x}, y) \in \mathcal{D}_s \}\) contains poisoned samples with a trigger applied through a backdoor generation function \(g(\cdot)\), and \(\mathcal{D}_c = \mathcal{D} \setminus \mathcal{D}_s\) contains the remaining clean data. Here, \(y_t\) represents the target label to which poisoned samples are classified during inference. The attacker's objective is to train a backdoor model \(\mathcal{F}_\theta: \mathcal{X} \rightarrow \mathbb{R}^K\) that minimizes the loss \(\mathcal{L}(\mathcal{F}_\theta, \mathcal{Y})\), such that samples containing the trigger are classified as the target class \(y_t\), while clean samples remain correctly classified during inference.

\begin{figure}[h] 
    \centering
    \subfloat[]{\includegraphics[trim=100 35 90 35,clip,width=0.15\linewidth]{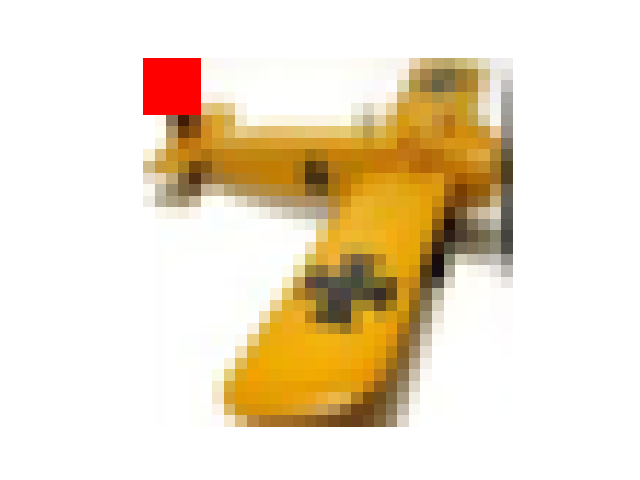}}
    \hfil
    \subfloat[]{\includegraphics[trim=100 35 90 35,clip,width=0.15\linewidth]{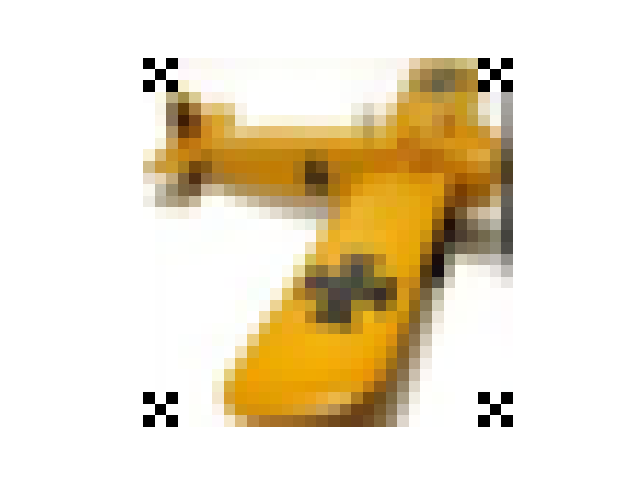}}
    \hfil
    \subfloat[]{\includegraphics[trim=100 35 90 35,clip,width=0.15\linewidth]{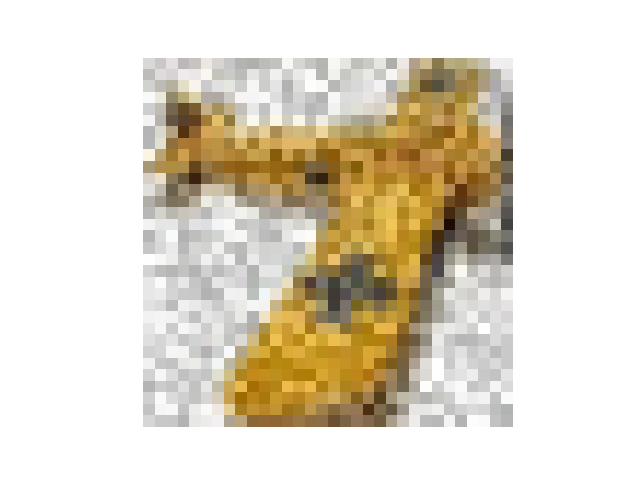}}
    \hfil
    \subfloat[]{\includegraphics[trim=100 35 90 35,clip,width=0.15\linewidth]{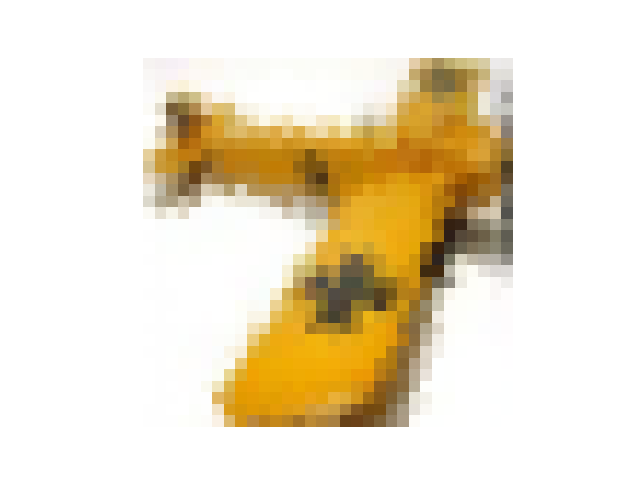}}
    \hfil
    \subfloat[]{\includegraphics[trim=0 0 0 0,clip,width=0.15\linewidth]{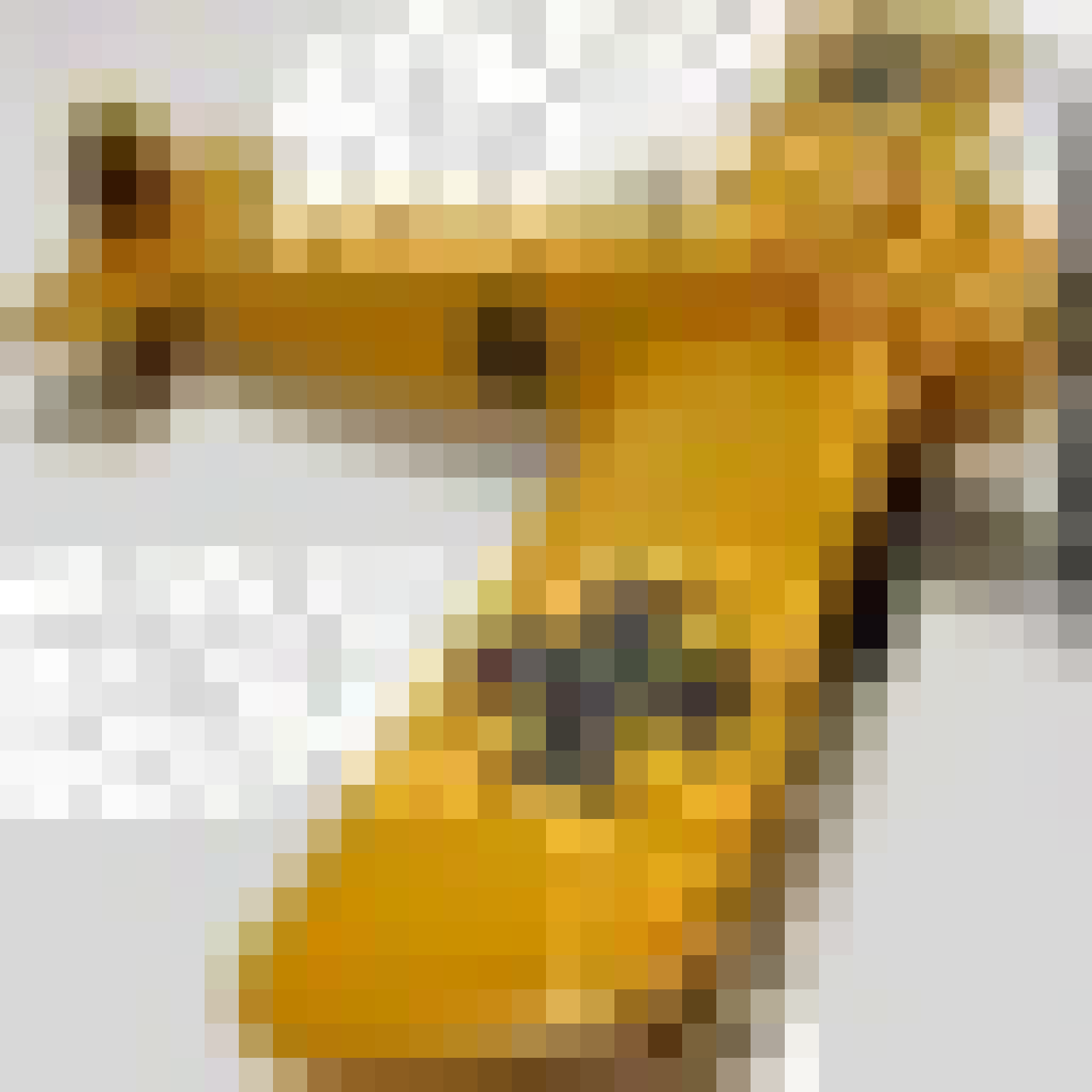}}
    \hfil
    \subfloat[]{\includegraphics[trim=0 0 0 0,clip,width=0.15\linewidth]{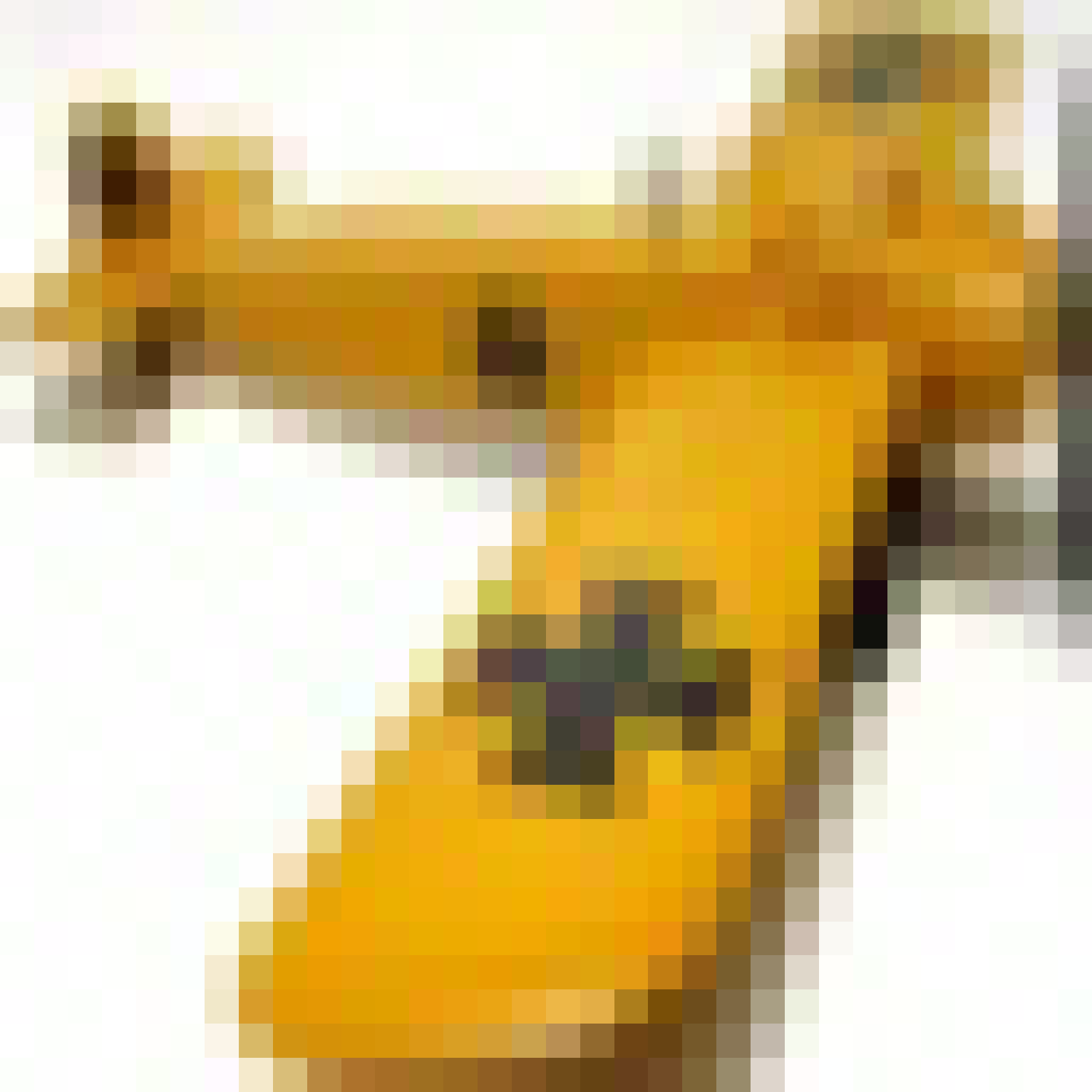}}
    
    \caption{\footnotesize{Poisoned samples generated by different attacks: (a) BadNets, (b) LC, (c) Blend, (d) WaNet, (e) Adap-Blend, and (f) Ftrojan.}}
    \label{fig:attacks}
\end{figure}

\paragraph{Our Threat Model.}
\ding{202} \textit{Attack Setting}. 
We adopt an all-to-one backdoor attack setting where all poisoned samples are mapped to a single target label ($y_{t}=0$). Specifically, for dirty-label attacks such as BadNets, 10\% of samples from each class are selected and modified by injecting a trigger and relabeling them to the target class. For clean-label attacks such as Label-Consistent (LC) attack, however, only 10\% of samples from the target class ($y_{t}=0$) are poisoned by adding imperceptible triggers, maintaining label consistency. 
\ding{203}  \textit{Attacker's Capability and Goal}. The adversary can poison the training data but has no knowledge about the training process. The goal is to cause the model to misbehave during test time when an input contains the trigger, but performs normally on benign ones.
\section{Analytical Framework}
\label{sec: analytical_framework}

This section presents an analytical framework for examining how backdoor attacks influence learning dynamics, as captured by the two MI terms in the IB. We introduce the MI estimation methods, adopt a class-wise perspective to capture per-class information flow, and link MI to the clustering of internal representations, providing a basis for understanding the effects of different backdoor mechanisms.

\subsection{MI Misestimation in Deterministic DNNs}
\label{sec:misestimation}
The estimation of MI in deterministic DNNs presents unique challenges. Unlike probabilistic models, deterministic DNNs can encode information about $X$ in arbitrarily fine variations of $T$ essentially without loss, regardless of the less number of neurons in deeper layers. Consequently, $I(X;T) = H(X) = \log(n)$ and $I(T;Y) = I(X;Y)$ remain constant, determined solely by the number of input images $n$, rather than by network parameters or the image content. Previous work in \cite{shwartz2017opening} and \cite{saxe2019information} employed quantization and noise injection as means for measurements. But these fluctuations could therefore be a consequence of estimation errors rather than changes in MI because they are not the intrinsic part of the network. We can focus on quantization method, which is widely used in MI estimation. The value of $I(X;Bin_m[T])$ depends on the quantization resolution $m$ which determines the distribution of $Bin_m[T]$, while the true $I(X;T)$ value is still $\log n$. To address this issue, \cite{goldfeldEstimatingInformationFlow2019} proposed the use of noisy DNNs. In this approach, $T = \phi(X)$ is replaced with $T^{(\sigma)} = T + Z^{(\sigma)}$, where $Z^{(\sigma)}$ represents Gaussian noise injected into the output of each hidden unit. This noise is intrinsic to the system, meaning the network is trained with the noisy activation values. In our work, we adopt noisy DNNs by incorporating noise into the model architecture to make MI estimations more meaningful.

\subsection{Method Selection for MI Estimation}
\label{subsec:mi_selection}
Accurate estimation of MI is critical for understanding how backdoor attacks disrupt the information flow in neural networks. Given the high-dimensional and continuous nature of neural representations, traditional MI estimation methods, which often rely on explicit assumptions about probability distributions, are computationally prohibitive and impractical. For our framework, we require an MI estimator that is scalable, stable, and capable of capturing subtle information dynamics across classes. To meet these requirements, we evaluated two advanced MI estimators: Mutual Information Neural Estimation (MINE) \cite{belghazi2018mine} and Noise-Contrastive Estimation (InfoNCE) \cite{oordRepresentationLearningContrastive2019}. 

MINE formulates MI estimation as maximizing the Donsker-Varadhan representation:
\begin{equation}
    \hat{I}_{\theta}(X; Y) = \mathbb{E}_{P_{X,Y}}[T_{\theta}(x, y)] - \log \mathbb{E}_{P_X \otimes P_Y}[e^{T_{\theta}(x, y)}]
    \label{eq:mine}
\end{equation}
The first term, \(\mathbb{E}_{P_{X,Y}}[T_{\theta}(x, y)]\), captures the alignment between \(X\) and \(Y\) under their joint distribution, while the second term, \(\log \mathbb{E}_{P_X \otimes P_Y}[e^{T_{\theta}(x, y)}]\), evaluates the network under the assumption that \(X\) and \(Y\) are independent.

InfoNCE relies on contrastive predictive coding:
\begin{align}
    \hat{I}_{\theta}(X; Y) 
    &= \mathbb{E}_{P_{X,Y}} \bigg[ T_{\theta}\left(x, y\right) \notag \\
    &\quad - \mathbb{E}_{P_X \otimes P_Y} \left[ \log \sum_{x^{\prime}} e^{T_{\theta}\left(x^{\prime}, y\right)} \right] \bigg]
    \label{eq:infonce}
\end{align}
The first term, \(\mathbb{E}_{P_{X,Y}}[T_{\theta}(x, y)]\), evaluates the alignment between true pairs of \((x, y)\), while the second term penalizes the network for assigning high similarity to incorrect pairs through a contrastive noise mechanism, \(\sum_{x'} e^{T_{\theta}(x', y)}\). By comparing true pairs to negative pairs sampled from the marginal distribution, InfoNCE efficiently estimates MI in a self-supervised framework.

\begin{figure}[h]
    \centering
    \includegraphics[width=0.8\linewidth, trim=10 10 0 0, clip]{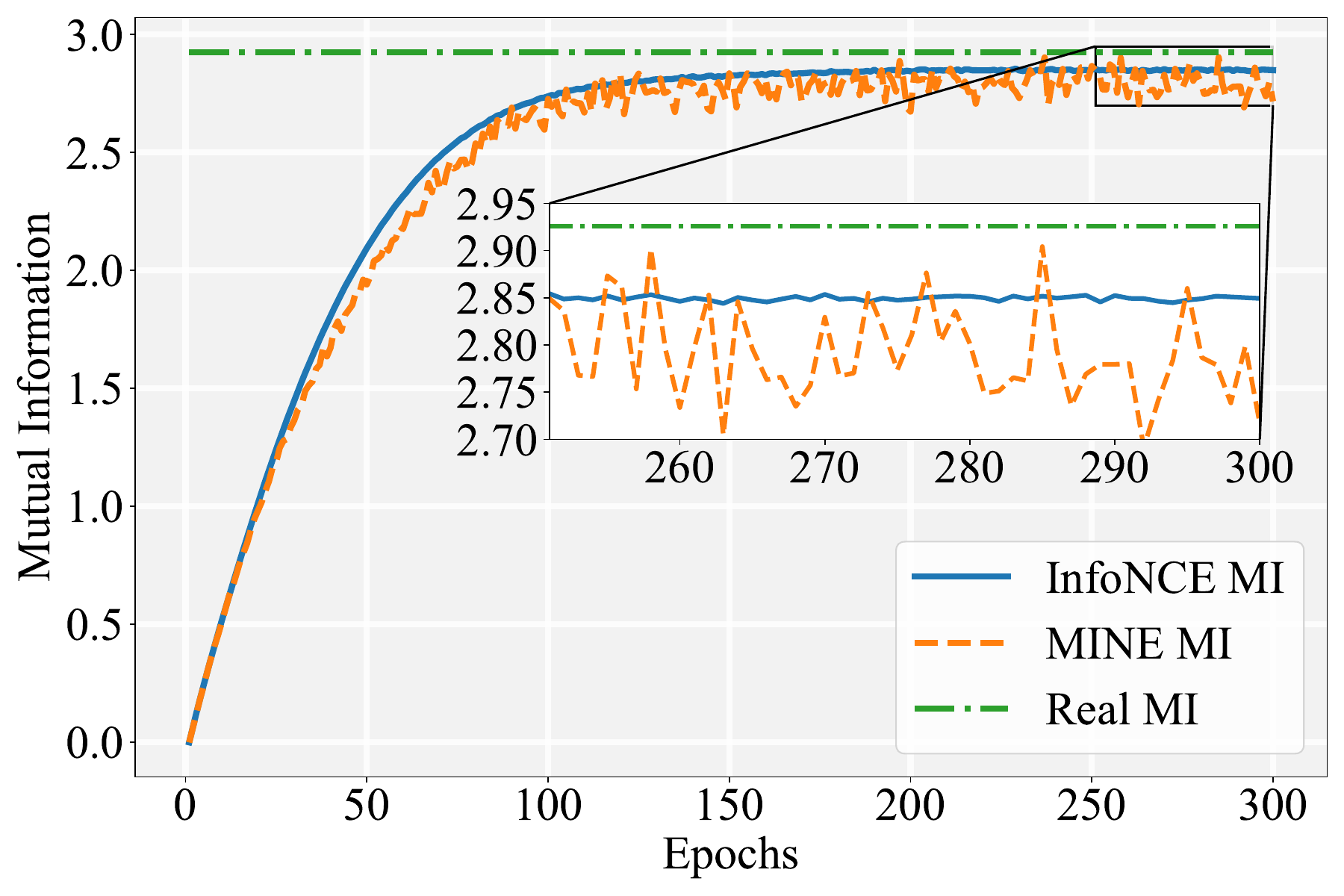}
    \vspace{-2mm}
    \caption{\footnotesize{Comparison of MI estimation between InfoNCE \cite{oordRepresentationLearningContrastive2019} and MINE \cite{belghazi2018mine} under a controlled experimental setup. The dashed green line represents the true MI value. InfoNCE demonstrates smoother convergence, reduced fluctuation, and faster attainment of the true MI value compared to MINE. The inset highlights the stability of MI estimates during the final training epochs, where InfoNCE consistently outperforms MINE in accuracy and variance.}}
    \label{fig:mi_estimation_comparison}
    \vspace{-5mm}
\end{figure}

To validate their suitability, we conducted a comparative study using correlated multivariate Gaussian distributions, which permit a closed-form, analytical ground-truth MI value. Specifically, we synthesized a 5-dimensional $Y \sim \mathcal{N}(0, 3I)$ and a correlated $X = Y_{1:3} + \mathcal{N}(0, 0.5I)$ derived from $Y$'s first three dimensions. The analytical MI between these distributions is computed as:
\begin{equation}
I(X; Y) = \frac{1}{2}\log\frac{\det(\Sigma_X)\det(\Sigma_Y)}{\det(\Sigma_{XY})}
\end{equation}
where $\Sigma_X$, $\Sigma_Y$, and $\Sigma_{XY}$ denote the covariance matrices of $X$, $Y$, and their joint distribution, respectively. To ensure equitable comparison, both estimators used identical fully-connected (input $\to$ 512 $\to$ 512 $\to$ 1) architectures with ReLU activations and an Adam optimizer (LR 2e-4, weight decay 1e-4), trained for 300 epochs. We utilized 256 negative samples for InfoNCE and an exponential moving average (rate 0.01) for MINE, following established practices . As shown in Figure~\ref{fig:mi_estimation_comparison}, InfoNCE achieves superior convergence stability, reduced fluctuations, and more accurate estimation of the ground-truth MI compared to MINE. Based on these empirical findings, we selected InfoNCE as the primary MI estimator for our framework.

\subsection{Class-Wise Mutual Information Estimation}
MI estimation traditionally considers the relationship between a model's input and output across the entire dataset. However, this global perspective may overlook class-specific behaviors that are critical for understanding and mitigating backdoors, particularly in all-to-one scenarios where poisoned data is mapped to a specific target class under supervised learning. To capture these dynamics, we propose a class-wise MI estimation framework, which focuses on the information dynamics within \textbf{each individual class} while training on the entire dataset. 

Following the previous study~\cite{saxe2019information}, We mainly care about the relationship between the model input $X$ and the last hidden layer output $T$, which clearly exhibits two distinct phases in its information dynamics. But if $Y$ is the target output, directly computing $I(T;Y)$ for individual classes results in $I(T;Y)=0$, as $Y$ becomes constant within a class, losing variability. To overcome this, we replace $I(T;Y)$ with $I(T;Y_\text{pred})$, where $Y_\text{pred}$ represents the probability vector of the model's predictions. This enables meaningful class-wise comparisons. A higher $I(T;Y_\text{pred})$ indicates that the representation $T$ contains more information about the model's own output distribution, which serves as a proxy for its predictive certainty. In other words, the model is more "confident" about its predictions for that class.

As discussed in Section~\ref{subsec:mi_selection}, noise injection plays a crucial role in MI estimation between $X$ and $T$, $T$ and $Y_\text{pred}$. Specifically, we inject Gaussian noise with standard deviation 
$\gamma$ after each activation layer and before the final output. This noise serves as a regularizer, akin to dropout, and is treated as a tunable hyperparameter optimized for the DNN's performance. 

\begin{figure}[htbp]
    \centering
    \includegraphics[width=0.85\linewidth]{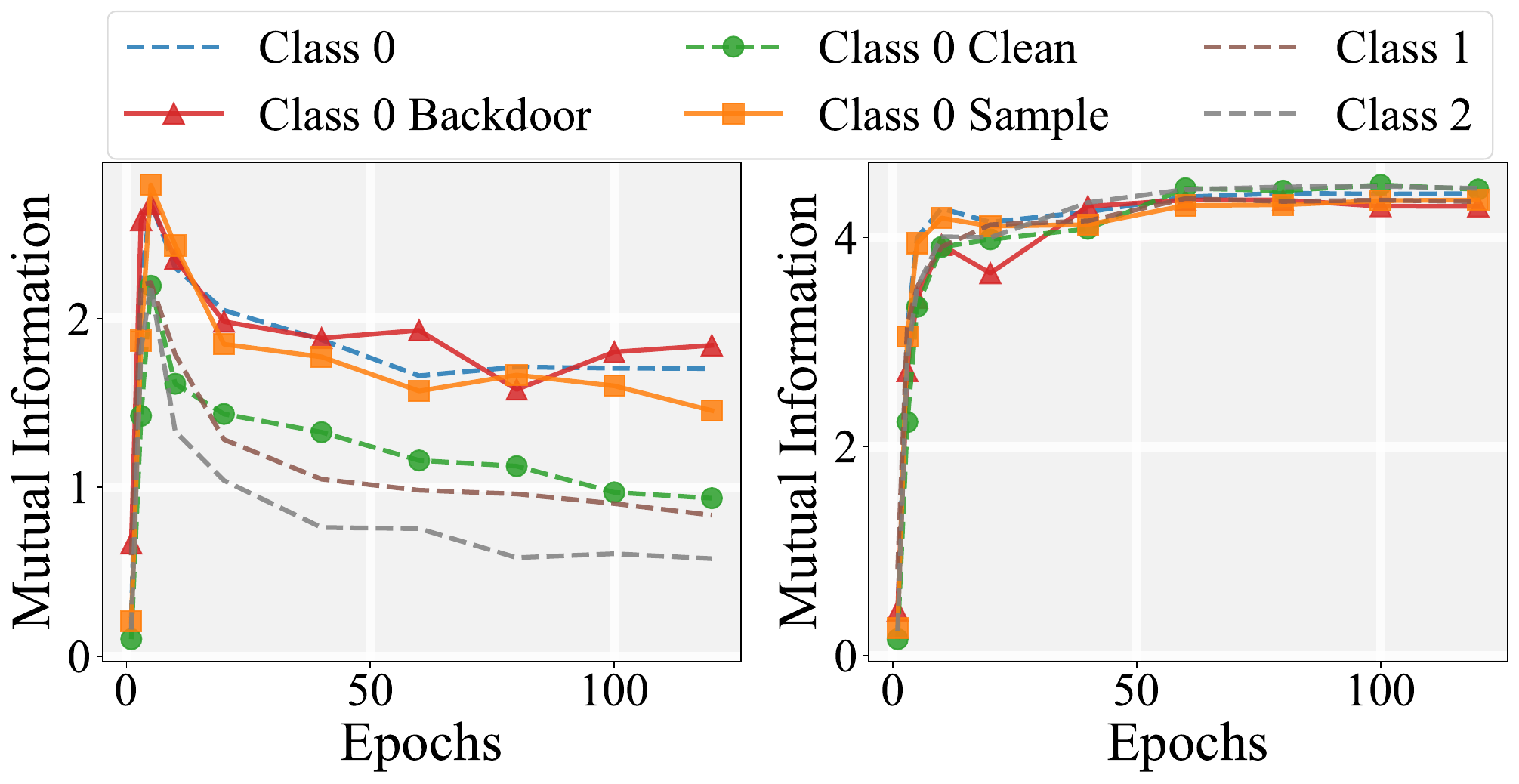}
    \vspace{-2mm}
    \caption{\footnotesize{MI dynamics under BadNets attack on the CIFAR-10 dataset using a ResNet-18 model with a 10\% poisoning ratio and \( \gamma = 0.4 \). The subfigures show \( I(X;T) \) (left) and \( I(T; Y_{\text{pred}}) \) (right) across training phases. \( I(X;T) \) exhibits the distinct two-phase behavior of backdoor samples and higher MI after the compression phased. \( I(T; Y_{\text{pred}}) \) of backdoor samples closely matches that of clean samples. These findings align with the mechanism of BadNets, where the trigger is learned early, and both semantic features from other clean classes and trigger-specific features influence the model representations and predictions.}}
    \label{fig:badnets-0.1-cifar10}
\end{figure}

\label{subsec:impact}
To validate our hypothesis and enable this fine-grained analysis, we must first define the specific data subsets required. The poisoning strategy (detailed in Section~\ref{sec:exp_setup}) creates a class-imbalanced target class (e.g., class 0) containing both poisoned and benign samples. Therefore, we create separate data loaders for: (1) All backdoor data (the isolated poisoned samples, later referred to as \textit{Class 0 Backdoor}); (2) Clean data in class 0 (the original, benign samples from the target class, or \textit{Class 0 Clean}); and (3) Class 0 sample subset data, consisting of a proportional mix of backdoor and clean samples from class 0, matched in size to other clean classes (or \textit{Class 0 Sample}). We also analyze the remaining (4) clean non-target classes (e.g., classes 1 and 2) for comparison. Our pilot exploration (Figure~\ref{fig:badnets-0.1-cifar10}) utilizes this framework on the CIFAR-10 dataset with a ResNet-18 model, under a 10\% BadNets attack targeting class 0.

To structure our analysis, we examine the model's behavior across four key phases of training: \ding{172} Early Learning Phase: The model initially adapts to the data. Backdoor samples, being simple for the model to learn, are rapidly incorporated, causing observable deviations in behavior compared to clean data. \ding{173} Transition Phase: The model shifts from fitting to compression after learning most class-specific features. This transition occurs near the peak of \( I(X;T) \). \ding{174} Compression Phase: The model compresses task-irrelevant features and refines learned representations. The influence of backdoor data becomes more pronounced as it integrates the triggers and \( I(X;T) \) begins to decrease. \ding{175} Final Convergence Phase: The model consolidates its learning.
While the IB principle broadly separates training into fitting and compression, our finer-grained division reveals that the MI dynamics of backdoor data differed significantly from clean data across all training phases.

\label{sec:4.2}
\paragraph{Pilot Exploration.}
As shown in Figure~\ref{fig:badnets-0.1-cifar10}, \( I(X;T) \) under the BadNets attack exhibits a clear two-phase pattern: rapid increase during early learning as the model captures relevant information, followed by a compression phase where redundant features are discarded. The target class (Class 0) consistently exhibits a higher \(I(X;T)\) than clean classes, an effect driven entirely by its backdoor samples. This phenomenon arises from the dual information sources within poisoned samples: the semantic features of the original image and the simple, localized trigger pattern. The model rapidly learns this salient trigger, yet it must also encode the original semantics. This dual-encoding task increases the total information the model processes for these samples, leading to a higher \(I(X;T)\) value that persists even after the compression phase.
In parallel, the MI between the representation and the model's prediction, \(I(T;Y_\text{pred})\), reflects the model's predictive certainty. For backdoor samples, the trigger acts as a powerful "shortcut" to the target label. However, because the original semantic features are also retained, the model does not overfit to the trigger alone and generalizes well to both clean and poisoned data under a 10\% poisoning ratio. Consequently, the predictive certainty for backdoor samples becomes comparable to that of clean samples, resulting in similar \(I(T;Y_\text{pred})\) values at convergence.


\paragraph{Mutual Information and Clustering.} 
Connecting MI with clustering provides complementary insights into how models organize internal representations. While MI quantifies the information encoded in \(T\), clustering reveals how samples are geometrically distributed into classes or subgroups. Consider a minimal example: assume a single noisy neuron with scalar input \(X\). Its output at epoch \(k\) is modeled as \(T(k) = S(k) + Z = \sigma(w_k X + b_k) + Z\), where \(\sigma\) is a monotonic activation function and \(Z \sim \mathcal{N}(0, \gamma^2)\) represents Gaussian noise. Because \(I(X; T(k))\) is invariant under invertible transformations, we simplify it to \(I(S(k); S(k) + Z)\), quantifying the number of distinguishable clusters in \(S(k)\) when corrupted by noise \(Z\). This is analogous to the information capacity of an additive white Gaussian noise (AWGN) channel, where the input constellation is \(\mathcal{S}_k = \{\sigma(w_k x + b_k) \mid x \in \mathcal{X}\}\). 
Prior work~\cite{goldfeldEstimatingInformationFlow2019} has shown that compression in neural networks arises from clustering of internal representations, with clusters predominantly containing samples from the same class. When \(I(X; T)\) is computed within a single class, clustering reflects how the model organizes subgroups within the class to represent intra-class variations under noise constraints.
This insight is particularly useful in backdoor analysis, as clustering allows us to observe whether backdoor samples integrate into the target class, form distinct sub-clusters, or disrupt class boundaries.

\begin{figure*}[htbp]
  \centering
  \setlength{\tabcolsep}{0pt}
  \begin{tabular}{ccc}
  
    \multicolumn{3}{c}{} \\
    
    \includegraphics[width=0.27\textwidth]{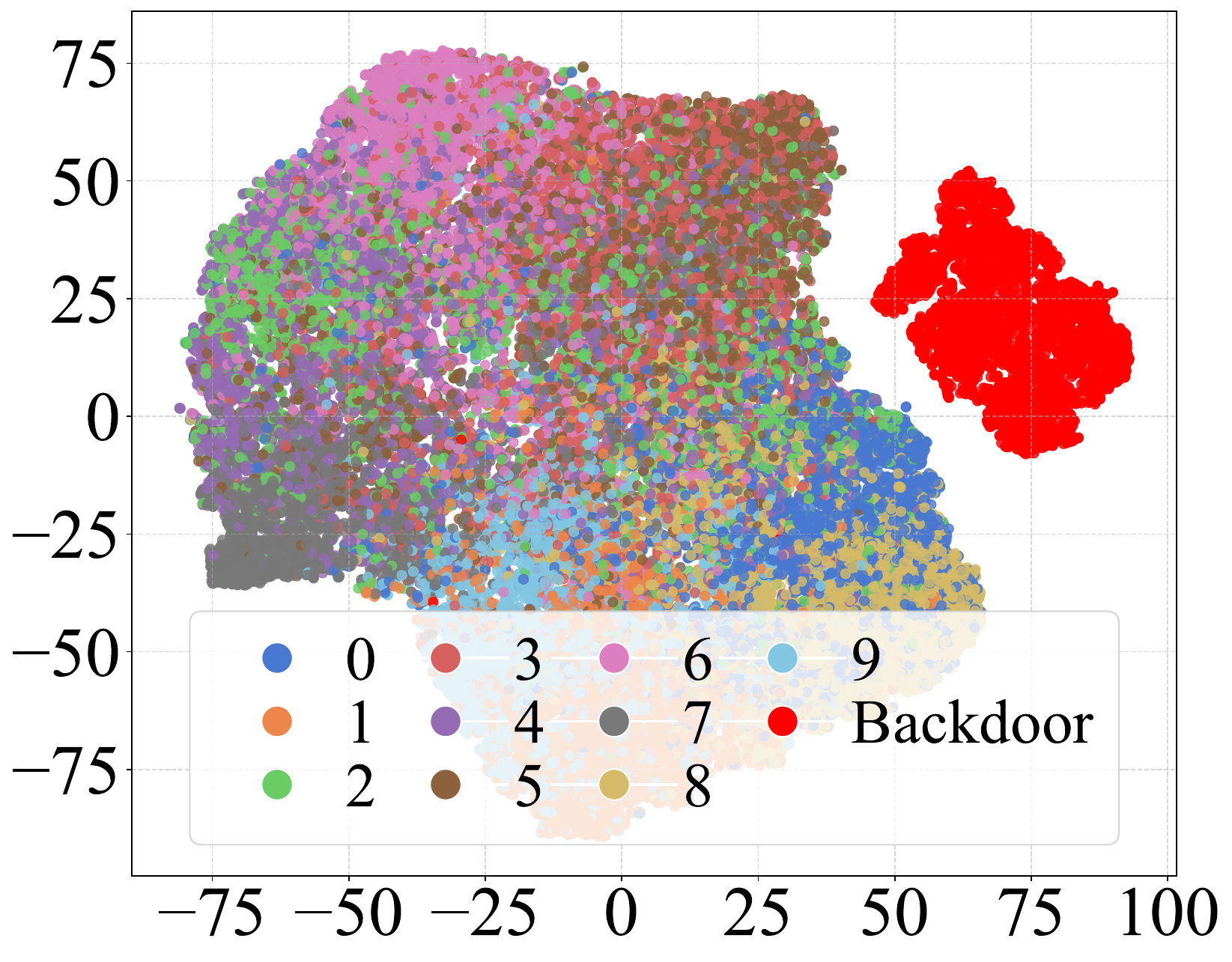} &
    \includegraphics[width=0.27\textwidth]{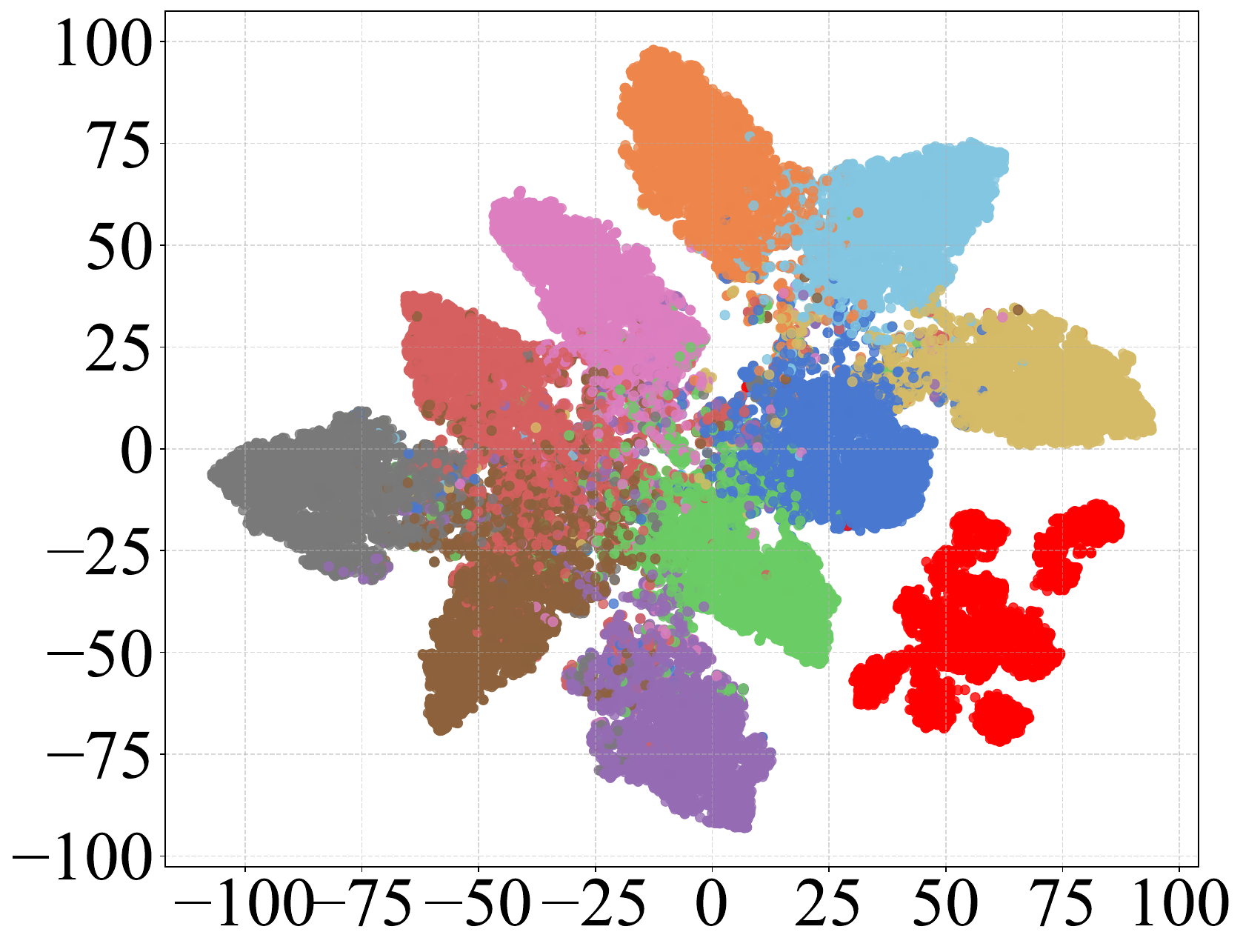} &
    \includegraphics[width=0.27\textwidth]{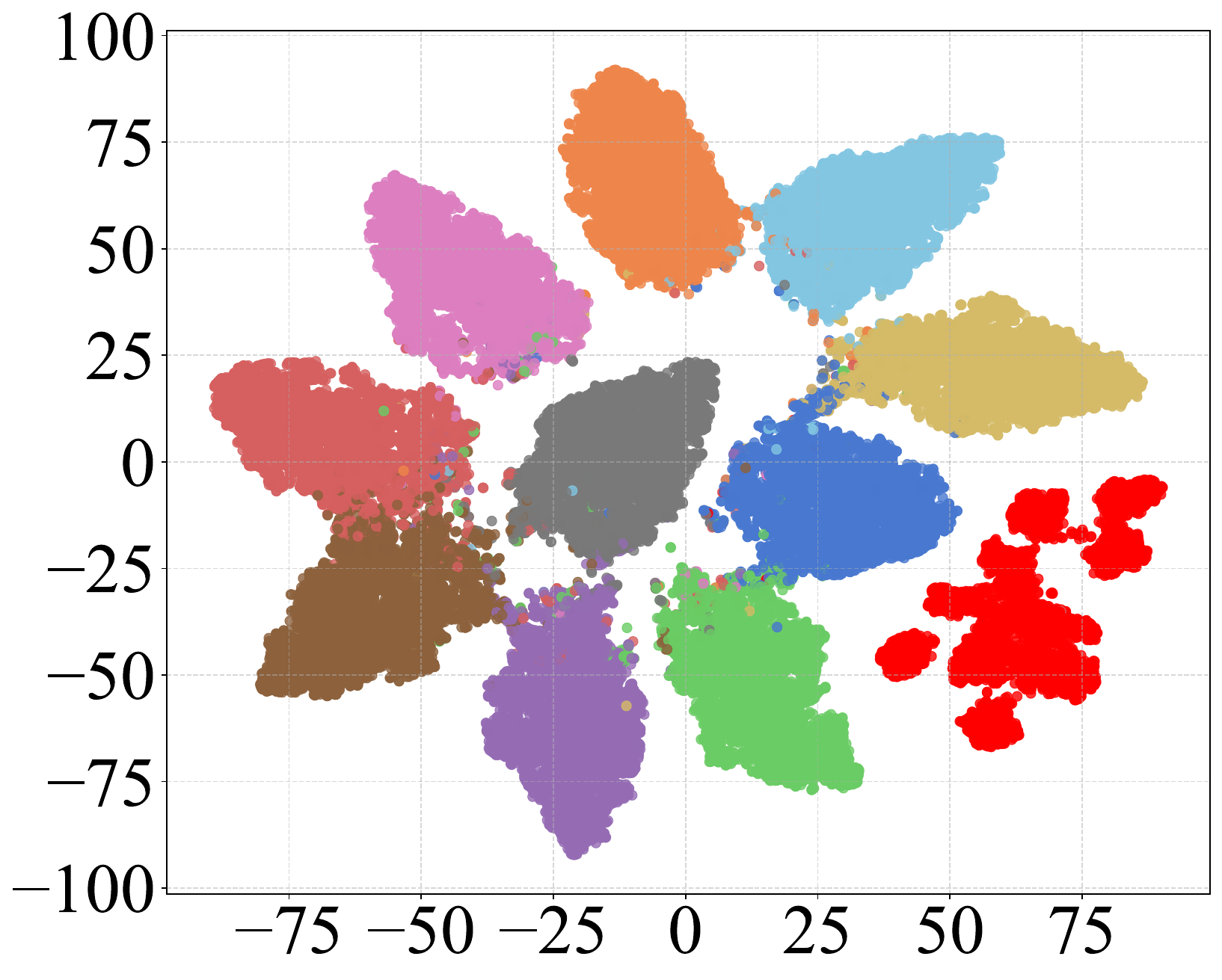} \\
    (a) Early learning phase & (b) Compression phase & (c) Convergence phase
    
  \end{tabular}
  \caption{\footnotesize{t-SNE of the last hidden layer representations $T$ under BadNets attack (CIFAR-10, 10\% poisoning ratio). (a) backdoor samples rapidly form a distinct cluster due to the simple trigger. (b)\&(c) clean samples form well-separated clusters, while backdoor samples remain a distinct cluster and further subdivide into sub-clusters reflecting the model's dual representation of the trigger and semantic features.}}
    \label{fig:tsne_badnets_0.1}
\end{figure*}

This insight is particularly useful for backdoor analysis, as it allows us to visually inspect whether backdoor samples integrate into the target class, form distinct sub-clusters, or disrupt class boundaries. Our t-SNE visualizations in Figure~\ref{fig:tsne_badnets_0.1} strongly validate this connection. In the early learning phase, the sharp rise in \(I(X;T)\) for backdoor data corresponds directly to the rapid formation of a distinct backdoor cluster, driven by the salient trigger. As training progresses, this cluster not only separates from clean data but also subdivides into multiple sub-clusters. This geometric separation and subsequent differentiation perfectly illustrate our "dual information" hypothesis: the model first learns the trigger to form a primary cluster and then learns the diverse semantic features within the backdoor set to form sub-clusters.

\renewcommand{\floatpagefraction}{.9}

\subsection{Evaluation of Stealthiness}

To quantify the stealthiness of a backdoor attack from an information-theoretic perspective, we propose a stealth score. This score measures the average deviation in MI dynamics between the backdoor-injected target class and the clean classes across key training phases. A lower score signifies a stealthier attack, as the MI dynamics of the poisoned class remain closer to those of clean classes, making the attack harder to detect through monitoring information flow.

We first identify a set of representative epochs $\mathcal{E}$, corresponding to the key training phases discussed in Section~\ref{subsec:impact}: early learning, transition, compression, and convergence.
At each selected epoch $e \in \mathcal{E}$, we compute the absolute difference between the MI of the target class (class 0) and the average MI of all other clean classes. This difference is calculated separately for $I(X;T)$ and $I(T;Y_{pred})$, and for three subsets of class 0: clean samples only, backdoor samples only, and all samples.
Let $MI_{I(X;T)}^{(e)}(c, k)$ denote the MI value for epoch $e$, MI type $I(X;T)$, class $c$, and subset $k$ in $\mathcal{K} = \{\text{clean, backdoor, total}\}$. The average MI for clean classes $c \in \{1, ..., N-1\}$ is $AvgMI_{I(X;T)}^{(e)} = \frac{1}{N-1}\sum_{i=1}^{N-1} MI_{I(X;T)}^{(e)}(i, \text{clean})$. The difference for subset $k$ is:
\begin{equation}
    \Delta_{I(X;T)}^{(e, k)} = \left| MI_{I(X;T)}^{(e)}(0, k) - AvgMI_{I(X;T)}^{(e)} \right|
\end{equation}
We compute \(\Delta_{I(T;Y_{pred})}^{(e, k)}\) analogously after min-max normalizing all MI values for fair comparison. The final stealth score holistically aggregates these deviations into a single metric:
\begin{equation}
Score = \frac{1}{2|\mathcal{E}| \cdot |\mathcal{K}|} \sum_{e \in \mathcal{E}} \sum_{k \in \mathcal{K}} \left( \Delta_{I(X;T)}^{(e, k)} + \Delta_{I(T;Y_{pred})}^{(e, k)} \right)
\end{equation}
This metric provides a comprehensive evaluation of how seamlessly a backdoor is integrated into the model's learned representations.

\section{Experiment Result}

\subsection{Experimental Setup}
\label{sec:exp_setup}

\subsubsection{Datasets}
\label{subsubsec:datasets}
For our experiments, we used both CIFAR-10~\cite{krizhevsky2009learning} and SVHN~\cite{netzer2011reading} datasets. Due to the class imbalance in SVHN, we selected the first 4,000 samples from each class to create a balanced training dataset (40,000 samples total). This preprocessing step ensured that our observations of MI dynamics were not biased by class representation disparities.

\subsubsection{Model Architectures}
\label{subsubsec:models}
We employed two widely-used convolutional neural network architectures ResNet-18~\cite{he2016deep} and VGG16~\cite{simonyan2014very}, which were adapted to incorporate noise injection for MI analysis.
We modified the standard ResNet-18 architecture to achieve better performance when training with injected noise. We replaced the first convolutional layer with a smaller kernel variant (kernel size 3$\times$3, stride 1, padding 1 instead of 7$\times$7, stride 2, padding 3) to better accommodate the smaller input dimensions of datasets. We also removed the initial max pooling layer to preserve more spatial information for these smaller-scale image datasets. The final fully connected layer dimensions were adjusted to output 10 classes (512 $\rightarrow$ 10). Most importantly, we integrated Gaussian noise injection after activations in each residual block (with standard deviation $\gamma = 0.4$) to enable meaningful MI estimation. These modifications were necessary to maintain reasonable test accuracy ($>90\%$) despite the presence of injected noise, which typically degrades model performance.
We implemented the VGG16 architecture with batch normalization, maintaining the original structure while adapting the input and output dimensions for our classification tasks. Similar to ResNet-18, we incorporated noise layers with standard deviation $\gamma = 0.4$ to enable MI estimation. The architecture was otherwise kept consistent with standard implementations, with appropriate adjustments for the dataset dimensions.

\subsubsection{Attack Configurations}
\label{subsubsec:attacks}
To ensure a comprehensive evaluation, we test our framework against six representative backdoor attacks. These were deliberately chosen to span the diverse categories of trigger mechanisms surveyed in our Related Work, from simple additive patterns to imperceptible, domain-specific attacks. The selected attacks include BadNets~\cite{gu2017badnets}, a classic visible patch-based attack; Blend~\cite{chen2017targeted} and Adap-Blend~\cite{qi2023revisiting} to represent global blending strategies; WaNet~\cite{nguyen2021wanet} for spatial warping; the highly stealthy clean-label attack, LC~\cite{turner2019label}; and FTrojan~\cite{wang2022invisible}, which operates in the frequency domain. Poisoning ratios are set to 10\% and 1\% (see Appendix~\ref{sec:low_poison_ratio}) for dirty-label attacks (e.g., BadNets) and 25\% and 10\% (of the target class) for clean-label attacks (e.g., LC), ensuring comparable attack success rates (ASR). For LC attacks, we adopt the adversarial perturbation framework with \(\ell_2\)-norm \(\epsilon = 300\) and a four-corner trigger~\cite{turner2019label}.

\subsubsection{Training and Estimation Details}
\label{subsubsec:training_details}
Our training process utilized Stochastic Gradient Descent with momentum 0.9 and weight decay 5e-4, starting with an initial learning rate of 0.01 managed by a ReduceLROnPlateau scheduler that reduced the rate by a factor of 0.5 after 5 epochs without improvement. We trained with a batch size of 256 samples using cross-entropy loss over 120 epochs, without applying any data augmentation techniques.

For MI estimation, we employed the InfoNCE estimator implemented as the same network in Section~\ref{subsec:mi_selection} with hidden dimension 128. Gaussian noise with standard deviation $\gamma = 0.4$ was injected into both the feature extraction layer for $I(X;T)$ estimation and the output layer for $I(T;Y_\text{pred})$ estimation. The estimation process used an AdamW optimizer with initial learning rates of 3e-4 for $I(X;T)$ and 5e-4 for $I(T;Y_\text{pred})$, with weight decay set to 1e-4. We used batch sizes of 128 for $I(X;T)$ and 512 for $I(T;Y_\text{pred})$ estimations, with 512 and 256 negative samples respectively. To ensure efficient computation, we implemented dynamic early stopping with a delta threshold of 1e-2, while setting maximum epochs to 350 for $I(X;T)$ and 250 for $I(T;Y_\text{pred})$.

To capture the dynamic properties of backdoor training, we computed MI at strategic epochs throughout the training process. This sampling strategy allowed us to observe the evolution of information flow during the early learning phase, transition phase, compression phase, and final convergence phase.

\subsection{Validation of Estimation Framework}
\label{subsec:validation}
To ensure the fidelity of our analysis, we conducted a rigorous validation of our InfoNCE-based MI estimation framework. This section confirms the estimator's reliability against known ground truth, its consistency with theoretical bounds, and its adherence to fundamental information-theoretic principles.

\subsubsection{Estimator Accuracy and Bound-Checking}
\begin{figure*}[t]
    \centering
    \begin{minipage}[t]{0.32\linewidth}
        \centering
        \includegraphics[width=\linewidth]{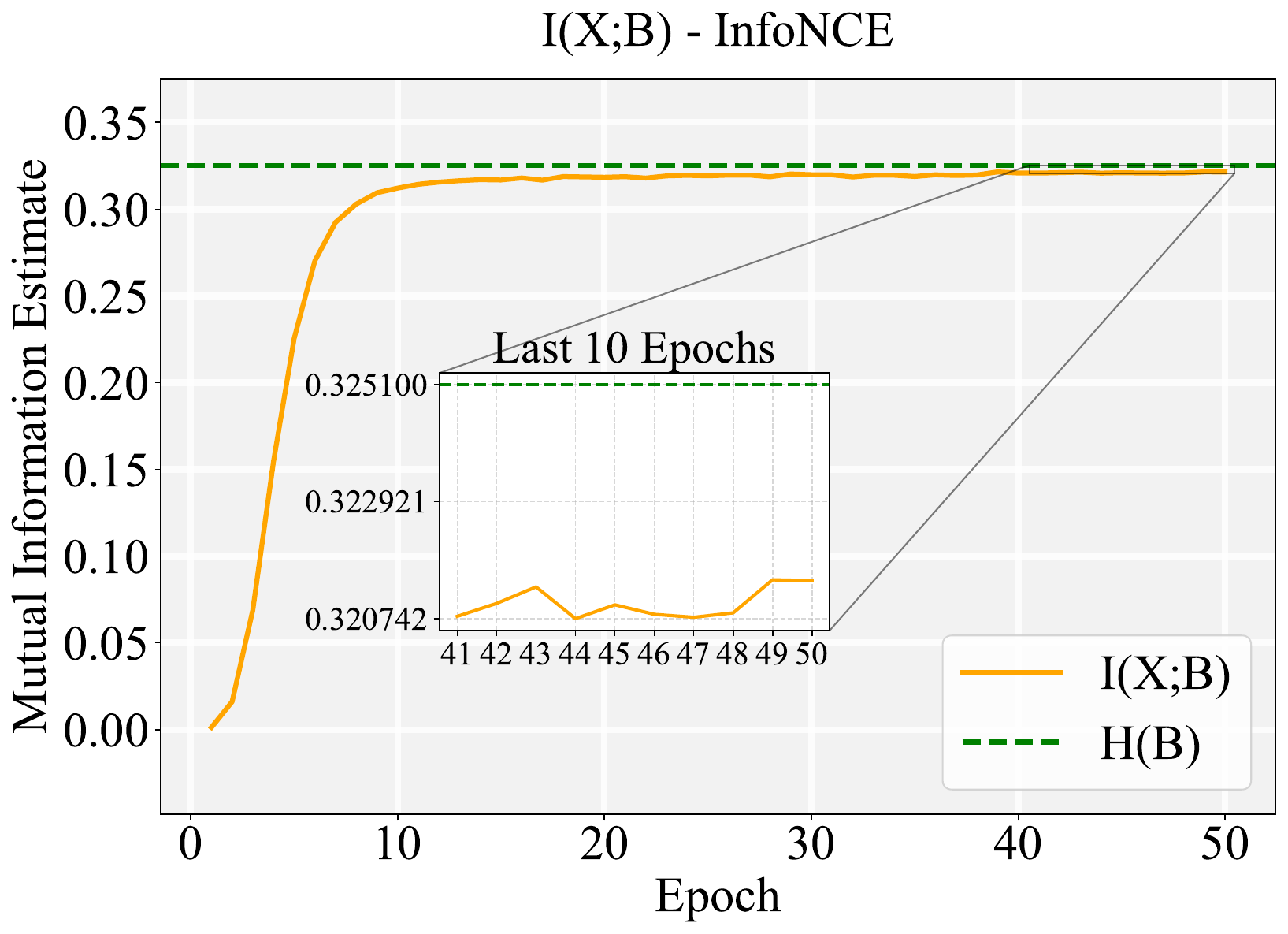}
        \\ \small (a) $I(X;B)$ vs. $H(B)$
    \end{minipage}
    \hfill
    \begin{minipage}[t]{0.32\linewidth}
        \centering
        \includegraphics[width=\linewidth]{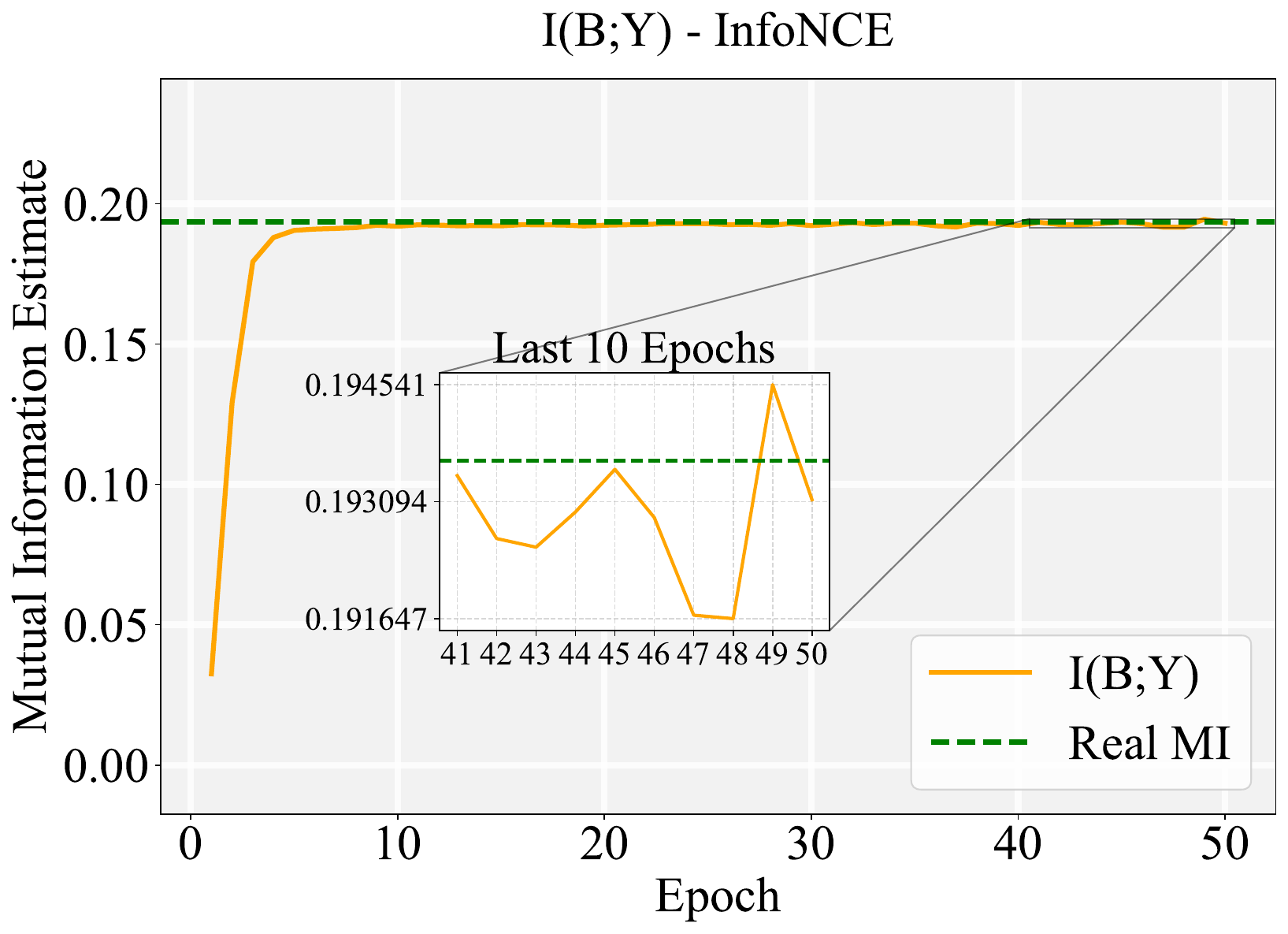}
        \\ \small (b) $I(B;Y)$ vs. Ground Truth
    \end{minipage}
    \hfill
    \begin{minipage}[t]{0.32\linewidth}
        \centering
        \includegraphics[width=\linewidth]{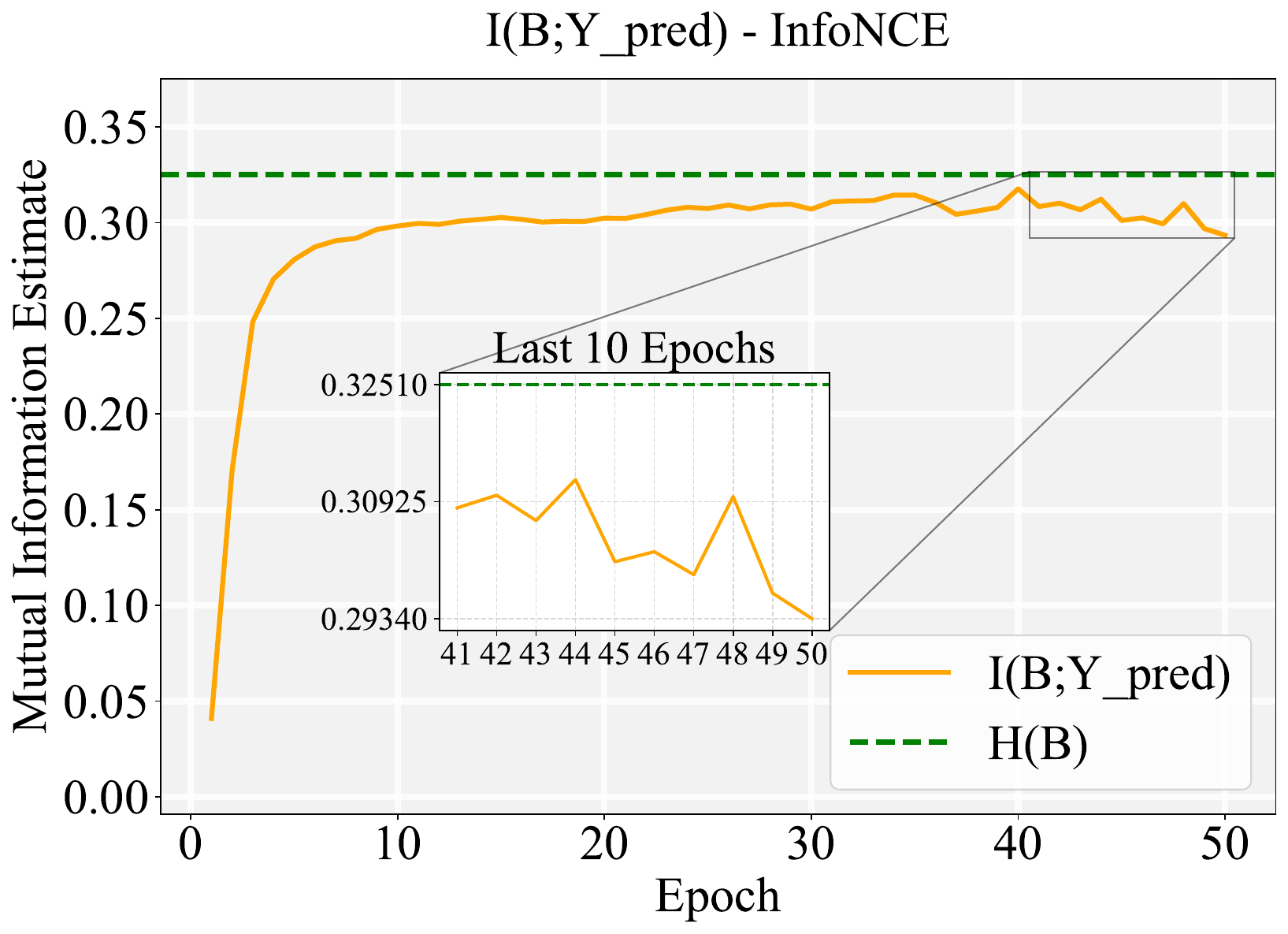} 
        \\ \small (c) $I(B;Y_{\text{pred}})$ vs. $H(B)$
    \end{minipage}
    \caption{
    \footnotesize{Validation of the InfoNCE estimator against theoretical bounds and ground truth. 
    (a) The $I(X;B)$ estimate closely matches its upper bound $H(B)$. 
    (b) The $I(B;Y)$ estimate is nearly identical to the analytically computed true MI. 
    (c) The $I(B;Y_{\text{pred}})$ estimate approaches its $H(B)$ bound, confirming reliance on the trigger.}
    }
    \label{fig:sanity_checks}
\end{figure*}

We first performed three sanity checks using a CIFAR-10 dataset with a 10\% Blend attack, with results visualized in Figure~\ref{fig:sanity_checks}. First, we estimated the MI between the input $X$ and a binary trigger-presence indicator $B$, denoted $I(X;B)$. Our estimate of $0.3208$ nats closely approaches the theoretical upper bound, $H(B) \approx 0.3251$ nats, resulting in a minimal estimation error of $\approx 1.3\%$. Second, we compared our estimate for the trigger-label dependency, $I(B;Y)$, against its analytically computed ground-truth value of $0.1936$ nats. Our InfoNCE estimate of $0.1929$ nats demonstrates a negligible error of only $\approx 0.36\%$, confirming the estimator's high accuracy on structured inputs. Finally, our estimate for $I(B;Y_{\text{pred}})$ ($\approx 0.3025$ nats) was again shown to be close to the $H(B)$ bound, validating that the model's predictions are strongly influenced by the trigger signal, as expected.

\subsubsection{Validation via Fano's Inequality and DPI}
\begin{figure}[t]
    \centering
    \includegraphics[width=0.6\linewidth]{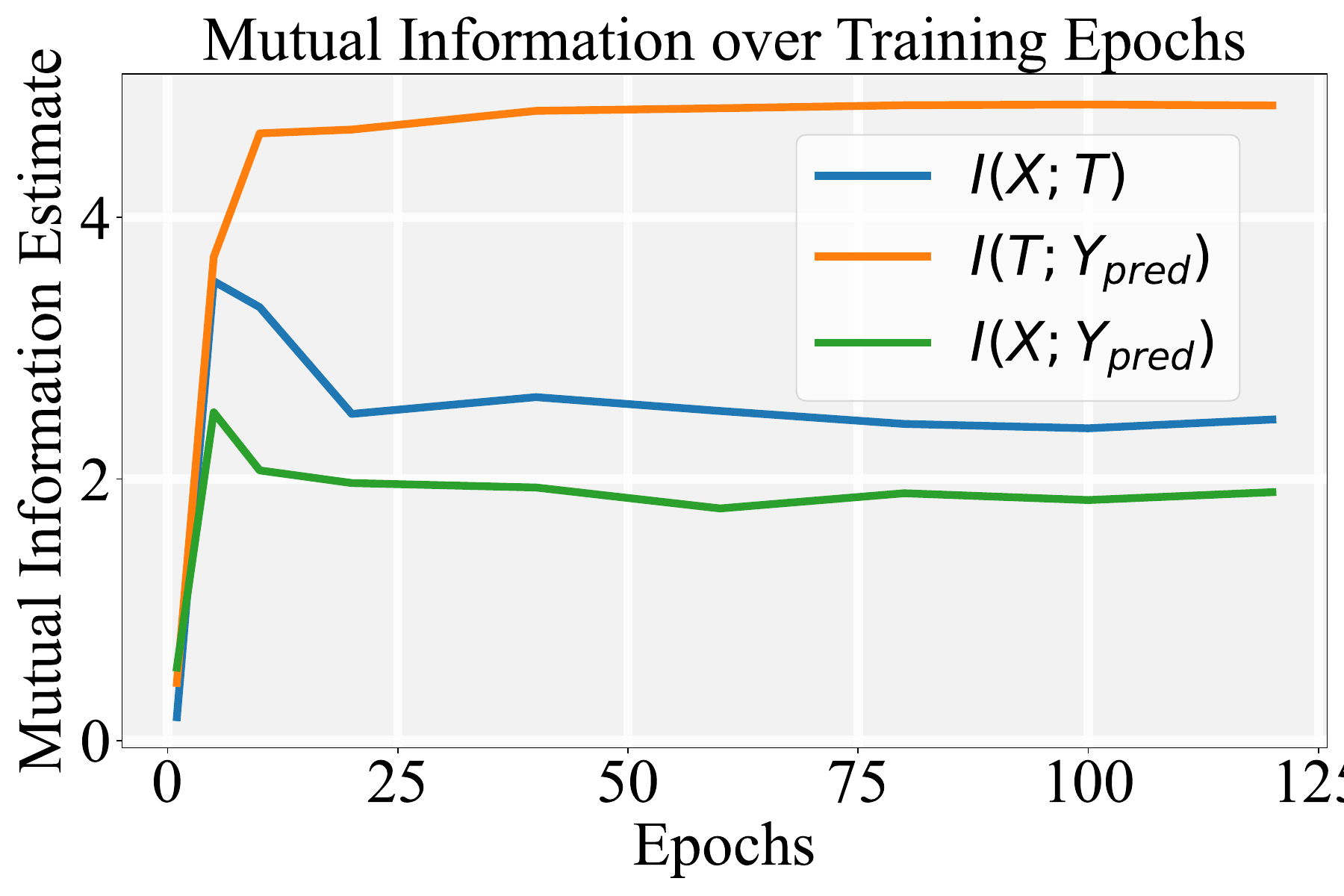} 
    \caption{\footnotesize{Dynamics of $I(X;T)$, $I(T;Y_{\text{pred}})$, and $I(X;Y_{\text{pred}})$ for the entire dataset under the Blend attack. The plot confirms two key theoretical principles: (1) The Data Processing Inequality holds, as $I(X;Y_{\text{pred}}) \le I(X;T)$ at all epochs. (2) The Fano bound is satisfied, as the final $I(X;Y_{\text{pred}})$ ($\approx 1.9$ nats) is well above the 1.3262 nats required for 80\% accuracy.}}
    \label{fig:mi_all_dpi}
\end{figure}

We further assessed the framework's theoretical consistency, as shown in Figure~\ref{fig:mi_all_dpi}. First, we derived the theoretical lower bound of MI required for the model to achieve its observed 80\% accuracy on the 10-class imbalanced dataset. Based on the class entropy $H(Y) \approx 2.266$ nats and error $P_e = 0.2$, Fano's inequality dictates a minimum required MI of $I(X;Y) \ge 1.3262$ nats. Our estimated $I(X;Y_{\text{pred}})$ (converging at $\approx 1.9$ nats) is comfortably above this theoretical lower bound, supporting the fidelity of our estimations. Furthermore, we verified that our estimates adhere to the Data Processing Inequality (DPI), which mandates that information can only decrease through processing, i.e., $I(X;Y_{\text{pred}}) \leq I(X;T)$. Figure~\ref{fig:mi_all_dpi} confirms that this inequality consistently holds at all stages of training, with the $I(X;Y_{\text{pred}})$ curve remaining strictly below the $I(X;T)$ curve.

Together, these results demonstrate that our estimated MI values are not only accurate (vs. ground truth) but also theoretically grounded, satisfying both the Fano bound and the Data Processing Inequality. This establishes a reliable foundation for the analytical framework used in this paper.

\subsection{Dynamic Properties Under Different Attacks}
We extend our analysis to examine MI dynamics under other representative attacks, complementing our experiments on BadNets in Section~\ref{sec:4.2}.

\begin{figure}[!htbp]
    \centering
    \renewcommand{\arraystretch}{1.3} 
    \setlength{\tabcolsep}{3pt} 
    \resizebox{0.45\textwidth}{!}{
    \begin{tabular}{p{0.4cm}c c} 
        \multicolumn{1}{c}{} &
        \multicolumn{1}{c}{\textbf{ $I(X;T)$}} & 
        \multicolumn{1}{c}{\textbf{$I(T;Y_{\text{pred}})$}} \\ 
        \raisebox{0.5\height}{\rotatebox{90}{\centering \colorbox{gray!20}{Blend-10\%}}} &
        \includegraphics[width=0.45\linewidth]{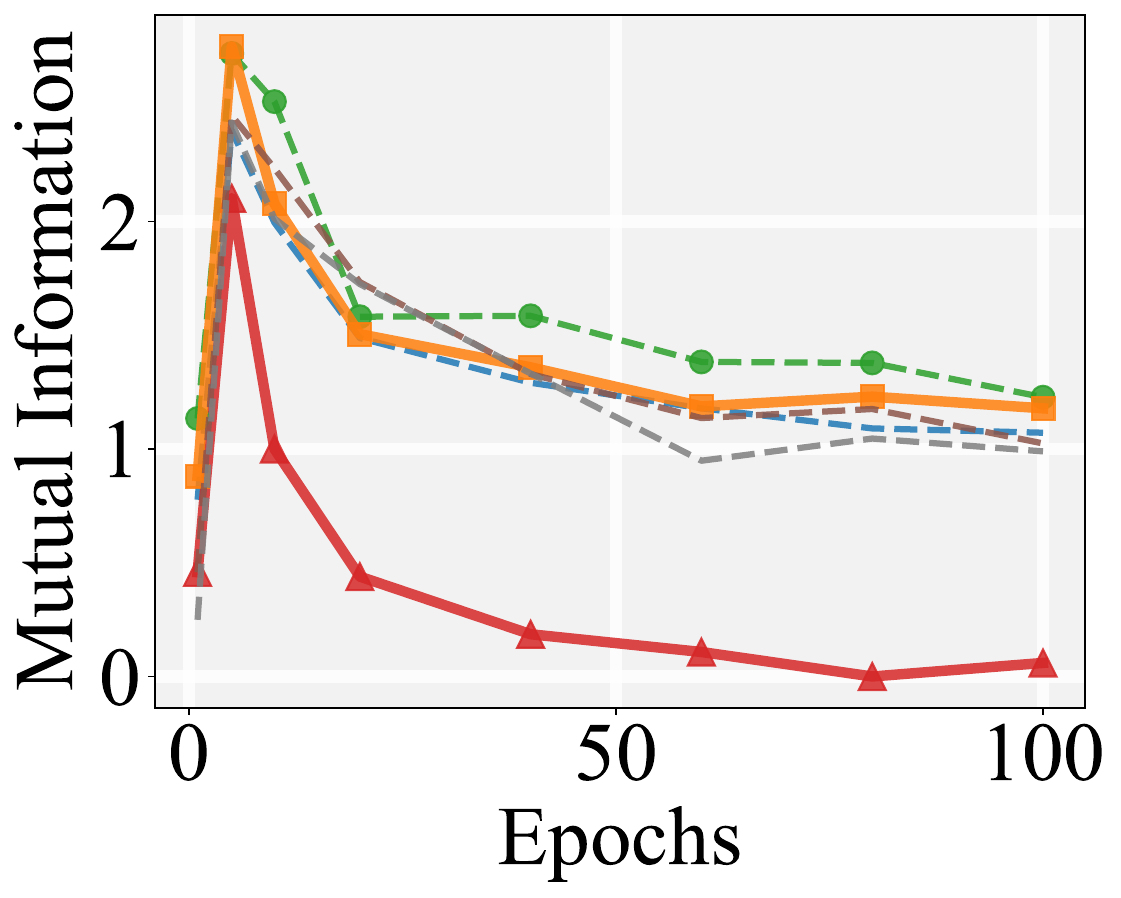} &
        \includegraphics[width=0.45\linewidth]{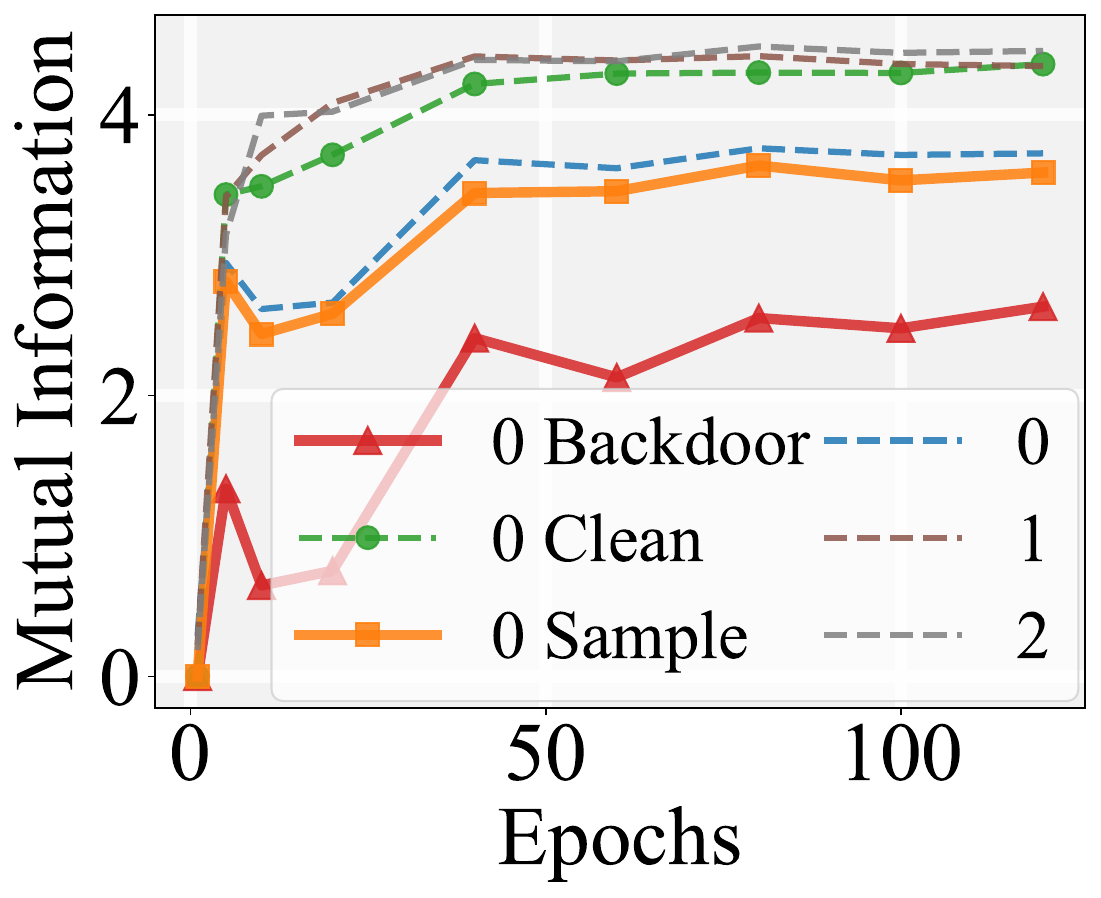} \\
        \raisebox{0.45\height}{\rotatebox{90}{\centering \colorbox{gray!20}{WaNet-10\%}}} &
        \includegraphics[width=0.45\linewidth]{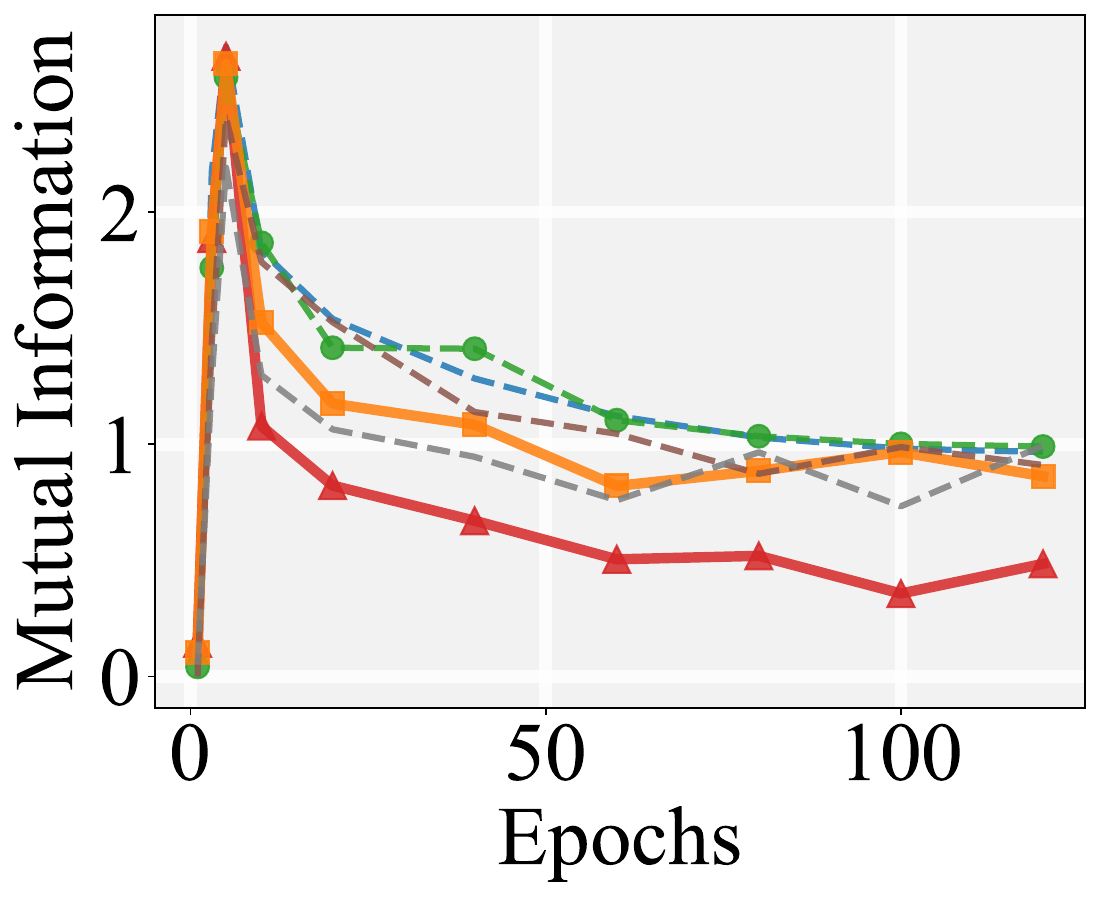} &
        \includegraphics[width=0.45\linewidth]{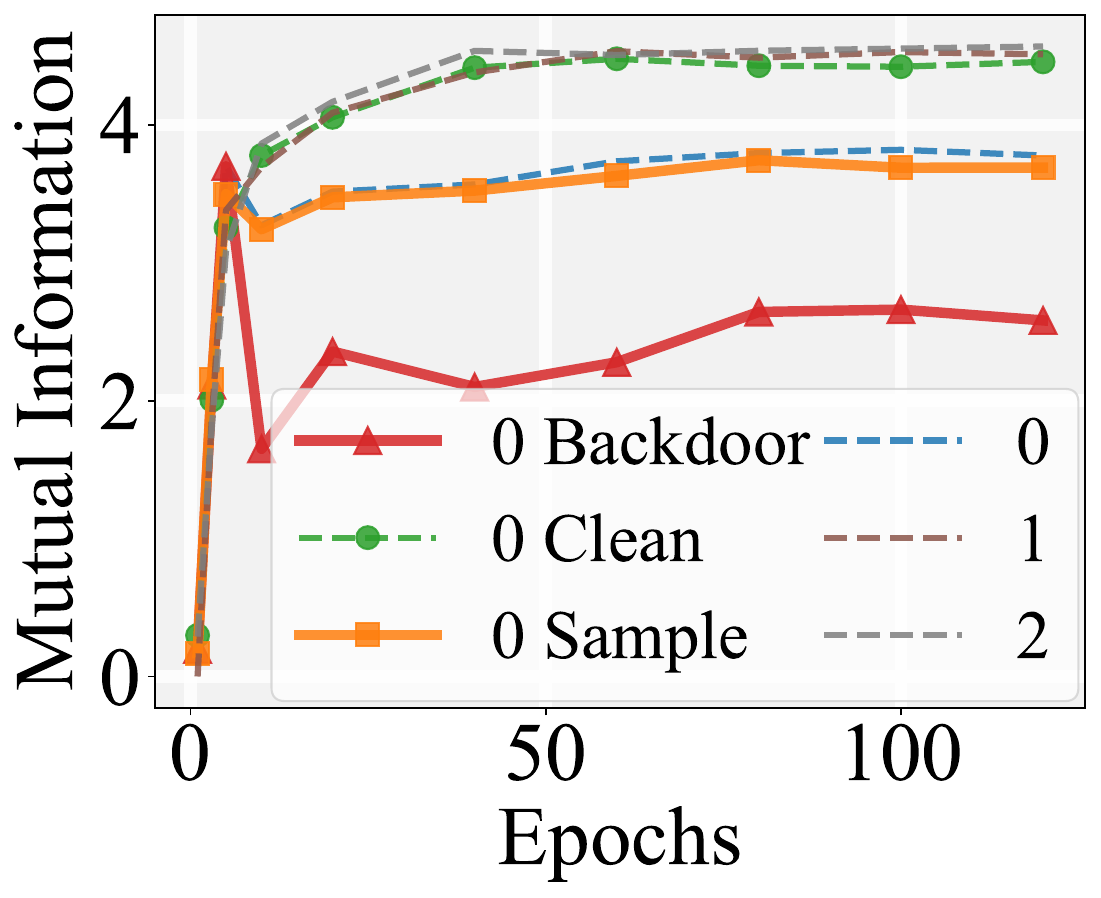} \\
        \raisebox{0.7\height}{\rotatebox{90}{\centering \colorbox{gray!20}{LC-25\%}}} &
        \includegraphics[width=0.45\linewidth]{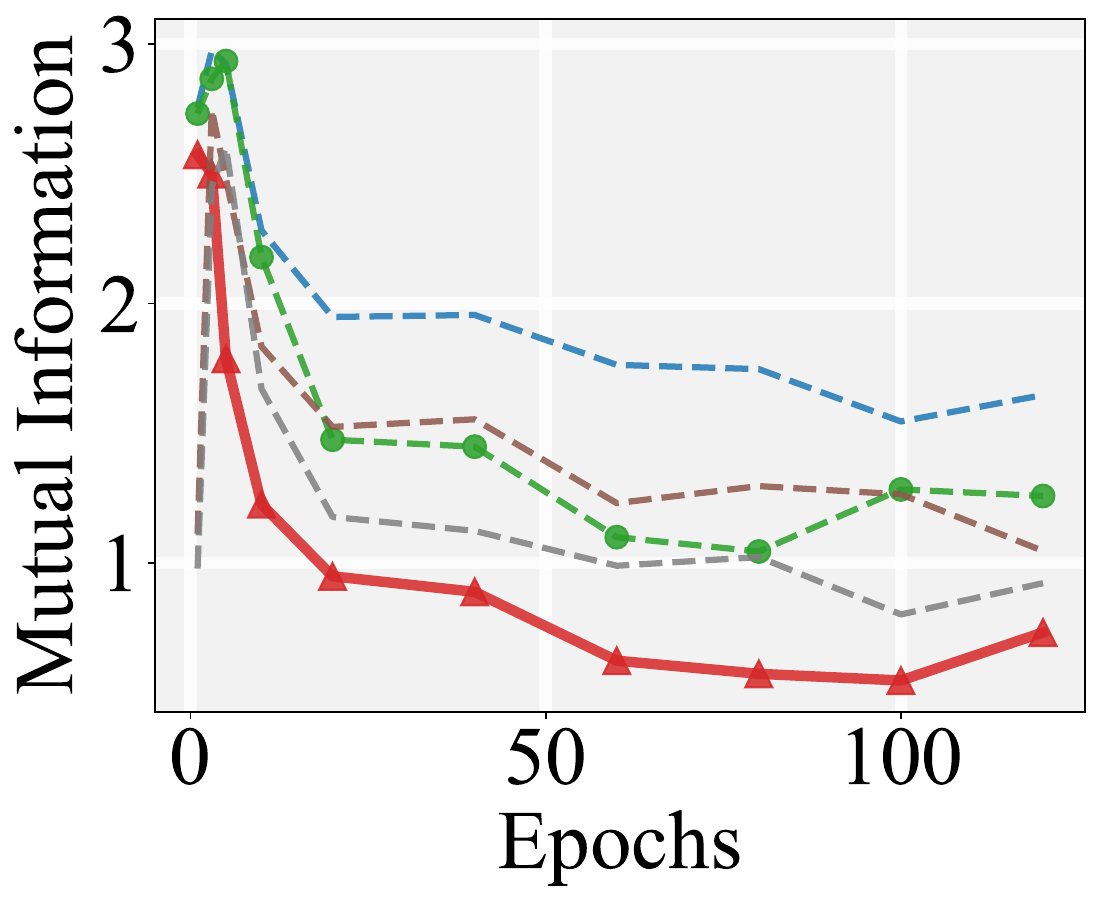} &
        \includegraphics[width=0.45\linewidth]{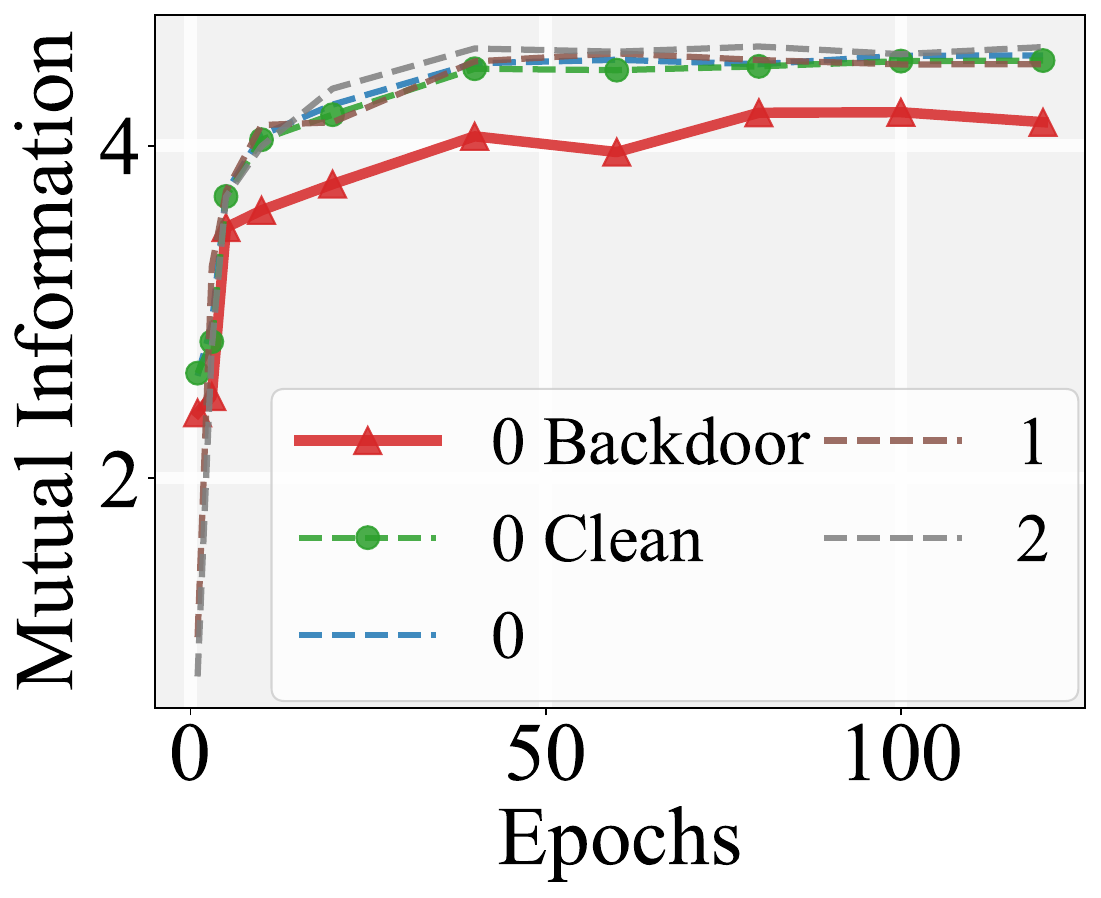} \\
    \end{tabular}}
    \vspace{-3mm}
    \caption{\footnotesize{Comparison of MI dynamics under different backdoor attacks with 10\% poisoning ratio (25\% for LC). The rows represent the attack methods, and the columns represent the MI terms \( I(X;T) \) and \( I(T;Y_{\text{pred}}) \). For the Blend attack, global perturbations disrupt semantic features, leading to consistently lower MI for backdoor samples. For the WaNet attack, localized perturbations allow the model to retain more information from the trigger, resulting in higher \( I(X;T) \) compared to Blend, though backdoor \( I(T; Y_{\text{pred}}) \) still drops below clean samples. For the LC attack, adversarial perturbations introduce additional variability, causing higher \( I(X;T) \) for class 0, but lower for backdoor samples. Backdoor \( I(T; Y_{\text{pred}}) \) is slightly lower than that of clean samples due to inconsistencies from the perturbations.}}
    \label{fig:attacks_comparison_mi}
    \vspace{-4mm}
\end{figure}

\paragraph{MI Dynamics.} 
As shown in Figure~\ref{fig:attacks_comparison_mi}, each attack presents distinct MI patterns that reflect its underlying mechanism. 
For the \textbf{Blend attack}, the global additive trigger disrupts the entire image, degrading semantic features in backdoor samples. As training progresses, the model quickly discards these weakened semantics and retains only the trigger. This leads to consistently lower \(I(X;T)\) for backdoor samples, which approaches nearly zero at convergence. Consequently, class 0's overall \(I(X;T)\) becomes indistinguishable from other clean classes, as the trigger fails to provide class-specific contributions. Similarly, \(I(T;Y_{\text{pred}})\) remains suppressed, indicating that predictions rely on a compressed, low-information trigger representation.
In contrast, the \textbf{WaNet attack} applies subtle spatial warping, embedding the trigger in a less disruptive manner. Semantic information is largely preserved early in training, resulting in similar MI trends between backdoor and clean samples. As the model begins compressing redundant features, the localized trigger in WaNet remains informative, leading to a moderate drop in \(I(X;T)\) and a final value higher than in Blend. However, \(I(T;Y_{\text{pred}})\) for backdoor samples still falls below that of clean samples at convergence, suggesting a shift toward trigger reliance for prediction.
The \textbf{LC attack} combines patch triggers with adversarial perturbations, creating backdoor samples that preserve their original semantic features while being redirected to the target class. As a result, class 0 exhibits higher \(I(X;T)\) during early training compared to clean classes, due to the increased variability introduced by these semantically diverse backdoor samples. However, the \(I(X;T)\) of backdoor samples is actually lower than that of clean samples. This occurs because the model learns to rely primarily on the patch trigger for prediction, gradually discarding the original semantic features as redundant. The result is a compressed representation that retains only the shortcut signal, reducing the information retained from the input. Similarly, \(I(T;Y_{\text{pred}})\) for class 0 starts higher than clean classes due to increased prediction confidence, but eventually drops for backdoor samples due to inconsistencies introduced by the adversarial noise, increasing the uncertainty \(H(Y_{\text{pred}}|T)\).

\begin{figure*}[tp]
    \centering
    \renewcommand{\arraystretch}{1.5} 
    \begin{tabular}{p{0.5cm}ccc} 
        \raisebox{0.5\height}{\rotatebox{90}{\centering \colorbox{gray!20}{Blend-10\%}}} &
        \includegraphics[width=0.25\linewidth]{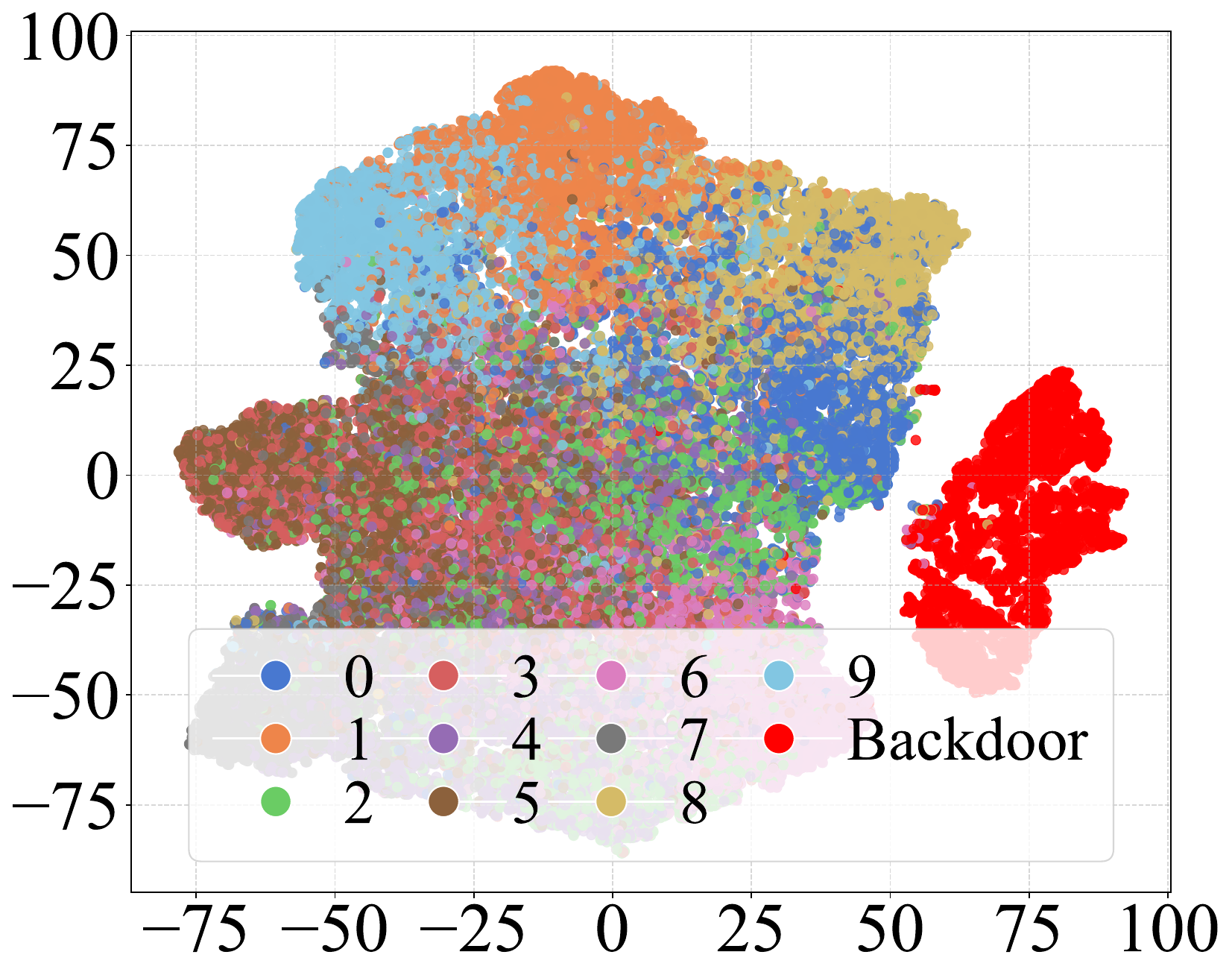} &
        \includegraphics[width=0.25\linewidth]{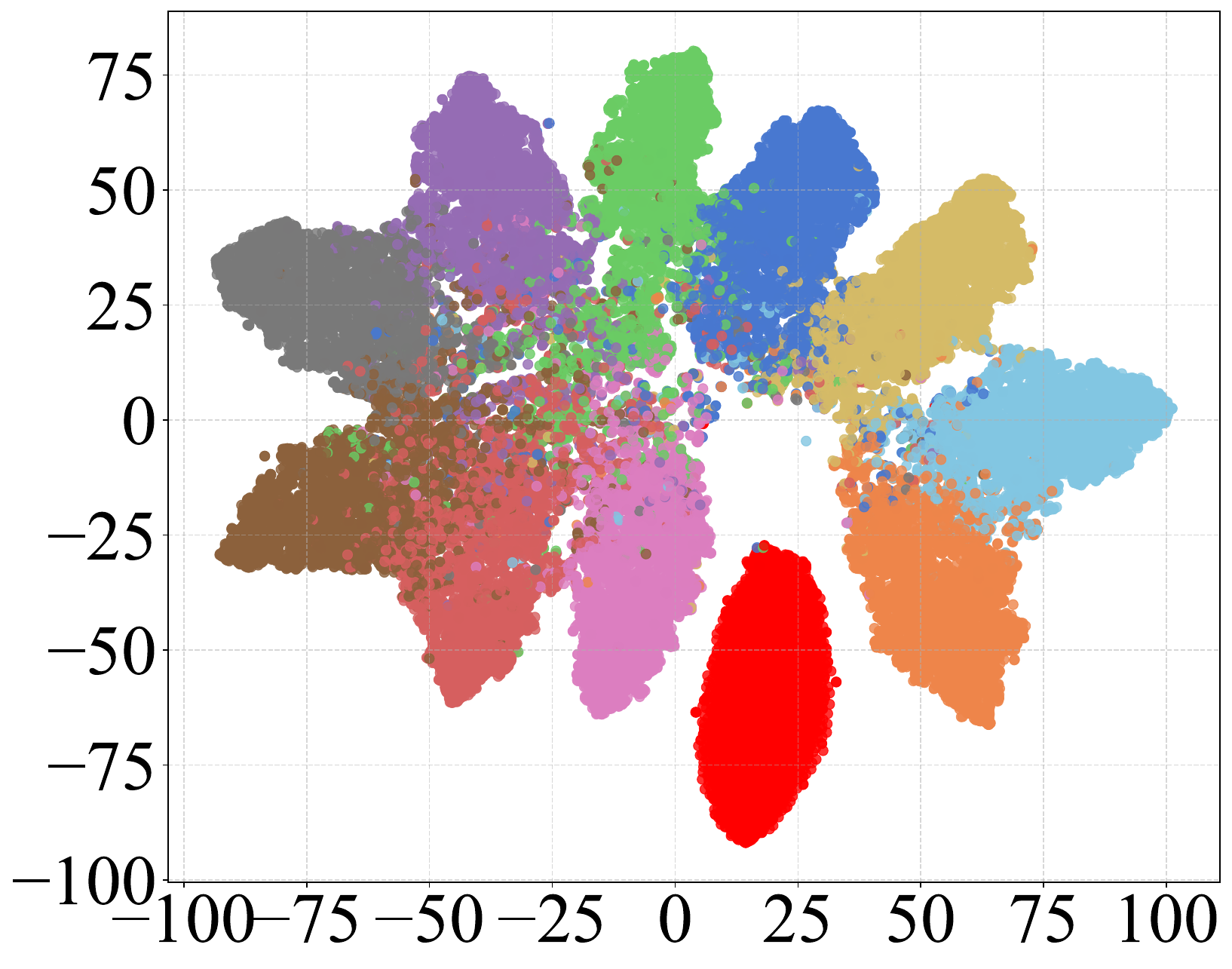} &
        \includegraphics[width=0.25\linewidth]{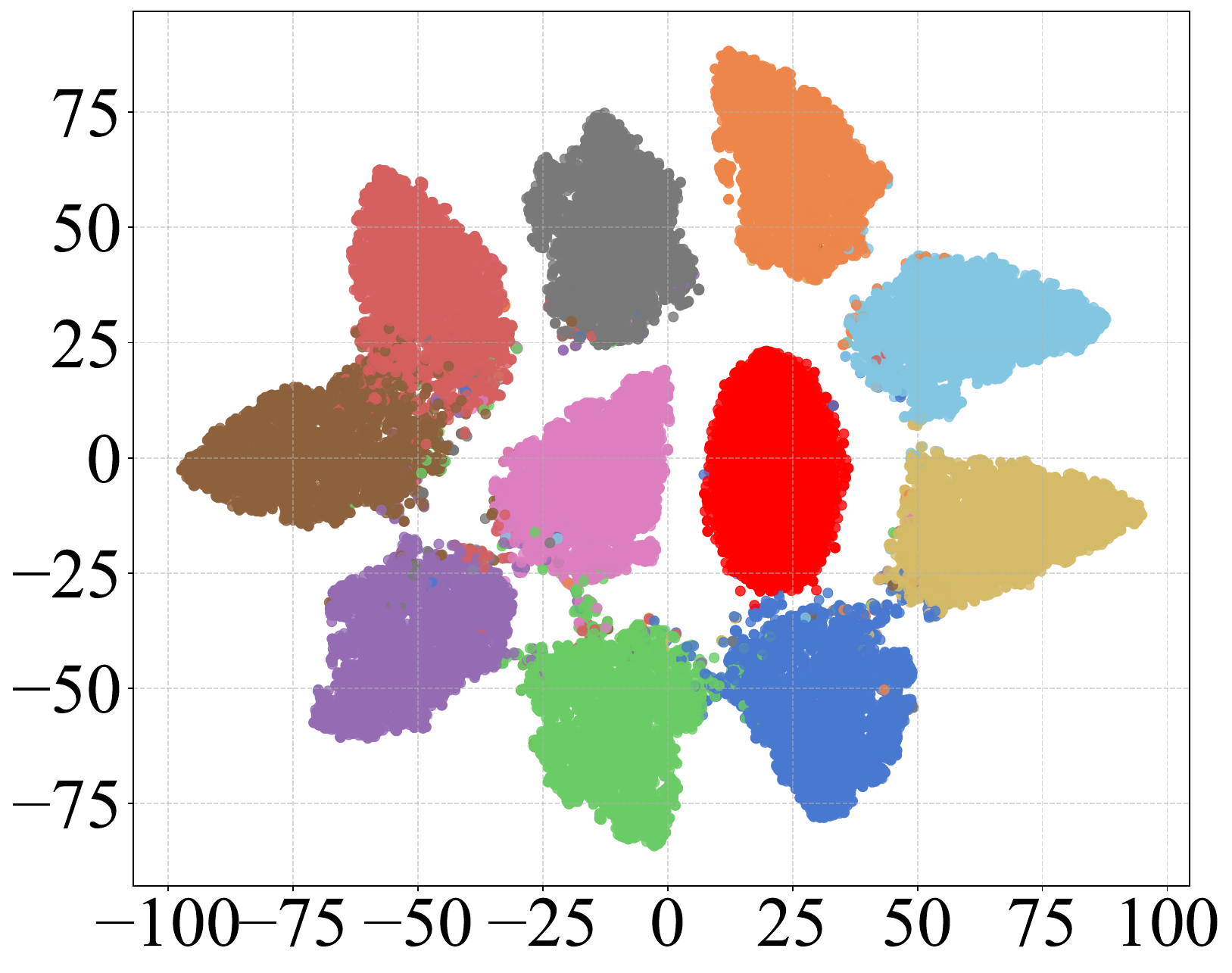} \\ 
        \raisebox{0.5\height}{\rotatebox{90}{\centering \colorbox{gray!20}{WaNet-10\%}}} &
        \includegraphics[width=0.25\linewidth]{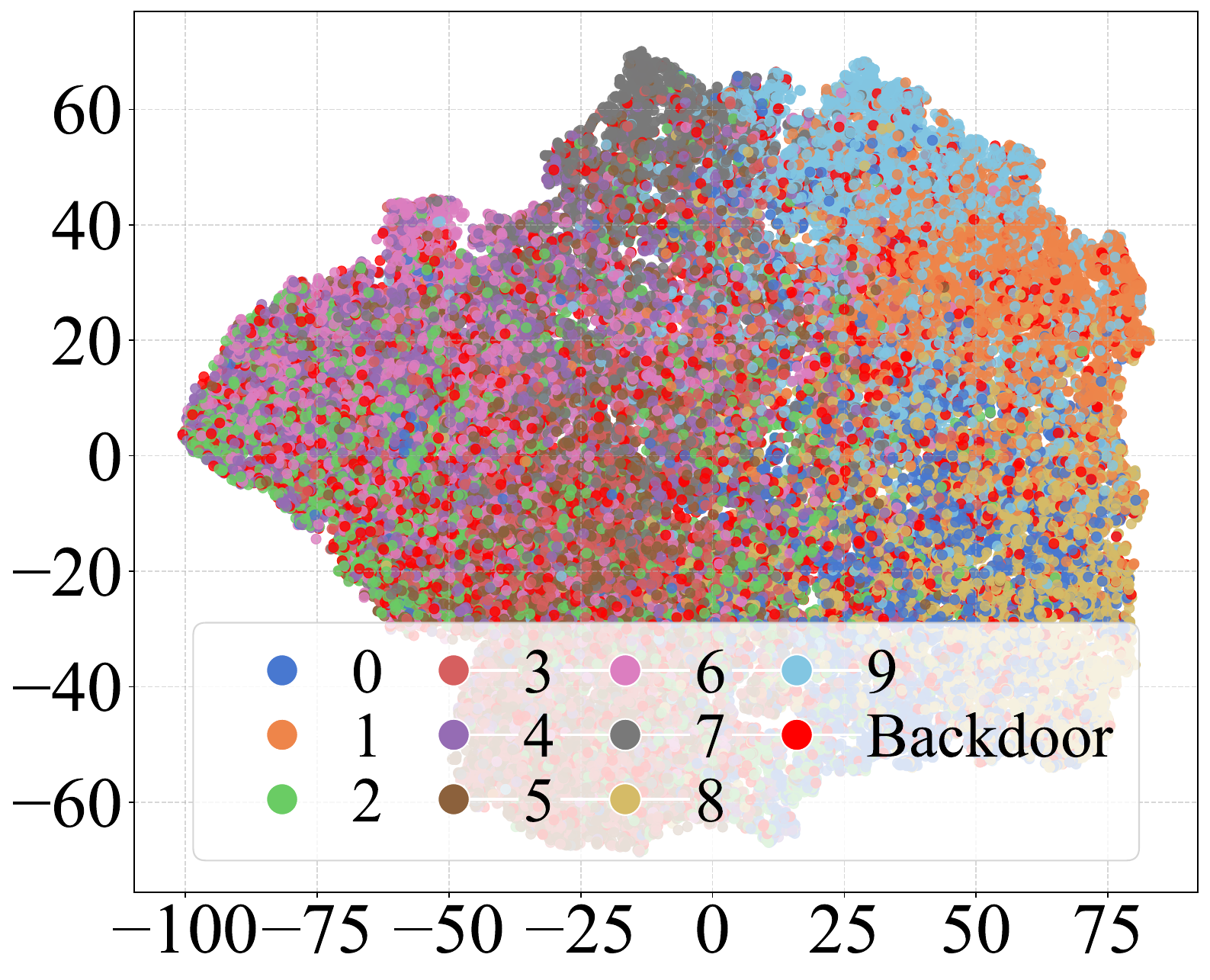} &
        \includegraphics[width=0.25\linewidth]{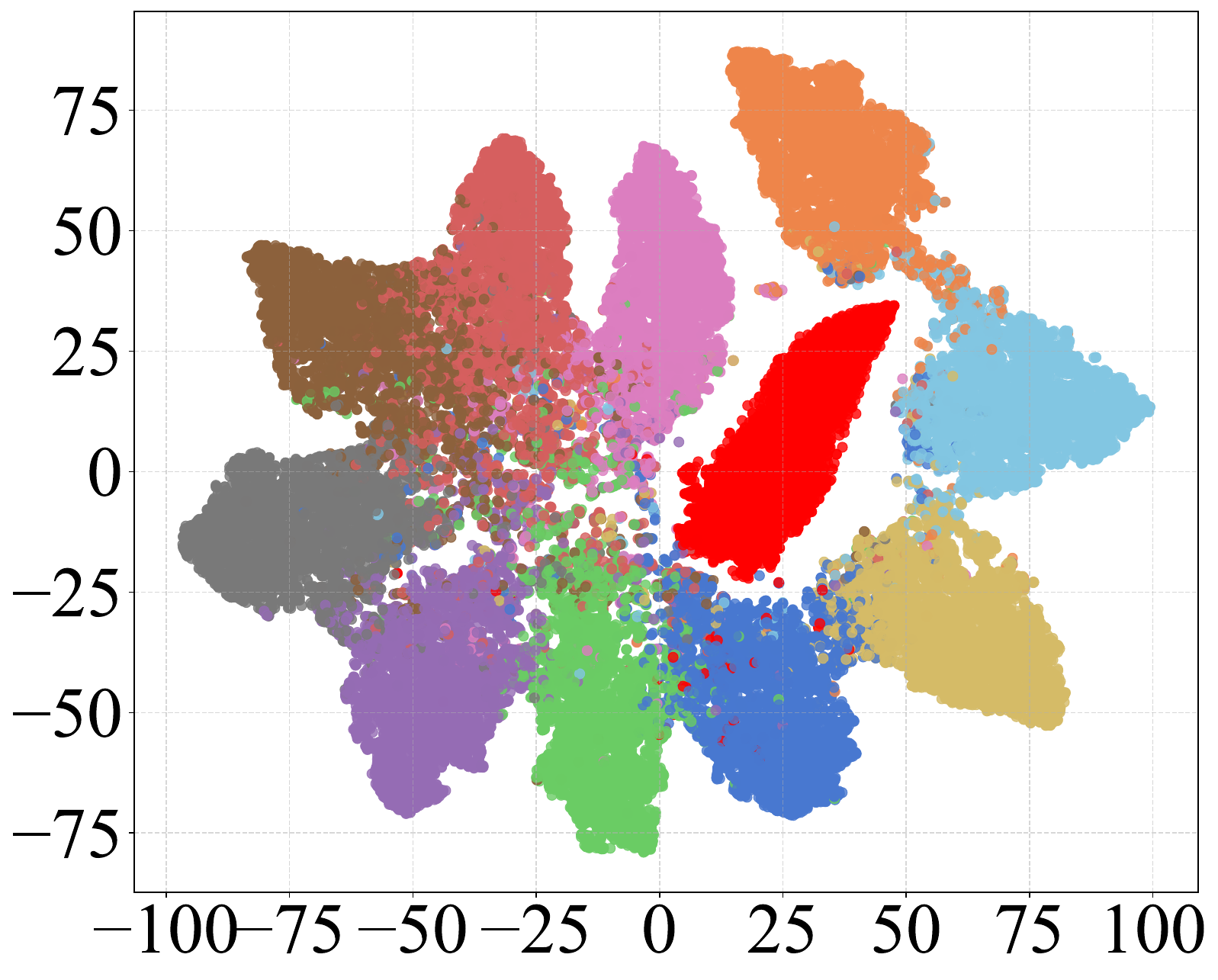} &
        \includegraphics[width=0.25\linewidth]{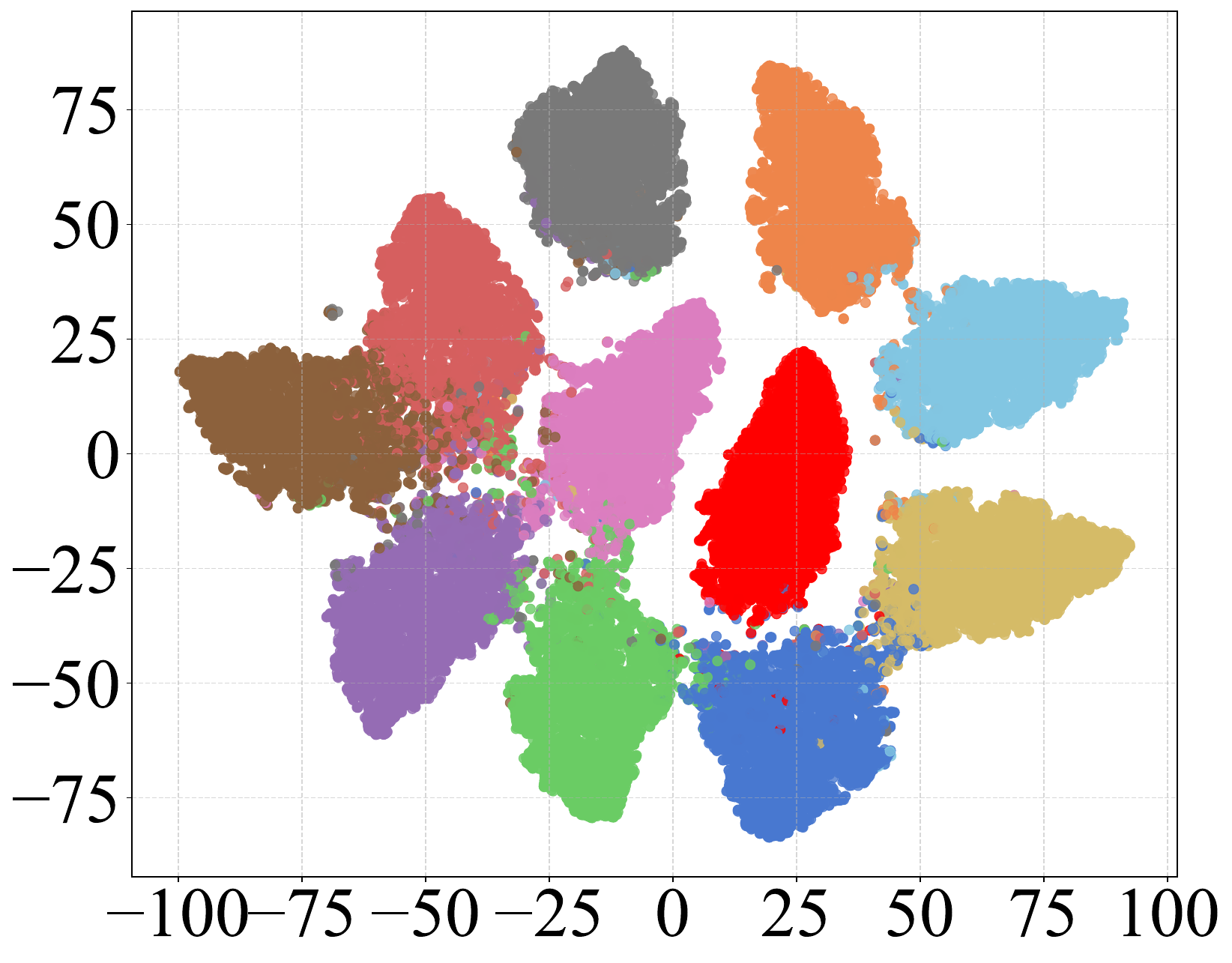} \\ 
        \raisebox{0.5\height}{\rotatebox{90}{\centering \colorbox{gray!20}{LC-25\%}}} &
        \includegraphics[width=0.25\linewidth]{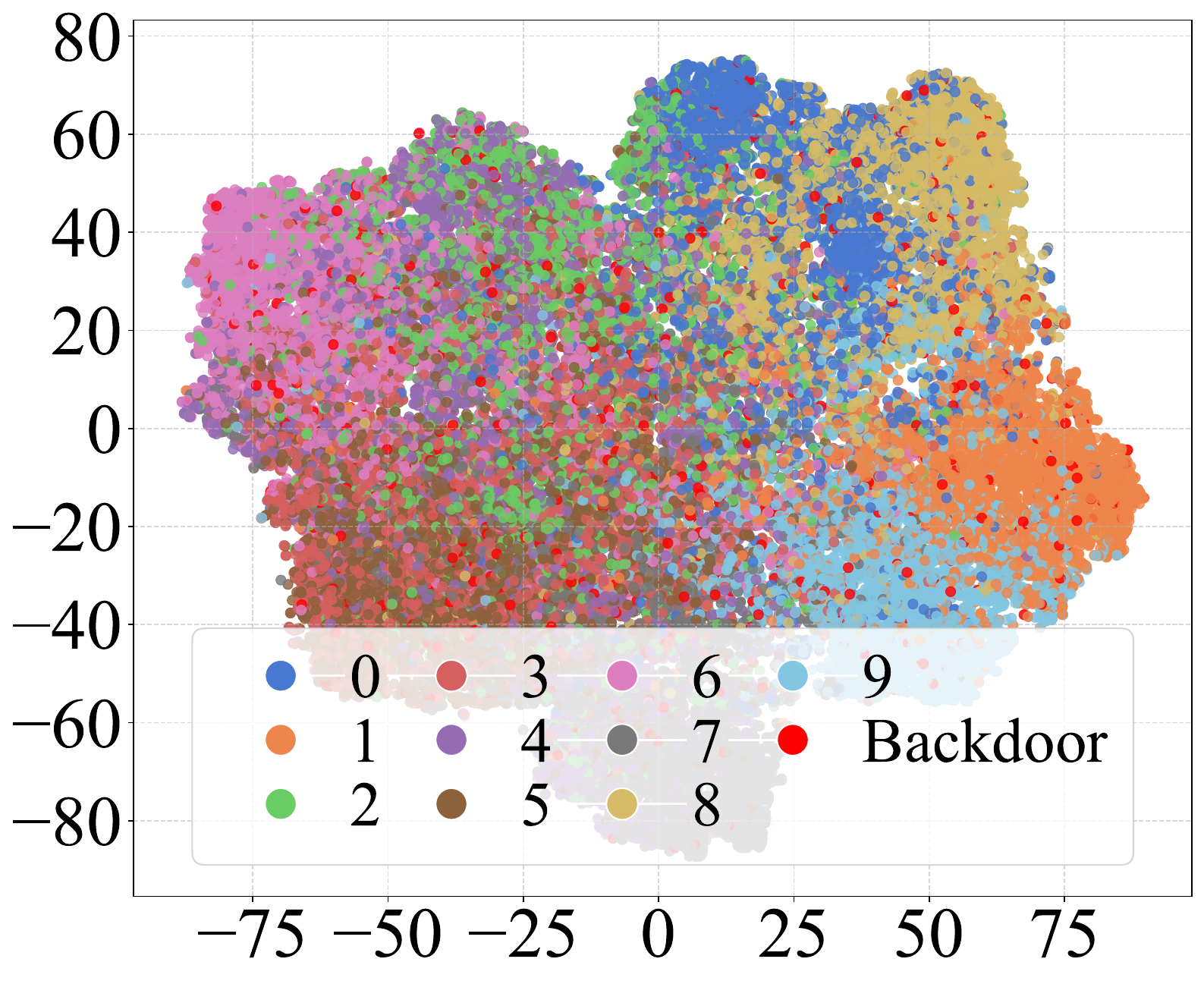} &
        \includegraphics[width=0.25\linewidth]{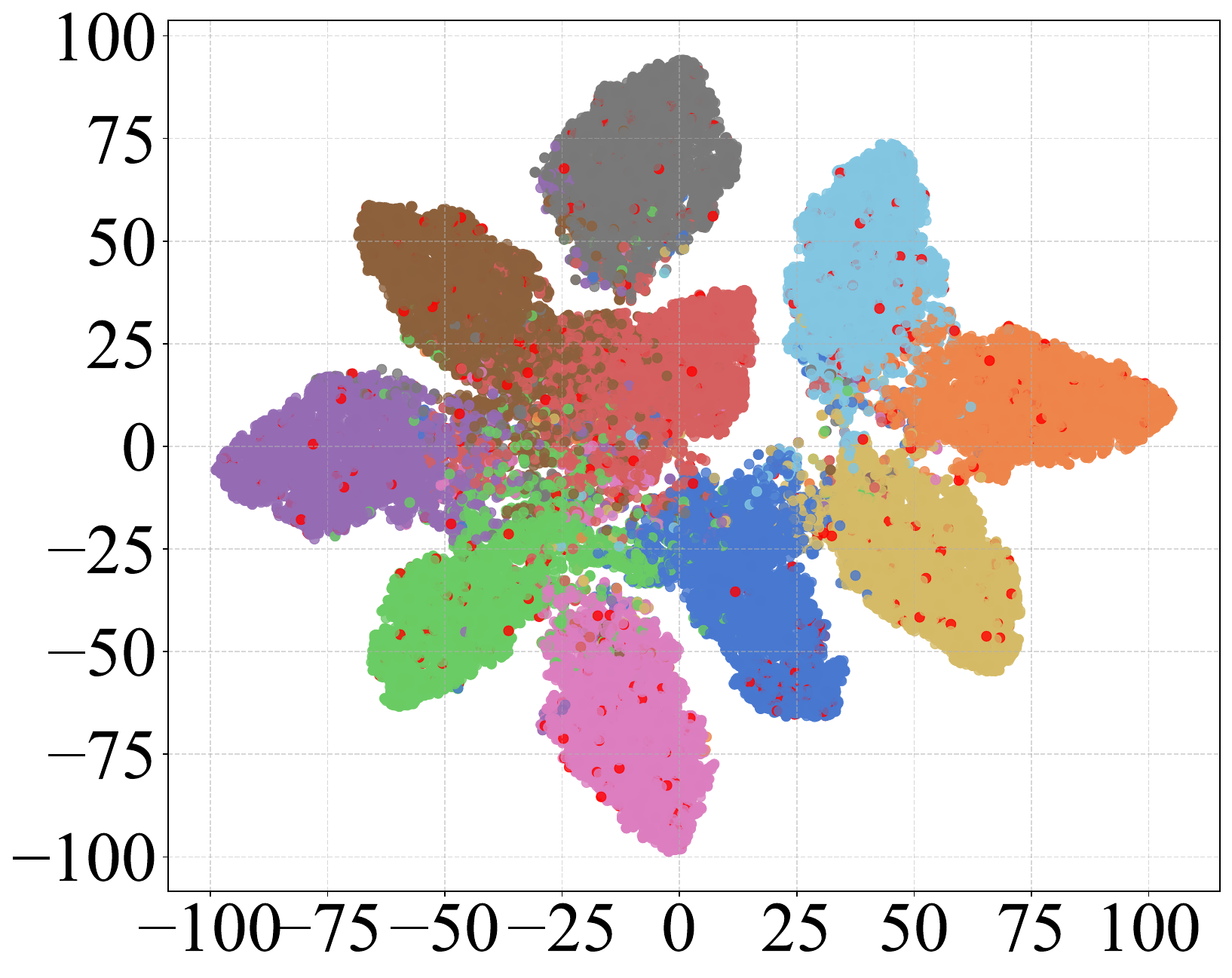} &
        \includegraphics[width=0.25\linewidth]{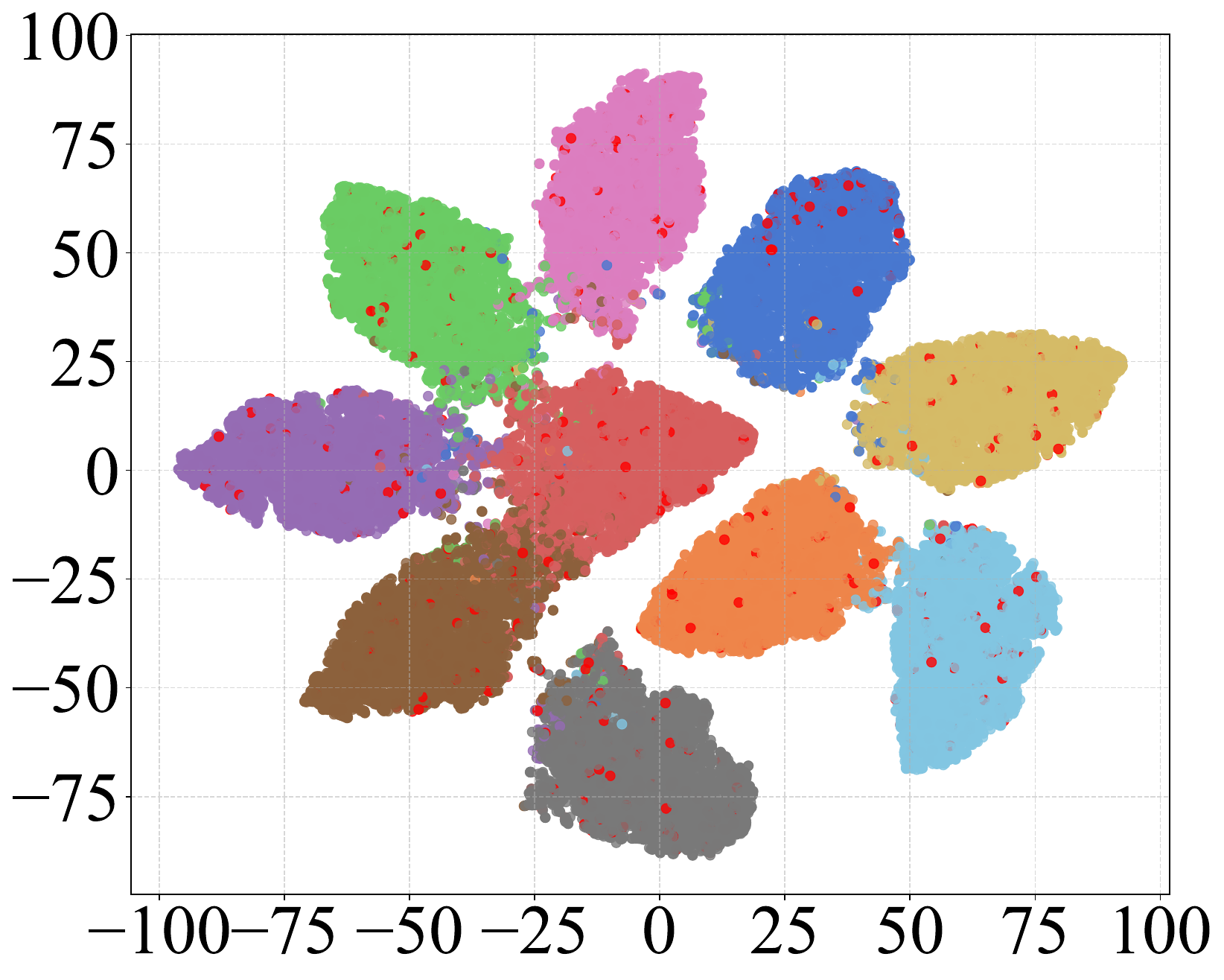} \\ 
        & (a) Early learning phase & (b) Compression phase & (c) Convergence phase
    \end{tabular}
    \caption{\footnotesize{t-SNE visualization of the last hidden layer representations \(T\) under different attacks (CIFAR-10, 10\% poison ratio). Each row represents a backdoor attack method. For the \textbf{Blend attack}, backdoor samples form a distinct cluster that is minimally integrated into the target class, reflecting low \(I(X;T)\). In \textbf{WaNet}, backdoor samples retain more semantic features, resulting in a less compact cluster and slightly higher \(I(X;T)\) compared to Blend. For the \textbf{LC attack}, adversarial perturbations cause backdoor samples to initially disperse, but they merge into the target class cluster  (the class targeted by the adversarial perturbations during trigger generation) as training progresses, increasing \(I(X;T)\) for class 0.}}
    \label{fig:attacks_comparison_tsne}
\end{figure*}

\paragraph{Clustering Dynamics.}
We further examine the representation clustering dynamics under various backdoor attacks using t-SNE visualizations, as illustrated in Figure~\ref{fig:attacks_comparison_tsne}. These visualizations reveal how backdoor samples' representation patterns evolve through training phases under Blend, WaNet, and LC attacks, with each attack exhibiting distinct clustering behaviors shaped by the interaction between semantic features and triggers.
For the \textbf{Blend attack}, global perturbations disrupt pixel-level semantics, causing backdoor samples to quickly form a distinct and compact cluster during the early learning phase, clearly separate from the target class cluster. As training progresses, the weak and diffuse trigger leads to strong compression, resulting in a small, detached cluster with limited integration into the target class throughout training. This minimal integration is consistent with the consistently low \(I(X;T)\) observed in the MI dynamics analysis, as the model discards semantic information in favor of the compressed trigger representation.
For the \textbf{WaNet attack}, subtle spatial warping preserves semantics, causing backdoor and clean samples to remain intermixed during early training. In contrast to Blend, the WaNet backdoor samples initially cluster loosely, retaining significant semantic features of their original classes. During compression and convergence phases, the model begins separating backdoor samples, forming a loosely compacted cluster that reflects the retention of semantic features. This less compact clustering compared to Blend explains WaNet's relatively higher \(I(X;T)\), as the model maintains more informative representations throughout training.
For the \textbf{LC attack}, adversarial perturbations initially scatter backdoor samples among clean clusters. Over time, these samples progressively merge into the target class cluster, driven by the adversarially induced semantic alignment. This integration process increases the MI \(I(X;T)\) within class 0 and backdoor samples, as the representations incorporate features from multiple classes. The merging behavior reflects how the model learns to associate diverse semantic patterns with the target class through the adversarial perturbation framework.

\subsection{MI Dynamics Under Different Poisoning Ratios}
\label{sec:low_poison_ratio}
\begin{figure*}[h]
    \centering
    \renewcommand{\arraystretch}{1.5} 
    \setlength{\tabcolsep}{3pt} 
    \begin{tabular}{c c c c} 
        \multicolumn{2}{c}{\colorbox{gray!20}{BadNets-1\%}} &
        \multicolumn{2}{c}{\colorbox{gray!20}{Blend-1\%}} \\ 
        \includegraphics[width=0.2\linewidth]{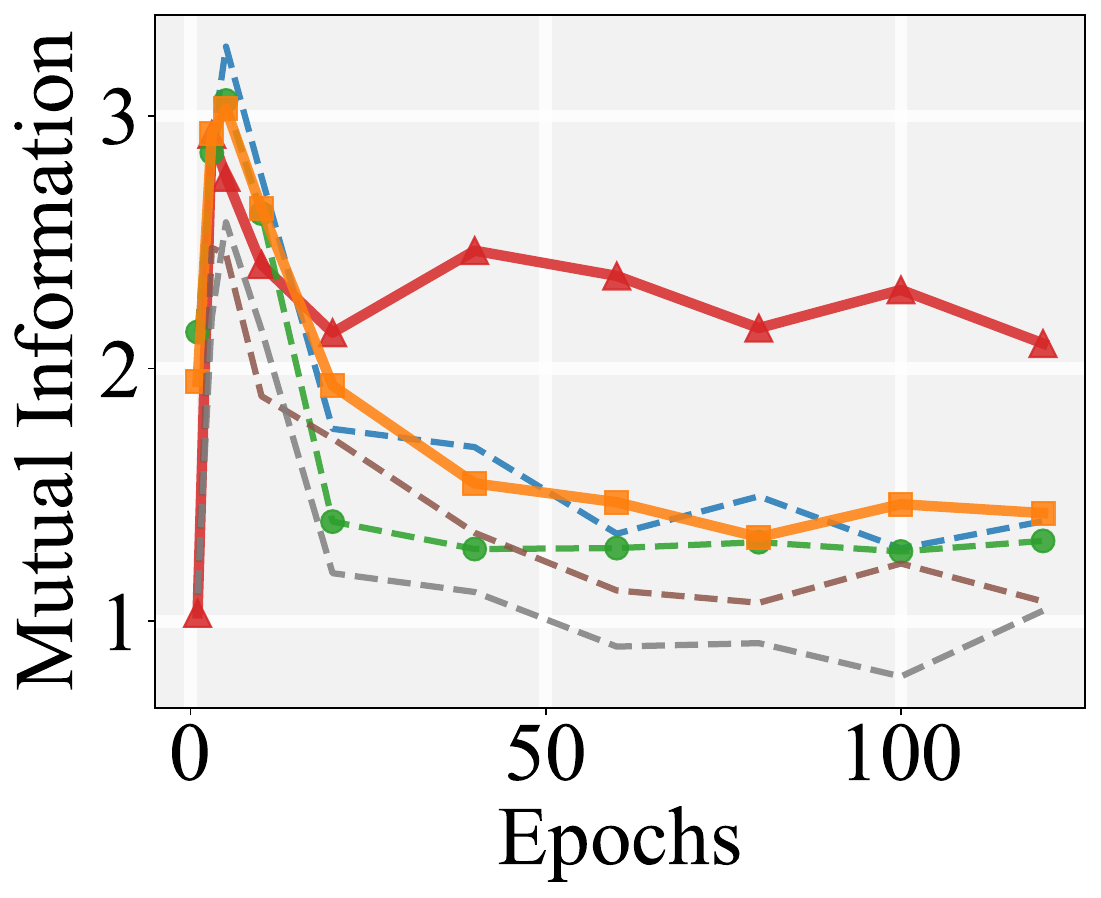} &
        \includegraphics[width=0.2\linewidth]{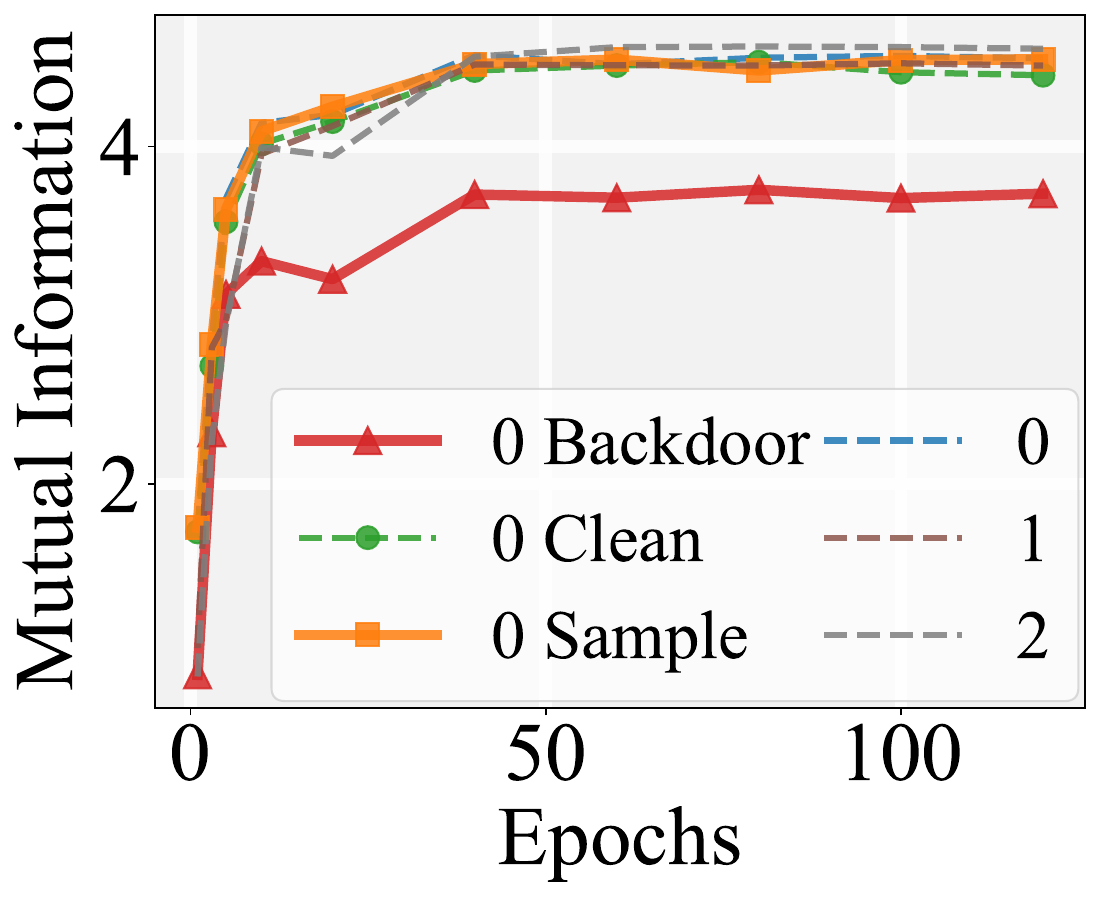} &
        \includegraphics[width=0.2\linewidth]{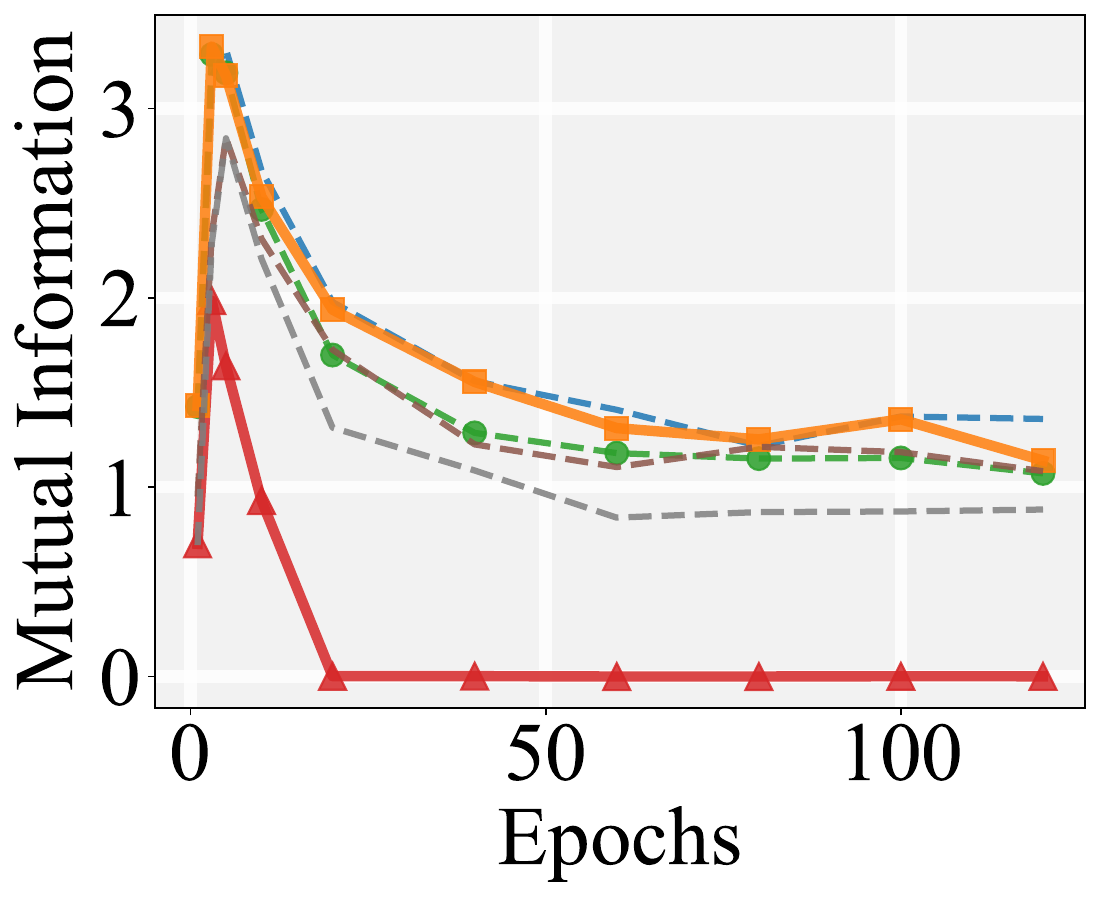} &
        \includegraphics[width=0.2\linewidth]{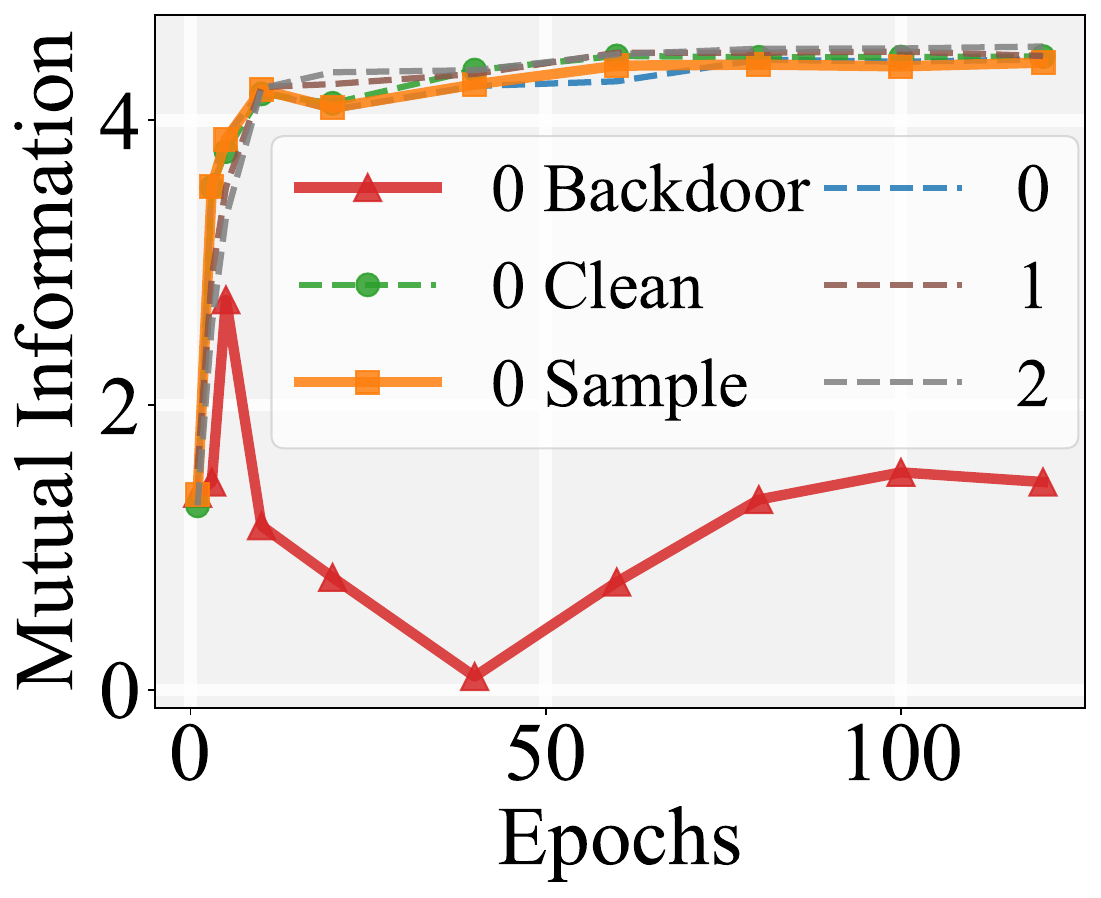} \\

         \multicolumn{2}{c}{\colorbox{gray!20}{WaNet-1\%}} &
        \multicolumn{2}{c}{\colorbox{gray!20}{LC-10\%}} \\
        \includegraphics[width=0.2\linewidth]{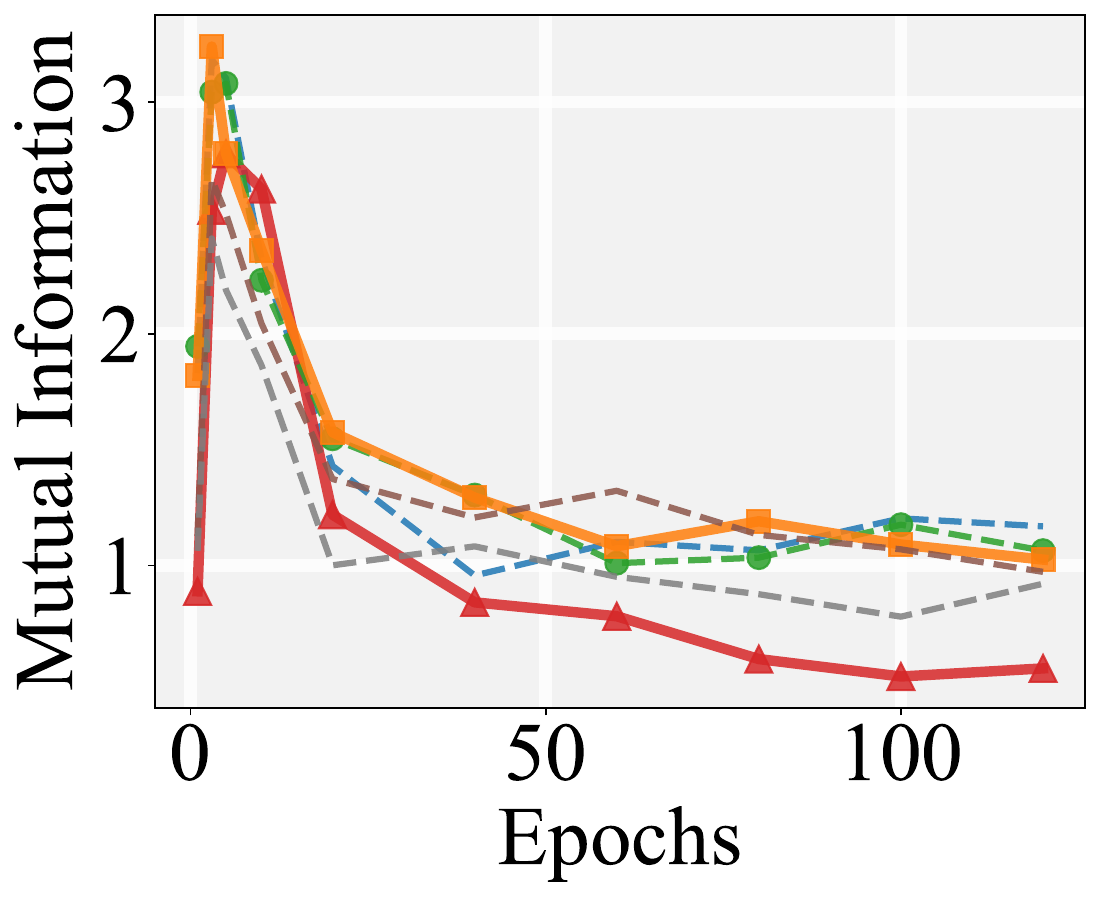} &
        \includegraphics[width=0.2\linewidth]{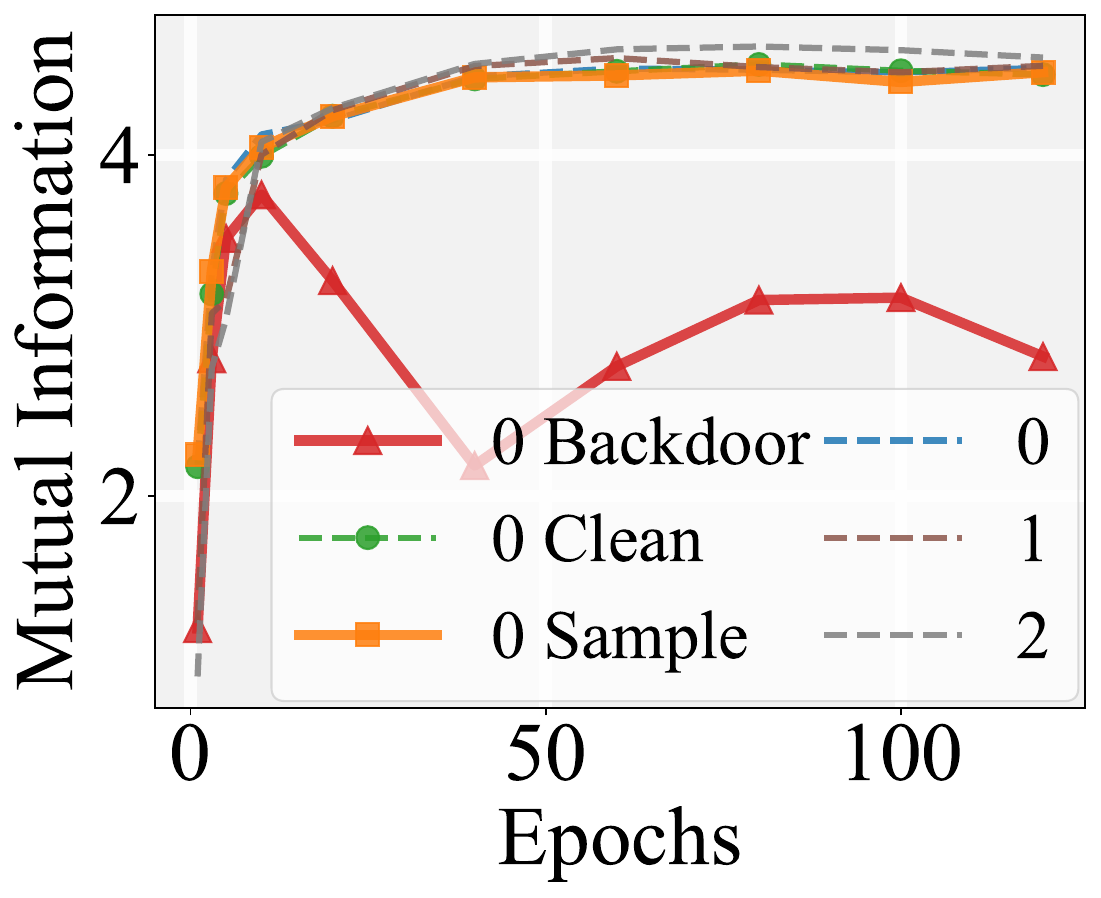} &
        \includegraphics[width=0.2\linewidth]{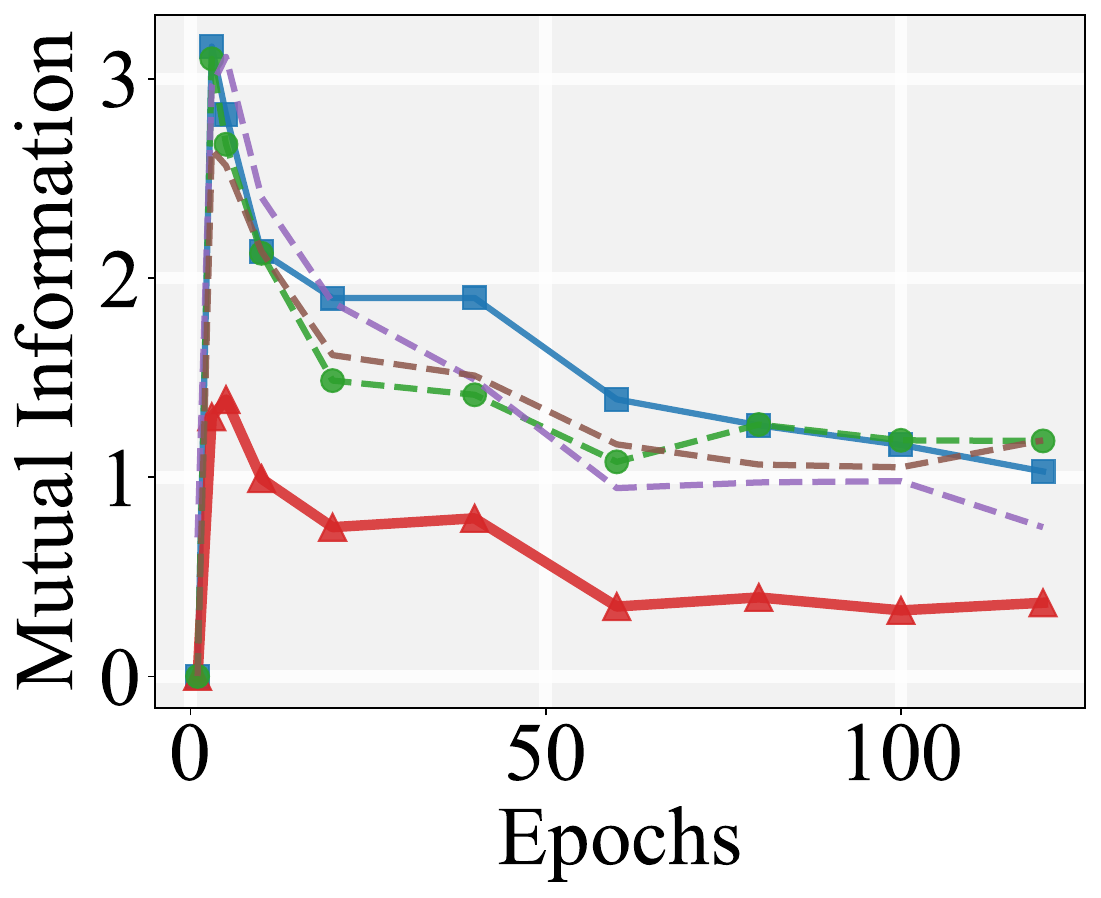} &
        \includegraphics[width=0.2\linewidth]{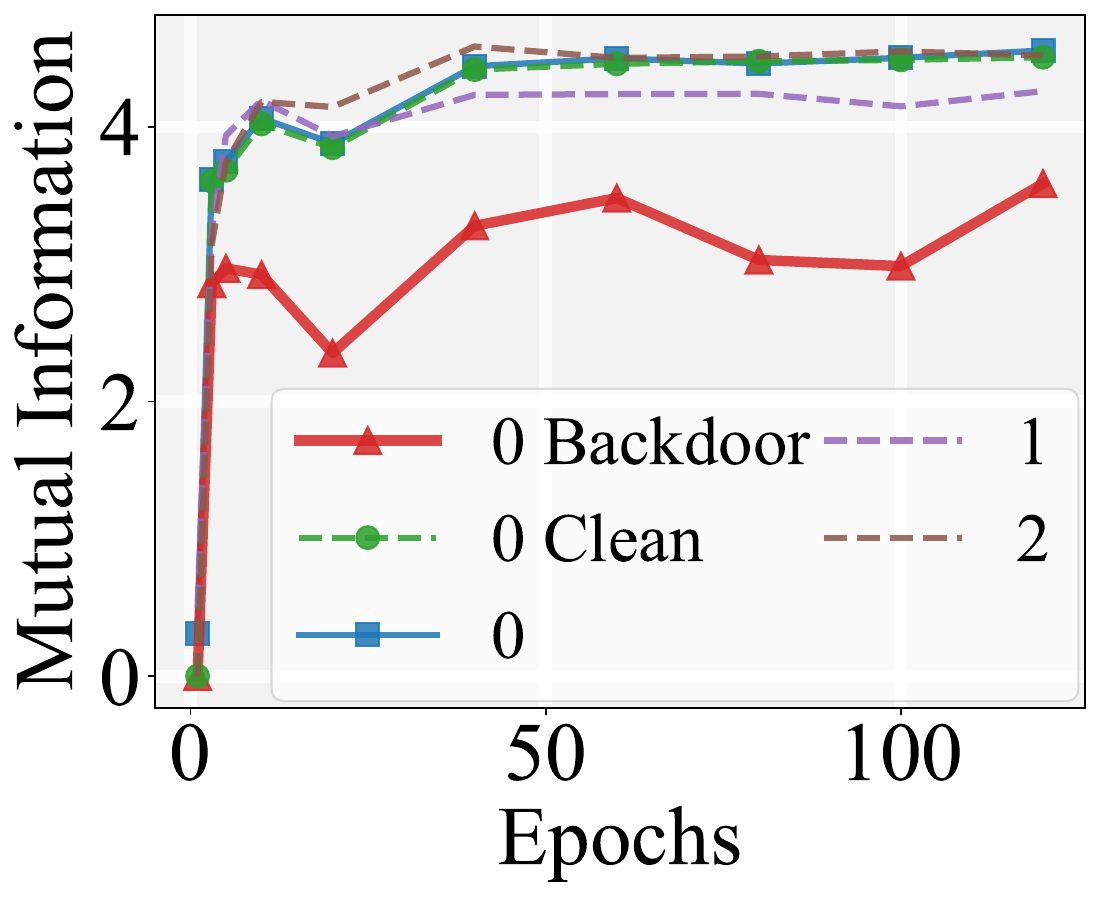} \\
    \end{tabular}
    \caption{\footnotesize{Comparison of MI dynamics under different poisoning ratios. The rows represent the attack methods, and the columns represent the MI terms $I(X;T)$ and $I(T;Y_{\text{pred}})$.}}
    \label{fig:ratio_comparison_mi}
\end{figure*}

In this section, we analyze the MI dynamics of all four backdoor attacks under reduced poisoning ratios: 1\% for dirty-label attacks and 10\% (relative to the target class) for clean-label attacks. The results are visualized in Figure~\ref{fig:ratio_comparison_mi}, revealing several notable changes in the dynamic properties compared to higher poisoning ratios.

First, for all attacks, the overall \(I(X;T)\) of class 0 converges to values nearly indistinguishable from other clean classes. This is due to the reduced presence of backdoor samples in the target class under lower poisoning ratios, diminishing the overall influence of backdoor samples on the class 0 representation.
Second, attacks with stealthier triggers, such as Blend and WaNet, exhibit a rapid compression of \(I(X;T)\) for backdoor samples to near-zero levels. This is because the distributed and subtle nature of these triggers weakens their ability to contribute meaningful information to the model’s representations, particularly under lower poisoning ratios. The reduced frequency of backdoor samples exacerbates this effect, as the model is less likely to retain weak trigger signals during the compression phase.
Third, for attacks with more salient triggers, such as BadNets and LC, \(I(T;Y_{\text{pred}})\) for backdoor samples shows a marked decline compared to higher poisoning ratios. This behavior reflects the model's reduced capacity to effectively learn the original semantic features of backdoor samples, which are sourced from multiple classes. With fewer backdoor samples, the model struggles to establish a strong association between these features and the target class, leading to lower prediction confidence and decreased \(I(T;Y_{\text{pred}})\).

\subsection{Evaluation of Stealth Score}
\label{sec:score_analysis}
We evaluate the proposed stealth score across various attacks and datasets, with results summarized in Table~\ref{tab:score}.

A counterintuitive yet significant finding is that BadNets, often regarded as a simple and conspicuous backdoor attack,
generally achieves the lowest stealth score (or one close to the lowest), indicating high \textit{model-level} stealthiness.
This phenomenon occurs because the fixed, high-salience trigger in BadNets is rapidly and efficiently learned, becoming seamlessly integrated into the model's feature representation. This high-quality integration minimizes the differences in MI dynamics between clean and poisoned samples. In contrast, attacks like Blend or WaNet rely on more diffuse triggers that alter the feature space more substantially, leading to higher MI discrepancies (e.g., lower $I(T;Y_{pred})$ for backdoor samples), thus resulting in worse (higher) stealth scores.While the LC attack shows a slightly lower score in some settings, direct comparisons are complicated by its different poisoning ratio.

\begin{table}[htbp]
    \caption{\footnotesize{Comparison of Stealth Score on Different datasets and Poisoning Ratios. $\Delta_{I(X;T)}$ and $\Delta_{I(T;Y_\text{pred})}$ represent stealth score calculated by MI metric $I(X;T)$ and $I(T;Y_\text{pred})$ respectively. Score represents the average of $\Delta_{I(X;T)}$ and $\Delta_{I(T;Y_\text{pred})}$ ($\downarrow$ represents the smaller the better). The best result in each setting and metric is marked in \textbf{bold}. The simplest attack BadNets generally achieves the lowest stealth score.}}
    \centering
    \begin{tabularx}{\linewidth}{lXXX} 
        \toprule [2pt]
        \textbf{Attack} & \textbf{$\Delta_{I(X;T)}$} & \textbf{$\Delta_{I(T;Y_\text{pred})}$} & \textbf{Score} $\downarrow$ \\ 
        \midrule [2pt]

        \multicolumn{4}{c}{\textbf{CIFAR-10 (10\% poison)}} \\ \midrule
        BadNets & 0.3903  & \textbf{0.1736}  & \textbf{0.2820}  \\
        Blend  & 0.3920  & 0.4956  & 0.4438 \\
        WaNet  & \textbf{0.3263}  & 0.4727  & 0.3995 \\
        LC-25\% & 0.5930  & 0.4632  & 0.5281  \\
        Adap-Blend & 0.5333 & 0.4150 & 0.4742 \\
        Ftrojan & 0.4163 & 0.6833 & 0.5498\\
        \midrule 
        
        \multicolumn{4}{c}{\textbf{CIFAR-10 (1\% poison)}} \\ \midrule
        BadNets & 0.4667  & 0.493  & 0.4799  \\
        Blend  & 0.7111  & \textbf{0.3695}  & 0.5403  \\
        WaNet  & 0.5179  & 0.5567  & 0.5373  \\
        LC-10\% & \textbf{0.4178}  & 0.5117  & \textbf{0.4648}  \\
        \midrule 

        \multicolumn{4}{c}{\textbf{SVHN (10\% poison)}} \\ \midrule
        BadNets & 0.5946  & \textbf{0.4690}  & 0.5318  \\
        Blend  & 0.5173  & 0.7135  & 0.6154  \\
        WaNet  & \textbf{0.4883}  & 0.6767  & 0.5825  \\
        LC-25\% & 0.4926  & 0.4927  & \textbf{0.4926}  \\
        
        \bottomrule
    \end{tabularx}
    \label{tab:score}
\end{table}

The analysis in Table~\ref{tab:score} also reveals the score's sensitivity to experimental 
conditions. With a relatively higher poisoning ratio (10\%), the model more easily learns the 
trigger and integrates it, leading to generally lower (better) stealth scores. Conversely, at 
lower poisoning ratios (1\%), the model struggles to capture the trigger consistently, resulting in 
increased MI differences. This trend is particularly evident for attacks like WaNet and Blend, 
where a lower poisoning ratio weakens the trigger's signal strength, leading to a higher (worse) stealth score. 
Furthermore, dataset characteristics influence stealthiness. On SVHN, the higher background complexity makes the model less sensitive to subtle perturbations (Blend, WaNet), resulting in higher MI 
discrepancies compared to CIFAR-10. In contrast, BadNets maintains its low stealth score 
across datasets, as its robust, localized trigger is consistently integrated. The impacts and 
implications of this score are further discussed in Section~\ref{sec:conclusion}.

\subsection{Analysis: Model-Level Stealth vs. Perceptual Quality}
\label{sec:score_vs_metrics}
A key implication of our findings in Section~\ref{sec:score_analysis} is the discrepancy between 
traditional, human-centric notions of stealth and the model-level integration quantified by our 
MI-based score. To empirically validate this trade-off, we compare our Stealth Score against 
three standard perceptual quality metrics (LPIPS, PSNR, and SSIM) that measure visual fidelity.

\subsubsection{Perceptual Metric Definitions}
We first briefly define the perceptual metrics used. 
The Learned Perceptual Image Patch Similarity (LPIPS) score~\cite{ghazanfari2023r} 
measures the distance between two images ($y, y_0$) by computing the weighted $L_2$ distance 
between their deep feature activations extracted from a pre-trained network (e.g., VGG) across $L$ layers:
\[ d(y, y_0) = \sum_l \frac{1}{H_l W_l} \sum_{h,w} w_l \cdot || \hat{y}_{hw}^l - \hat{y}_{0,hw}^l ||_2^2 \]
where $\hat{y}^l, \hat{y}_0^l$ are the normalized activations in layer $l$, and $w_l$ is the layer weight.

The Peak Signal-to-Noise Ratio (PSNR) quantifies reconstruction quality based 
on Mean Squared Error (MSE), where a higher value indicates smaller pixel-wise deviation:
\[ PSNR = 10 \cdot \log_{10}\left(\frac{MAX_I^2}{MSE}\right) \]

The Structural Similarity Index Measure (SSIM)~\cite{wang2004image} evaluates 
similarity by comparing local patterns of luminance, contrast, and structure:
\[ SSIM(x,y) = [l(x,y)]^\alpha \cdot [c(x,y)]^\beta \cdot [s(x,y)]^\gamma \]

\subsubsection{Comparative Analysis}
\begin{figure*}[!htbp]
    \centering
    \subfloat[LPIPS vs. Stealth Score]{\includegraphics[width=0.33\textwidth]{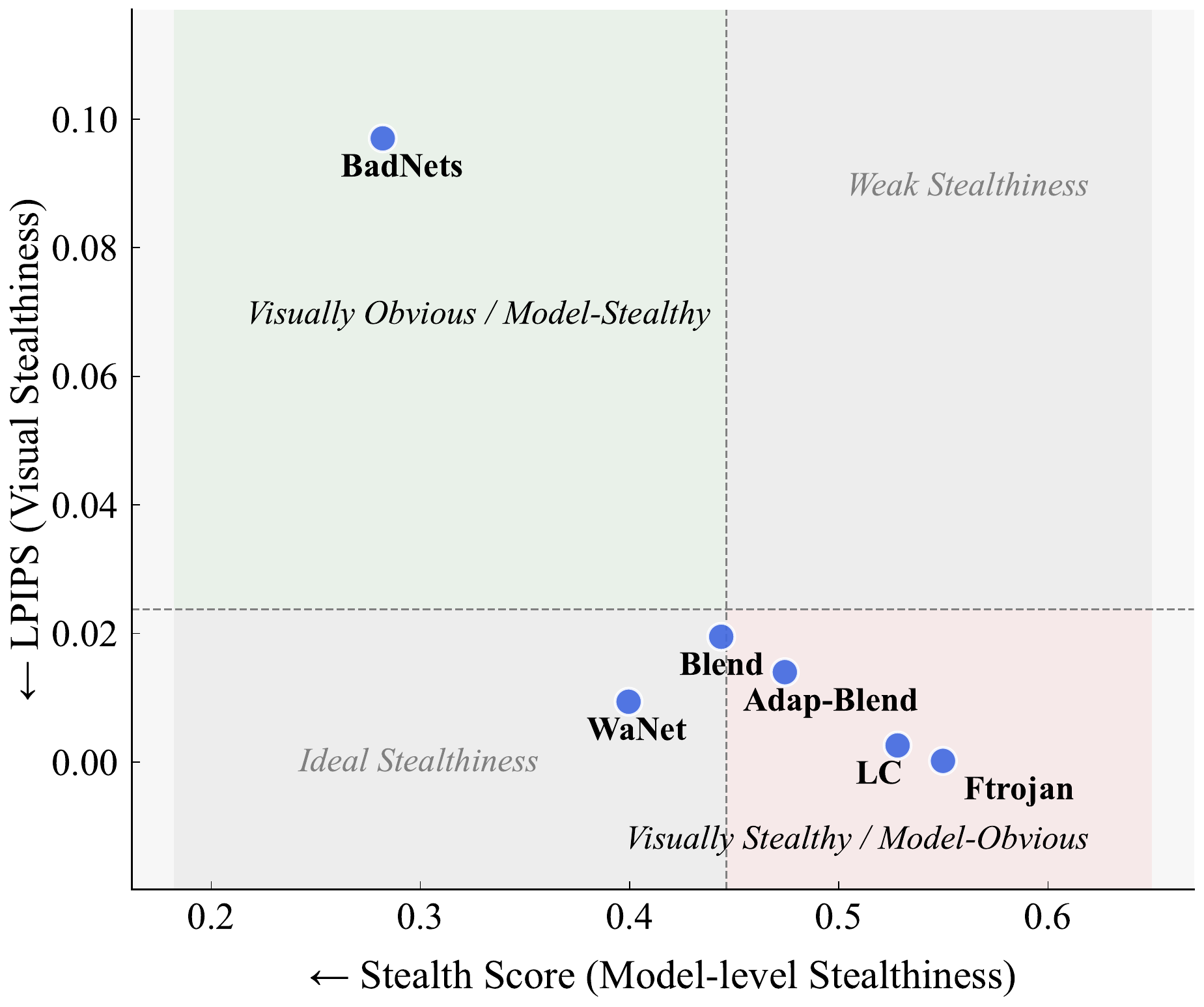}
    \label{fig:lpips_sub}}
    \subfloat[PSNR vs. Stealth Score]{\includegraphics[width=0.33\textwidth]{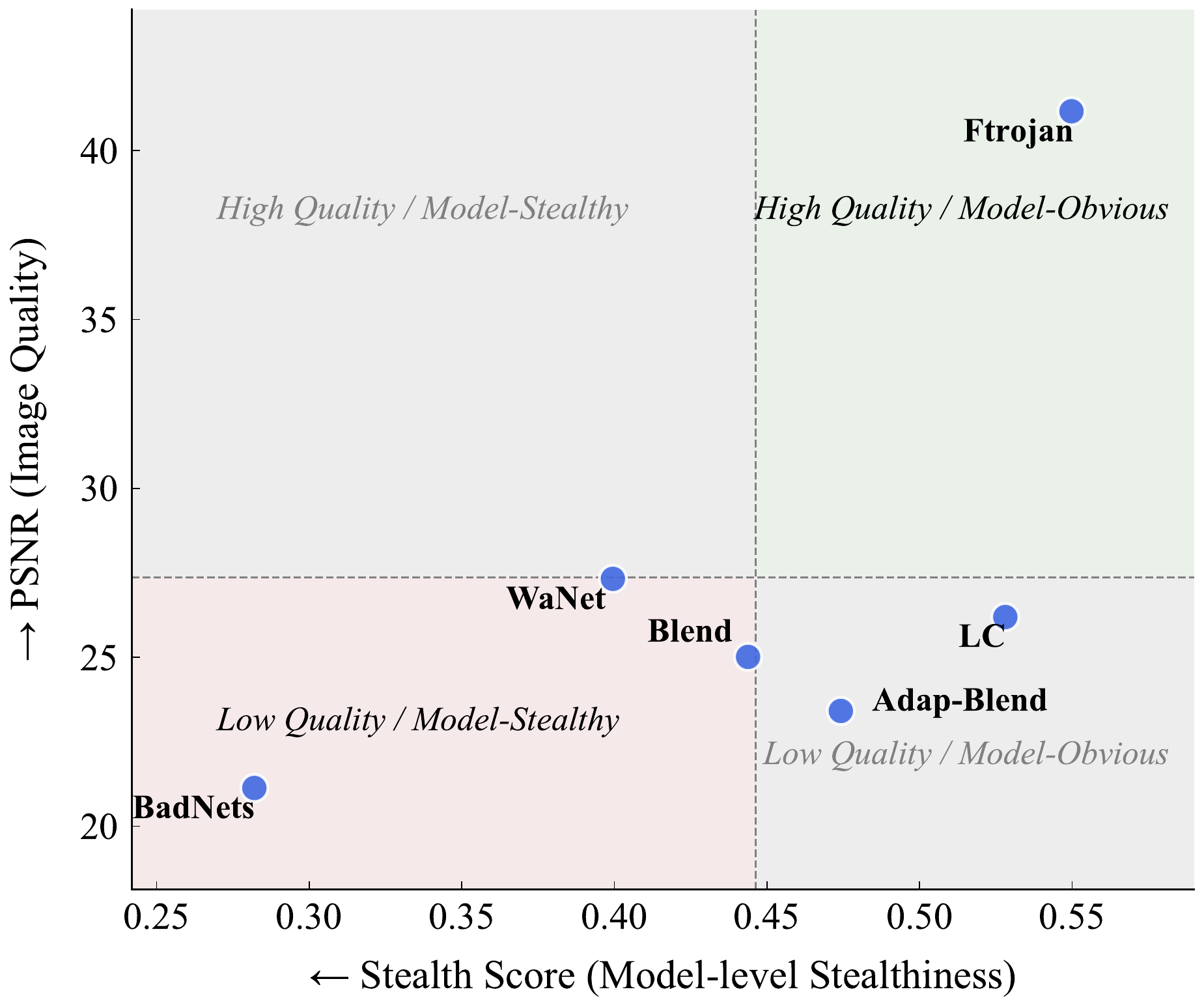}
    \label{fig:psnr_sub}}
    \subfloat[SSIM vs. Stealth Score]{\includegraphics[width=0.33\textwidth]{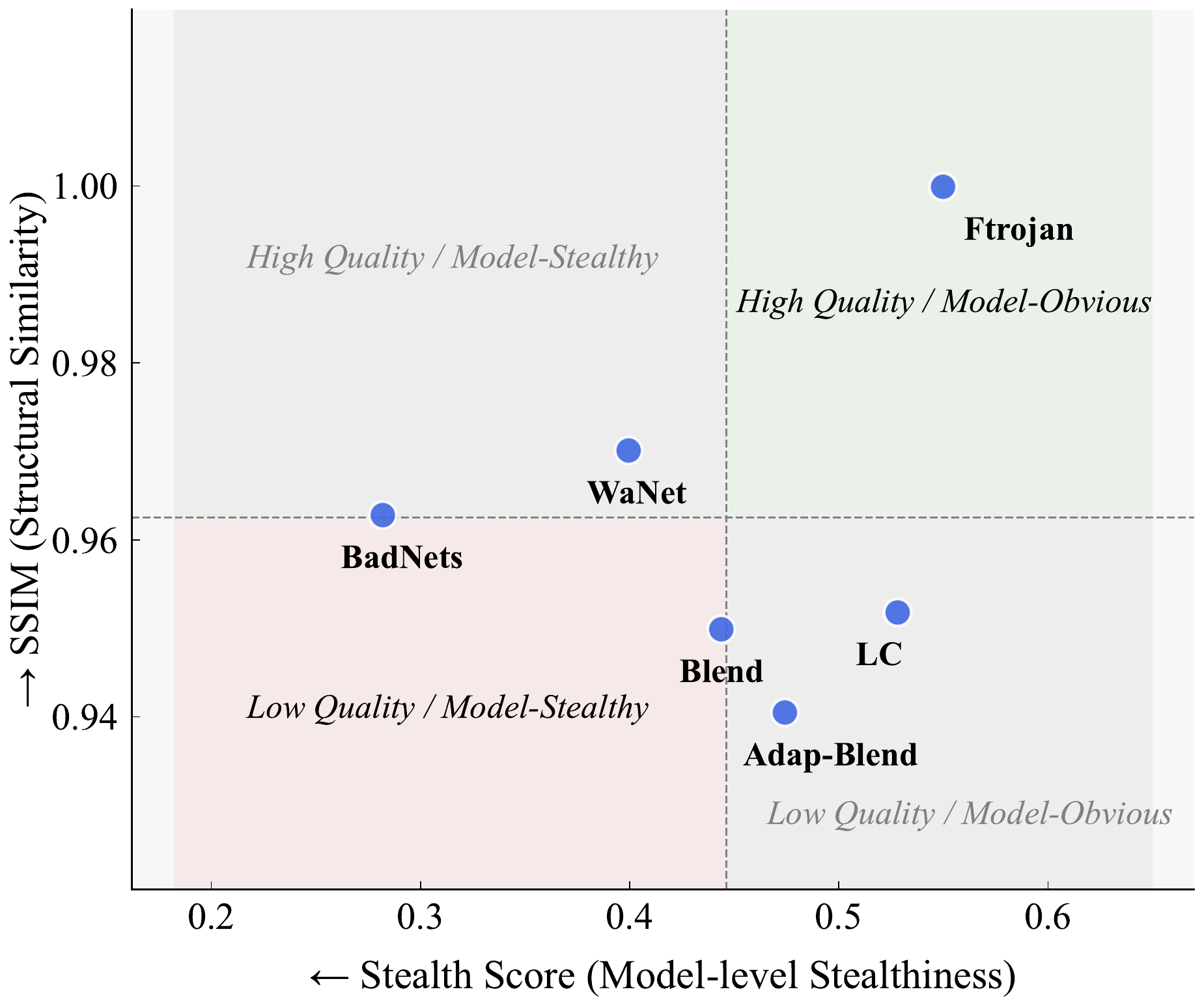}
    \label{fig:ssim_sub}}
    \caption{\footnotesize{Relationship between Stealth Score and traditional IQA metrics (PSNR and SSIM). 
    Higher PSNR and SSIM (better visual quality) do not correlate with better 
    model-level stealthiness (lower score).}}
    \label{fig:combined_metrics}
\end{figure*}

The results, plotted in Figure~\ref{fig:combined_metrics}, 
reveal a clear trade-off between visual stealth and model-integrated stealth. We observe 
no positive correlation; in fact, the attacks cluster into two distinct categories. 
On one hand, attacks like BadNets are visually obvious, exhibiting 
high LPIPS scores, low PSNR, and low SSIM. However, they consistently 
achieve low (good) MI-based Stealth Scores, confirming they are well-integrated 
into the model despite being easy for a human to see. On the other hand, 
attacks like Ftrojan are visually stealthy, with low LPIPS scores, 
high PSNR ($>40$), and near-perfect SSIM (1.00), making them perceptually invisible. 
Despite this, they register the highest (worst) MI-based Stealth Scores, indicating 
their dynamics diverge significantly from clean data and are thus highly 
conspicuous at the model level.
Other attacks like WaNet, Blend, and LC fall between these extremes but 
corroborate the same trend (e.g., WaNet is more visually stealthy than BadNets but 
less integrated at the model level).

In summary, the analyses against LPIPS, PSNR, and SSIM robustly demonstrate 
the value of our proposed Stealth Score. While traditional IQA metrics effectively 
measure visual fidelity, they are unable to quantify how backdoor triggers are 
processed and integrated within a neural network's internal representations. 
Our information-theoretic metric successfully addresses this gap, providing a 
new and essential dimension for evaluating the true stealthiness of backdoor attacks.

\section{Conclusion and Limitation}
\label{sec:conclusion}
In this paper, we presented an analytical framework leveraging the IB principle to investigate the training dynamics of backdoor attacks. By tracking MI evolution, we uncovered that backdoor samples exhibit unique trajectories—characterized by early-phase acceleration and distinct representation clustering—that fundamentally differ from clean data.

Crucially, our findings highlight the orthogonality between perceptual (visual) stealth and model-level (information-theoretic) stealth. Counterintuitively, visually conspicuous attacks like BadNets can achieve superior model stealthiness by integrating seamlessly into internal representations. This demonstrates that traditional metrics, such as trigger perceptibility or Attack Success Rate, are insufficient for characterizing the depth of backdoor embedding, validating our IB-based metric as a vital complementary dimension for evaluation.

Practically, this framework serves as a robust tool for defenders to quantify embedding depth. Real-time monitoring of MI dynamics further offers a pathway for identifying anomalous patterns during training, potentially guiding interventions such as neuron pruning or early stopping.

Our study currently focuses on image classification using standard CNNs (ResNet and VGG). Generalizing these MI dynamics to other modalities (e.g., NLP) and complex architectures (e.g., Transformers or LLMs) remains an open question. Future work will explore the IB framework's applicability to these domains and investigate how information flow analysis can inspire robust defenses against sophisticated attacks.




\appendix
\subsection{MI Dynamics Under Other Datasets}
\label{sec:svhn}

\begin{figure*}[h]
    \centering
    \renewcommand{\arraystretch}{1.5} 
    \setlength{\tabcolsep}{3pt} 
    \begin{tabular}{c c c c} 
        \multicolumn{2}{c}{\colorbox{gray!20}{BadNets-10\%}} &
        \multicolumn{2}{c}{\colorbox{gray!20}{Blend-10\%}} \\ 
        \includegraphics[width=0.2\linewidth]{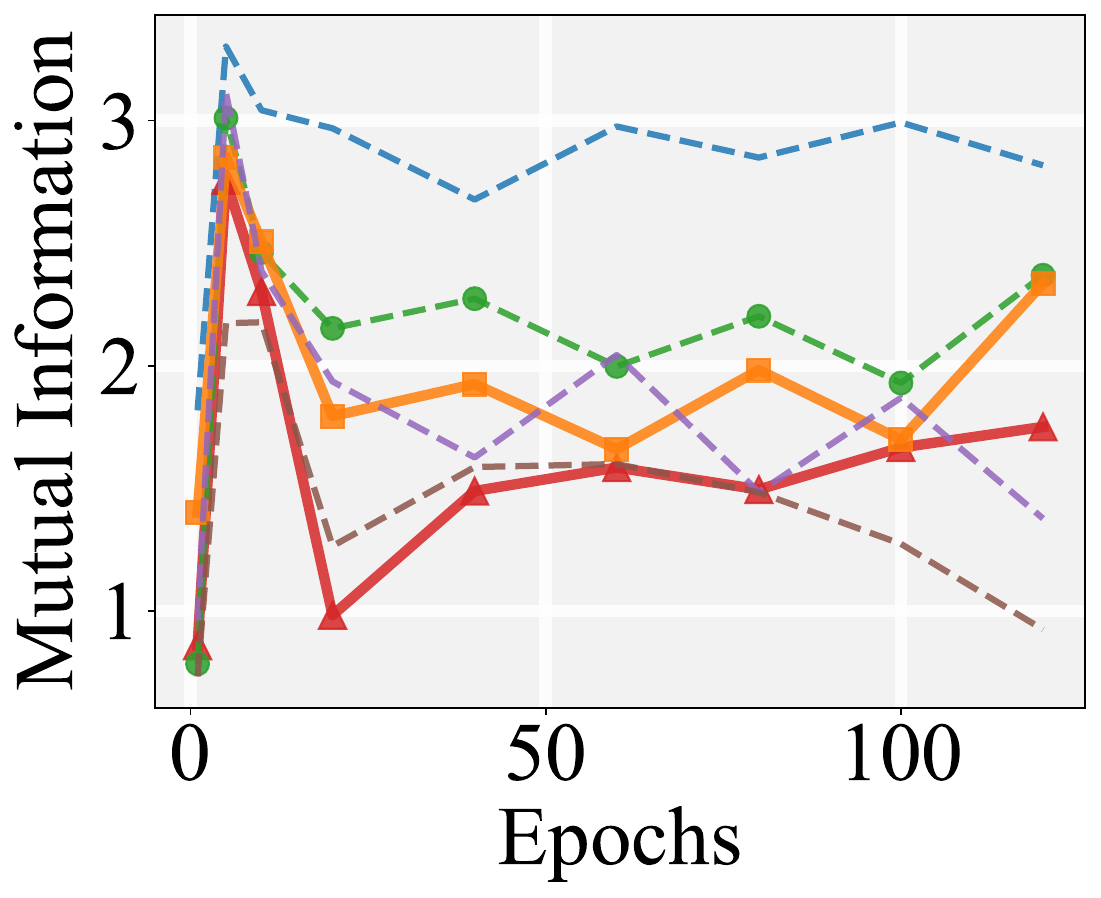} &
        \includegraphics[width=0.2\linewidth]{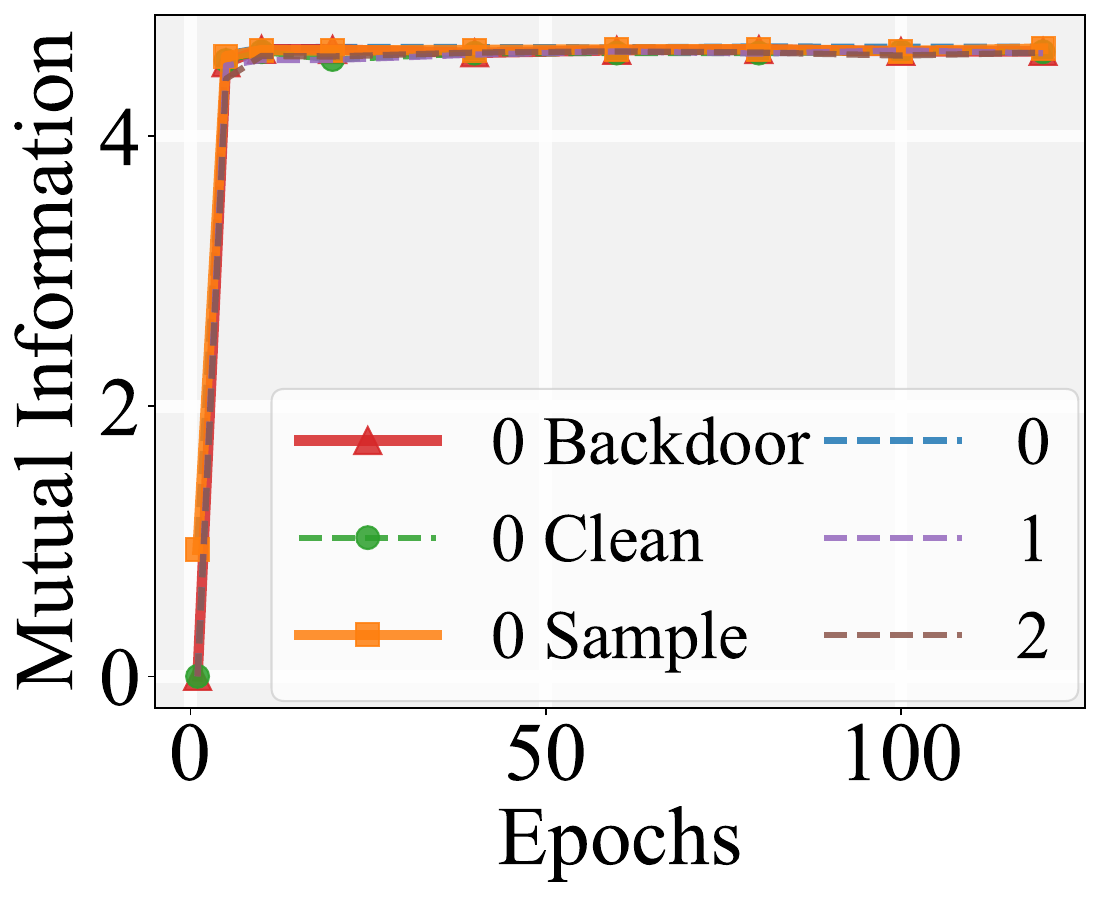} &
        \includegraphics[width=0.2\linewidth]{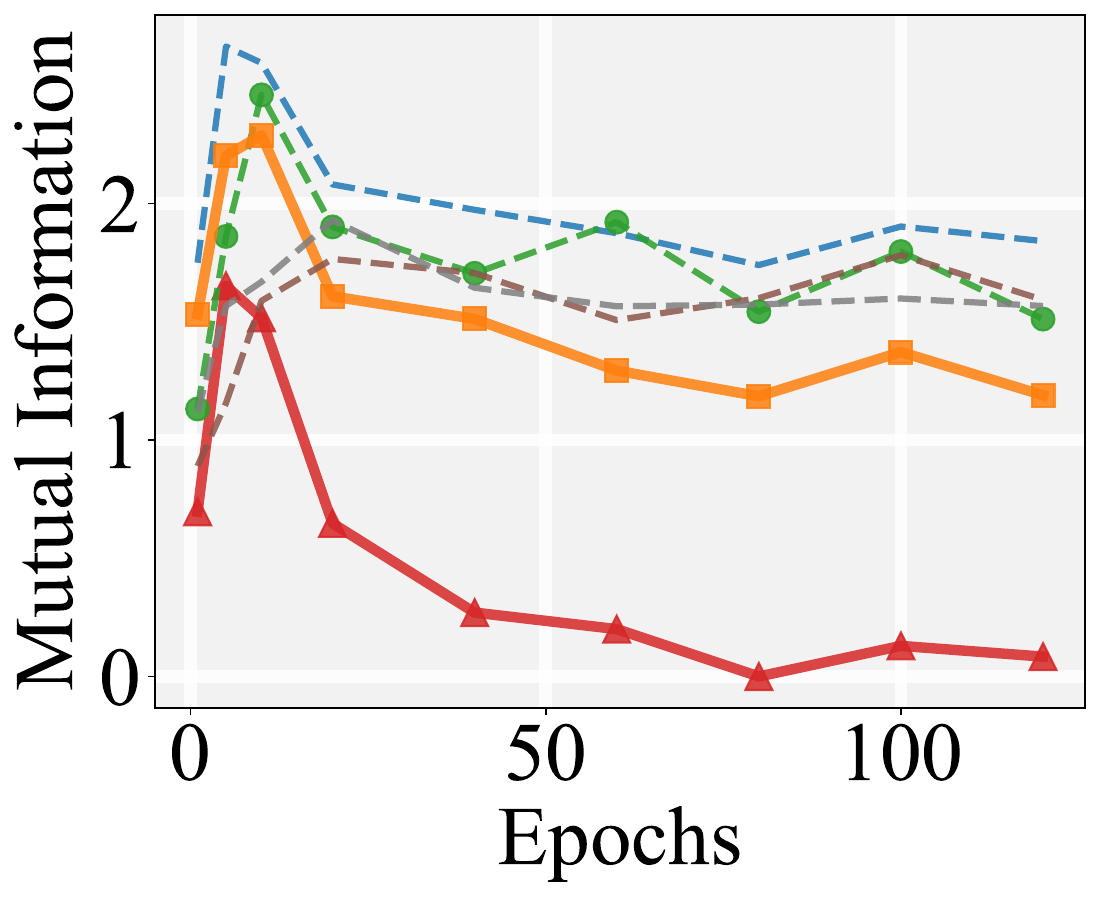} &
        \includegraphics[width=0.2\linewidth]{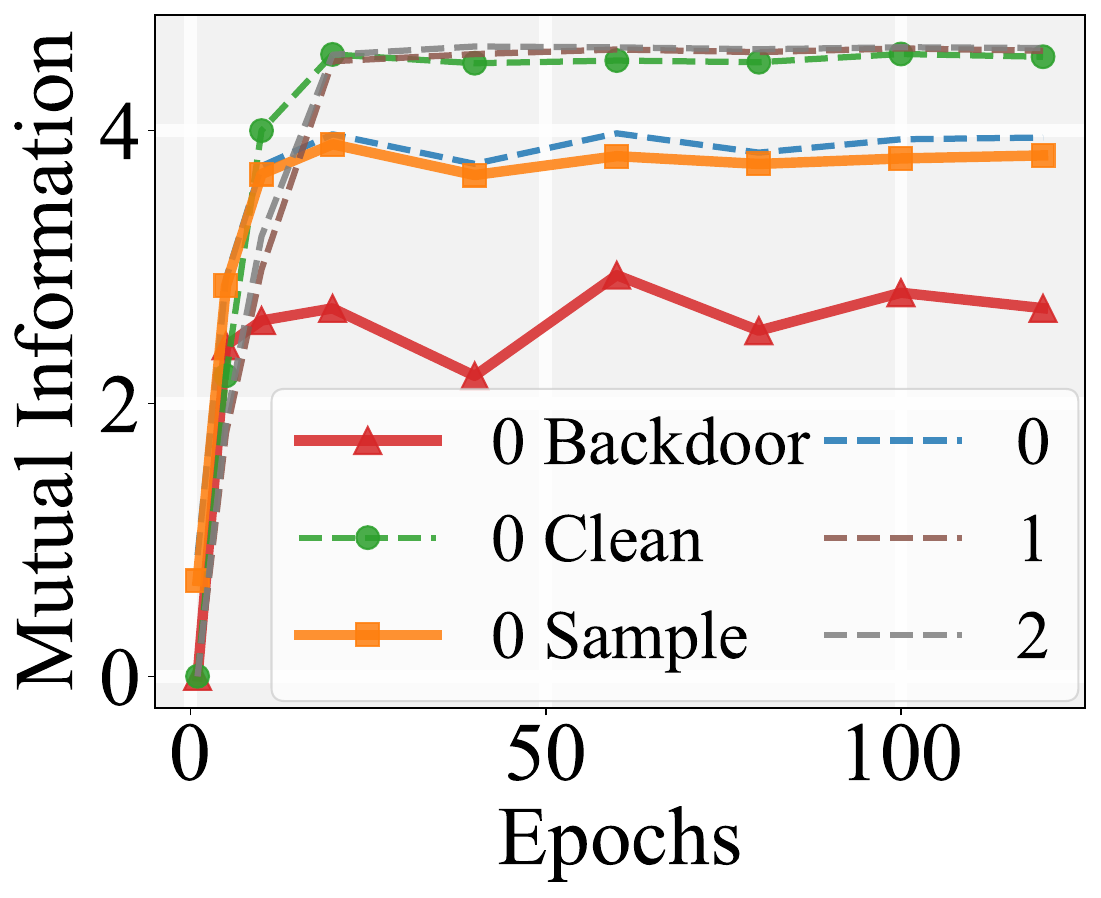} \\

         \multicolumn{2}{c}{\colorbox{gray!20}{WaNet-10\%}} &
        \multicolumn{2}{c}{\colorbox{gray!20}{LC-25\%}} \\
        \includegraphics[width=0.2\linewidth]{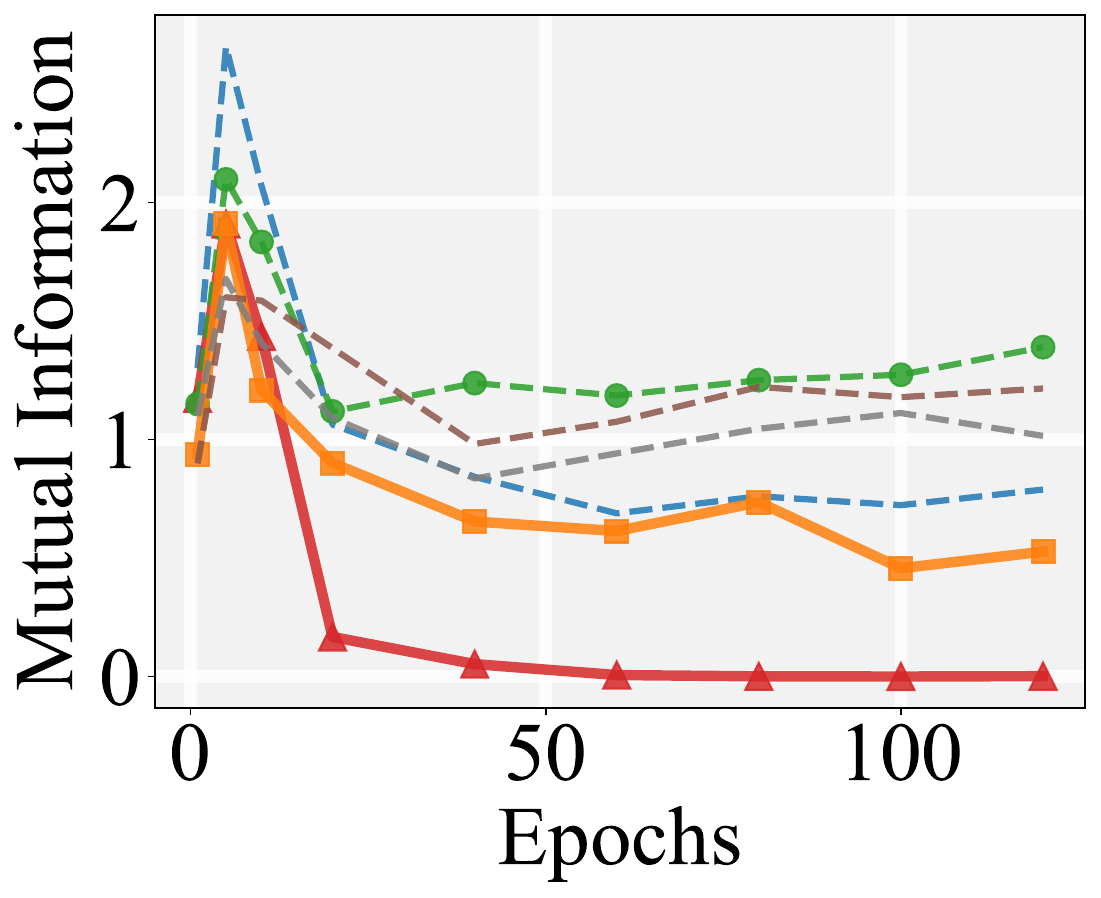} &
        \includegraphics[width=0.2\linewidth]{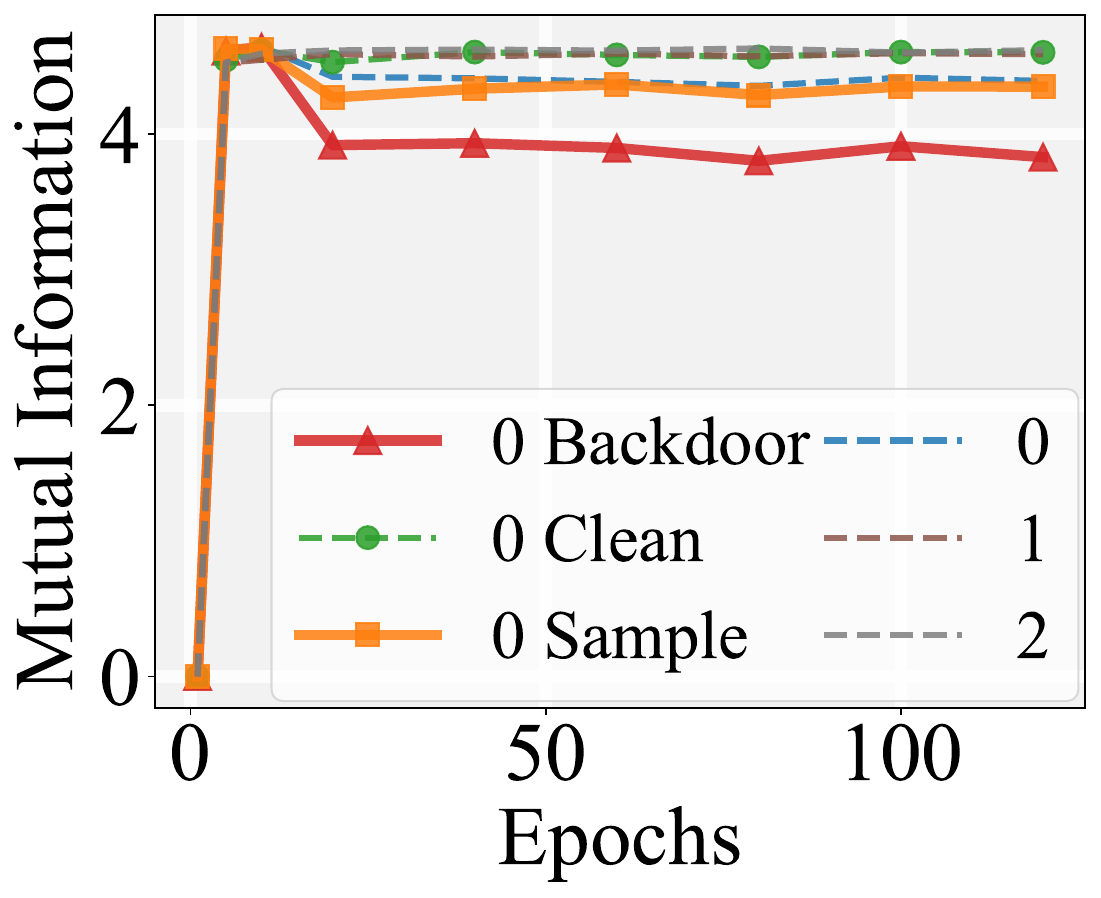} &
        \includegraphics[width=0.2\linewidth]{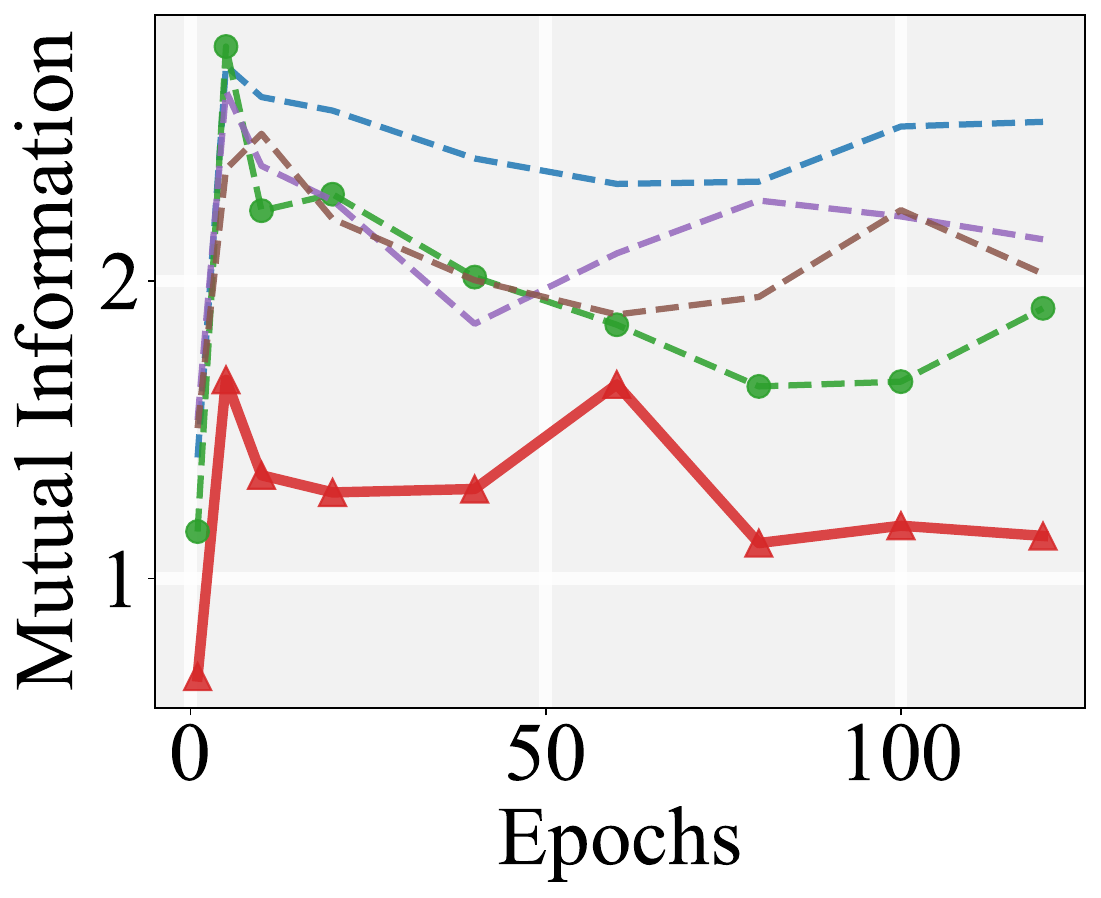} &
        \includegraphics[width=0.2\linewidth]{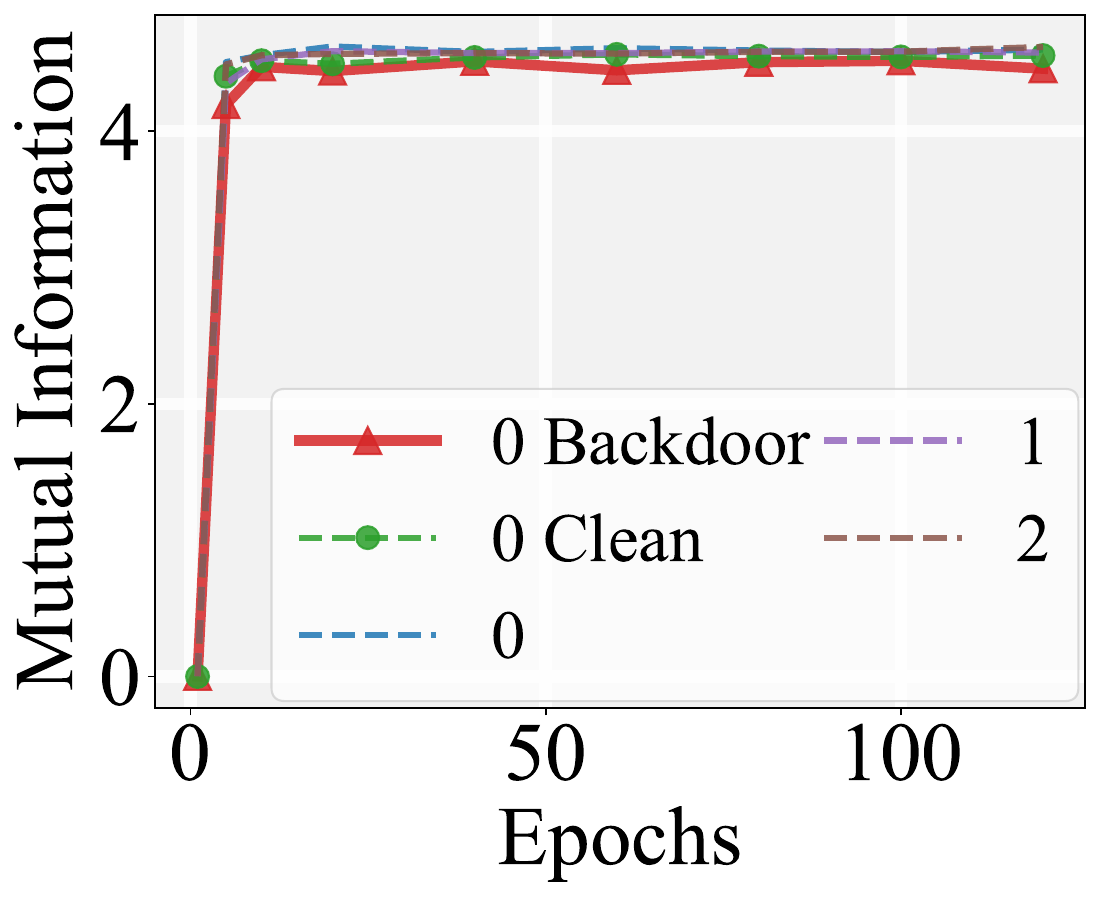} \\
    \end{tabular}
    \caption{\footnotesize{Comparison of MI dynamics under different datasets. The rows represent the attack methods, and the columns represent the MI terms $I(X;T)$ and $I(T;Y_{\text{pred}})$.}}
    \label{fig:svhn_mi}
\end{figure*}

In Figure~\ref{fig:svhn_mi}, we present the MI dynamics of various backdoor attacks (BadNets, WaNet, Blend, and LC) on the SVHN dataset, using the ResNet-18 architecture. The experimental setup for SVHN is consistent with the one used for CIFAR-10, including a learning rate of 0.001. All other settings, including the poisoning ratios (10\% for BadNets, WaNet, and Blend, and 25\% for LC), and the architecture parameters remain unchanged. 

Our results on SVHN mirror those observed on CIFAR-10, with similar trends in MI dynamics across all attack methods. Specifically, the general patterns for \( I(X; T) \) and \( I(T; Y_{\text{pred}}) \) are consistent, demonstrating that the dynamics of backdoor attacks on the ResNet-18 model do not significantly vary between these two datasets. This reinforces the robustness of the observed behaviors and suggests that the influence of backdoor attacks on MI dynamics is dataset-independent when using the same model architecture and attack settings.

\subsection{MI Dynamics on VGG16 Architecture}
\label{sec:vgg16}
\begin{figure*}[h]
    \centering
    \renewcommand{\arraystretch}{1.5} 
    \setlength{\tabcolsep}{3pt} 
    \begin{tabular}{c c c c} 
        \multicolumn{2}{c}{\colorbox{gray!20}{BadNets-10\%}} &
        \multicolumn{2}{c}{\colorbox{gray!20}{Blend-10\%}} \\ 
        \includegraphics[width=0.2\linewidth]{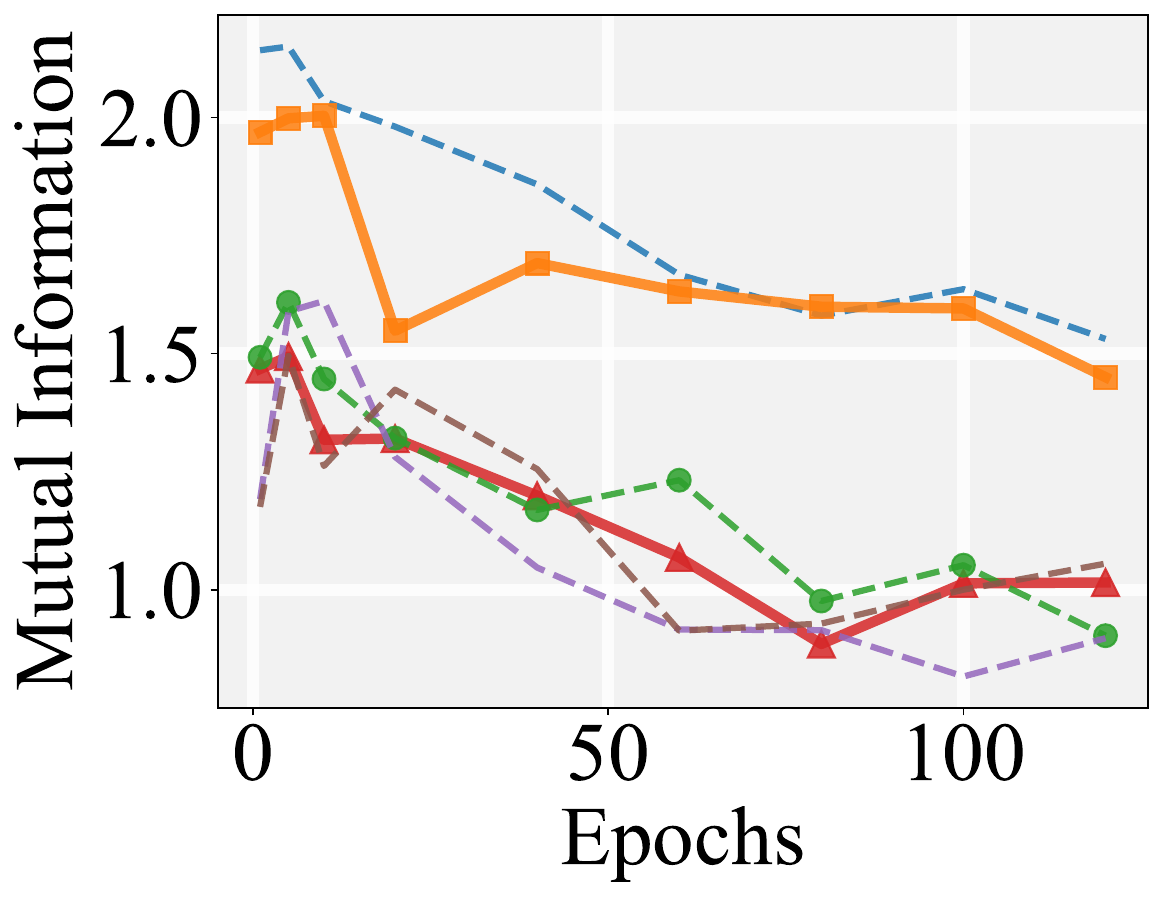} &
        \includegraphics[width=0.2\linewidth]{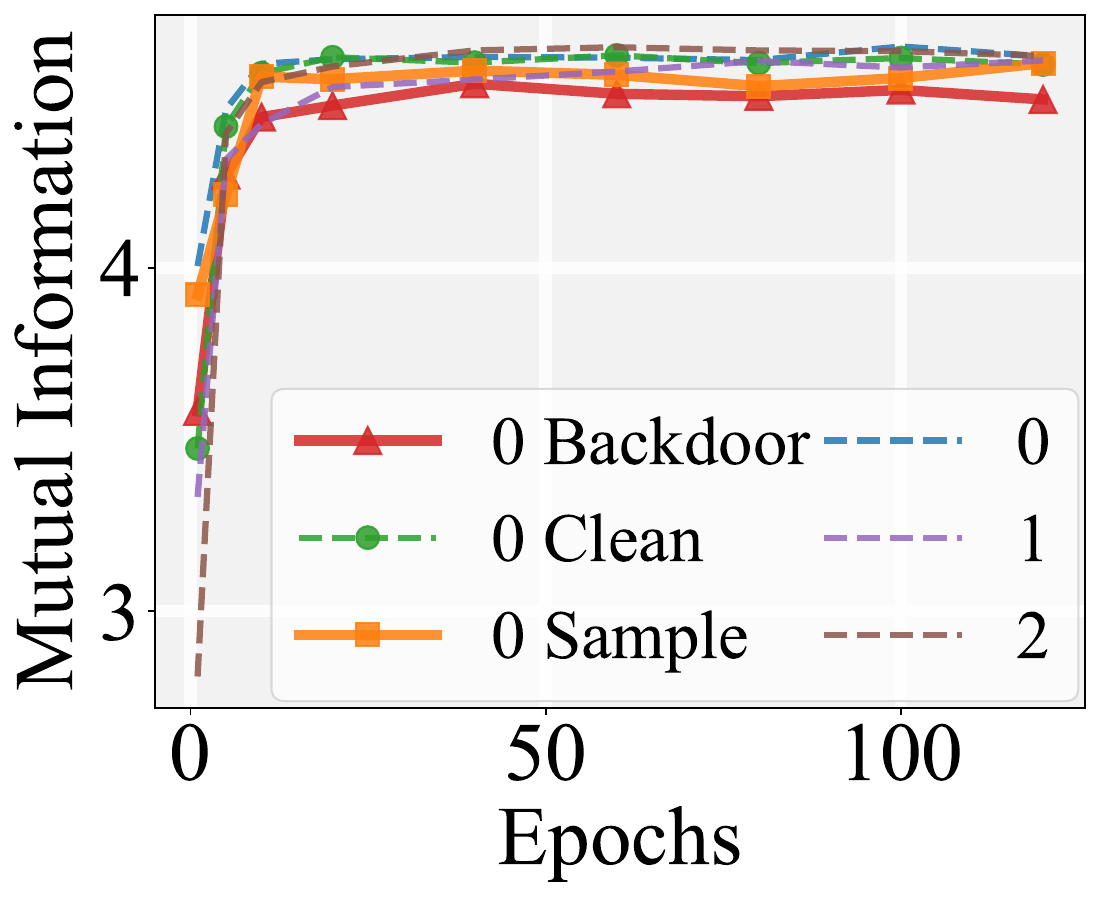} &
        \includegraphics[width=0.2\linewidth]{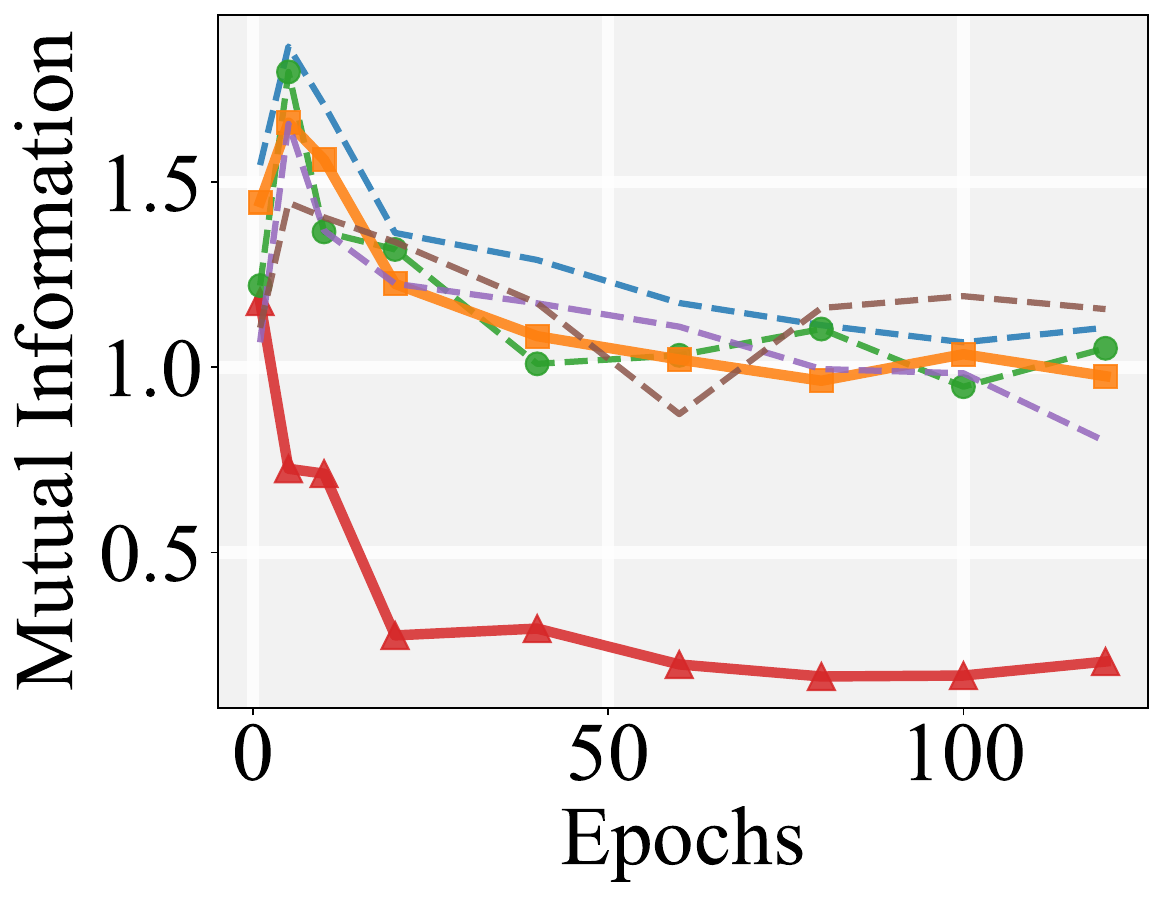} &
        \includegraphics[width=0.2\linewidth]{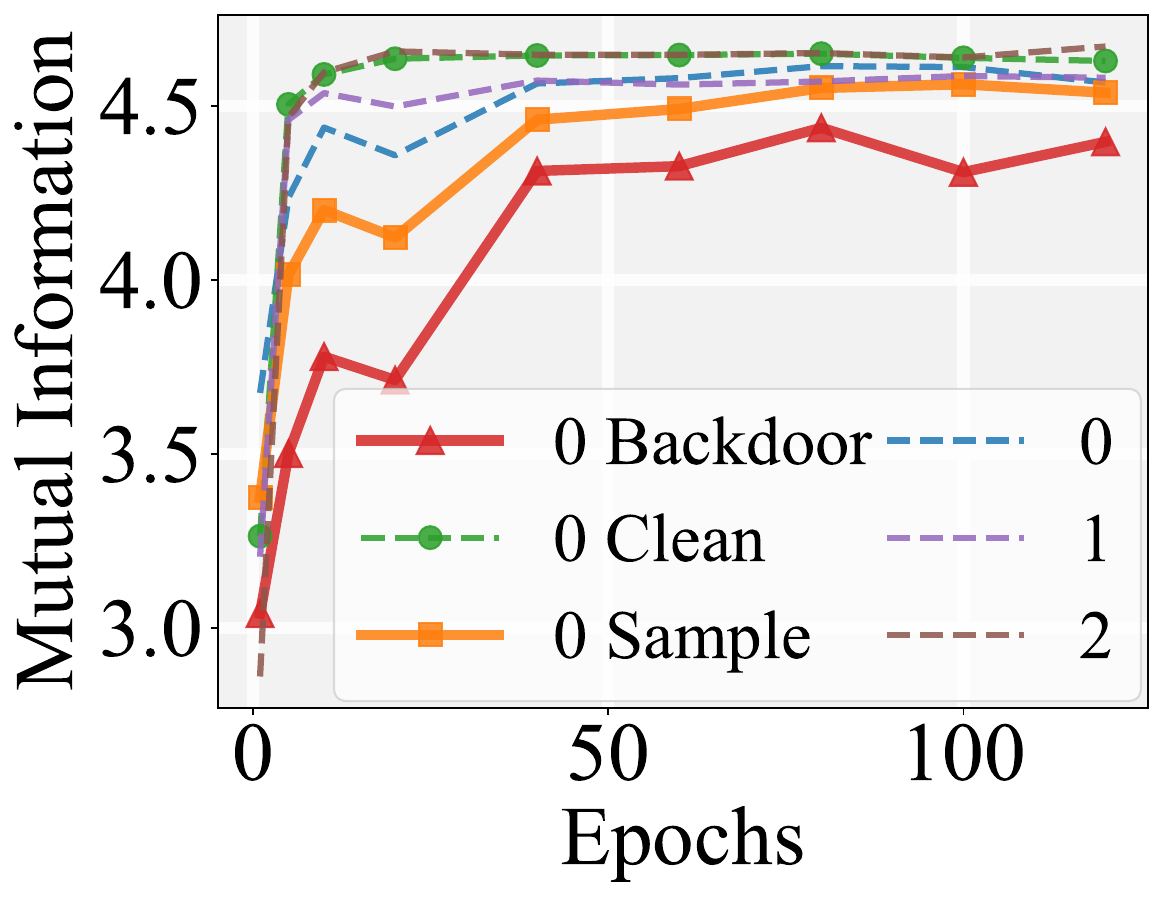} \\

        \multicolumn{2}{c}{\colorbox{gray!20}{WaNet-10\%}} &
        \multicolumn{2}{c}{\colorbox{gray!20}{LC-25\%}} \\
        \includegraphics[width=0.2\linewidth]{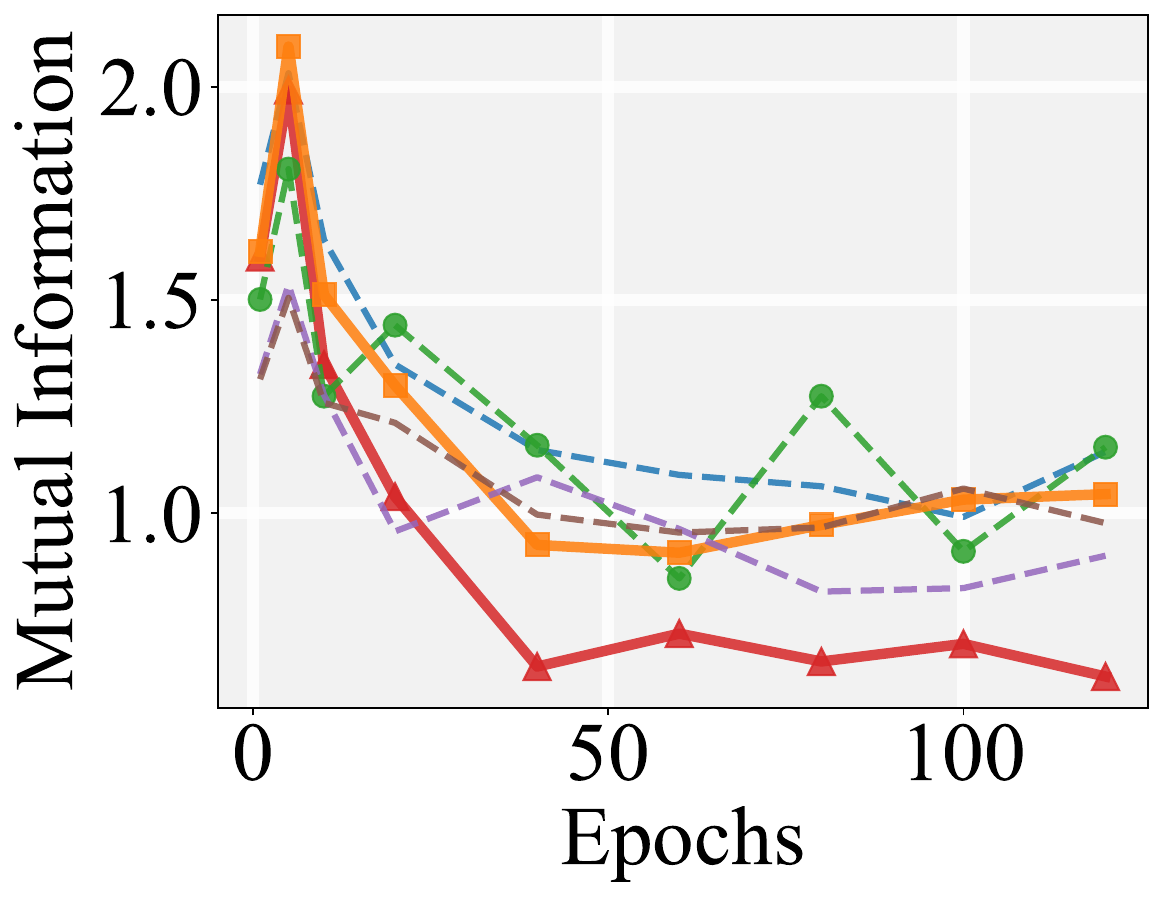} &
        \includegraphics[width=0.2\linewidth]{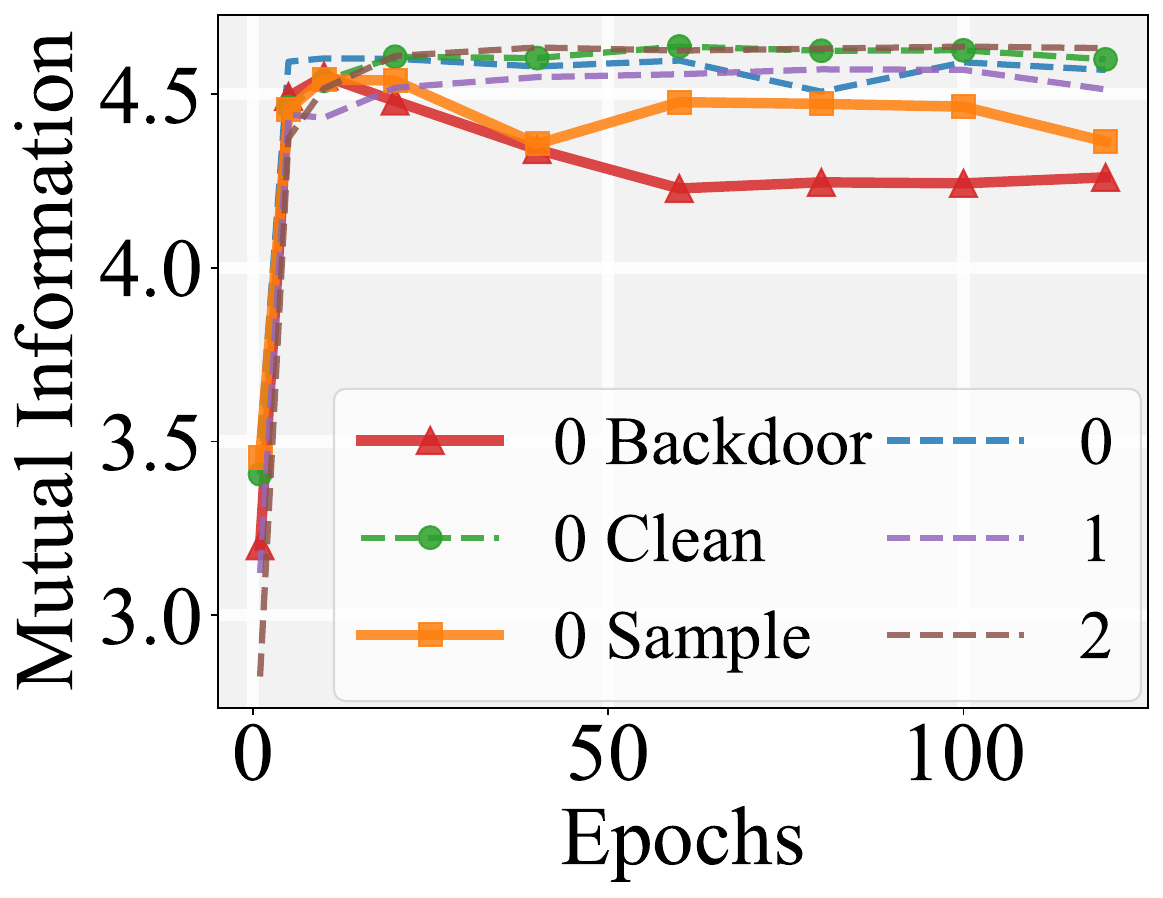} &
        \includegraphics[width=0.2\linewidth]{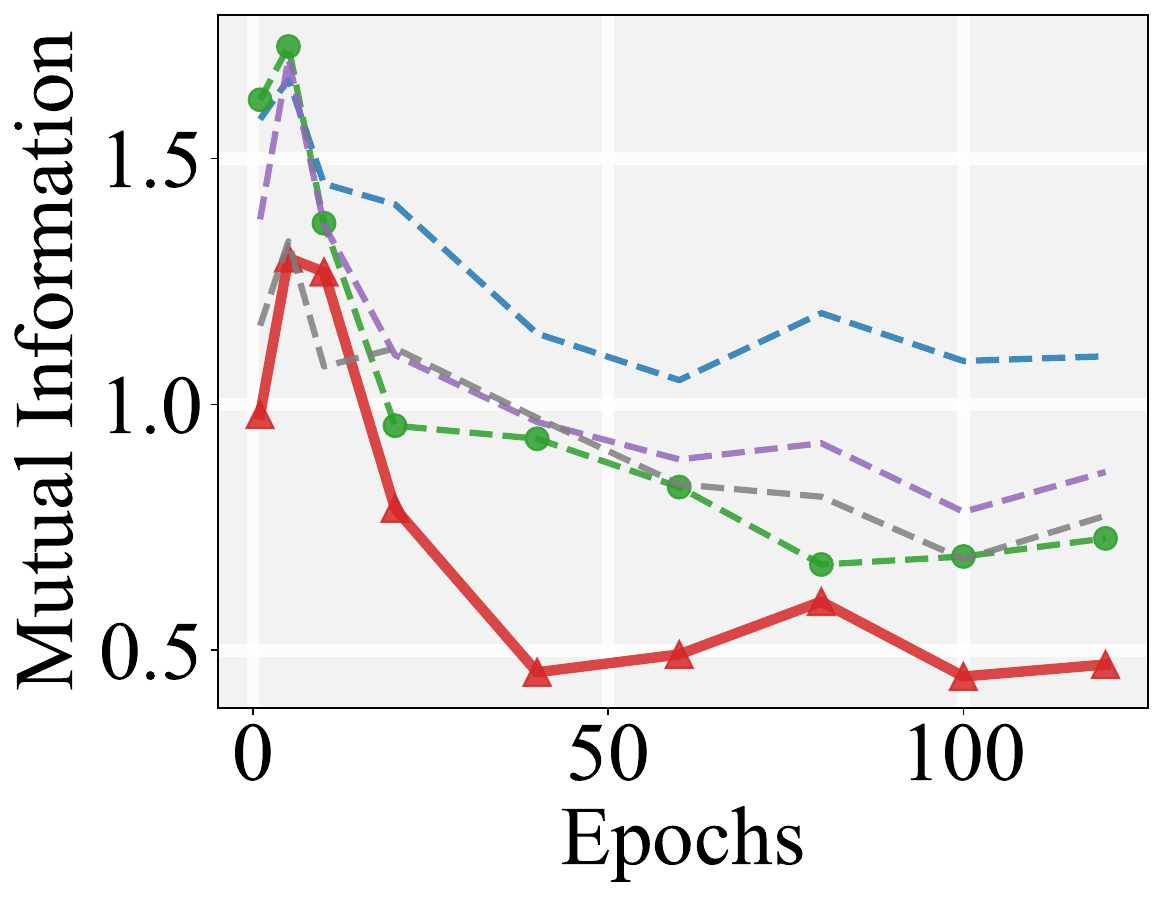} &
        \includegraphics[width=0.2\linewidth]{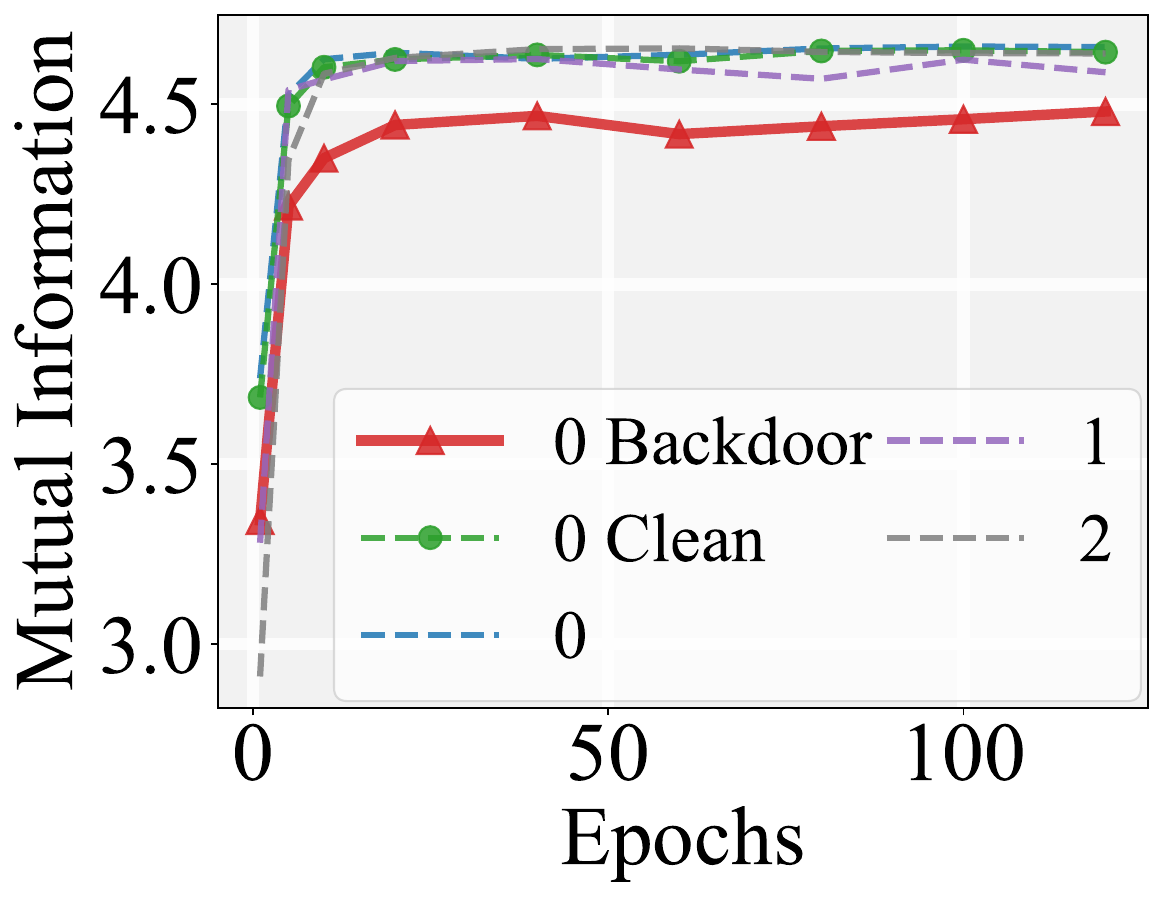} \\
    \end{tabular}
    \caption{\footnotesize{Comparison of MI dynamics using VGG16. The rows represent different attack methods, while the columns display the MI terms \( I(X; T) \) and \( I(T; Y_{\text{pred}}) \).}}
    \label{fig:vgg16_mi}
\end{figure*}

We extend our analysis to the VGG16 architecture using the same attack configurations and noise variance ($\gamma=0.4$) as the ResNet-18 experiments, with the learning rate adapted to 0.001. Figure~\ref{fig:vgg16_mi} illustrates the resulting MI dynamics. 

While the general trends align with ResNet-18, VGG16 exhibits distinct behaviors due to its architectural characteristics. 
For BadNets, the backdoor samples are less distinguishable from clean samples in the representation space ($I(X; T)$) compared to ResNet-18. This suggests that VGG16 is less efficient at disentangling trigger patterns from semantic features, resulting in a more uniform distribution.
Conversely, for Blend and WaNet attacks, VGG16 maintains a higher $I(T; Y_{\text{pred}})$ than expected. Unlike ResNet-18, which effectively isolates the trigger as a predictive shortcut, VGG16 relies more heavily on the remaining semantic features despite the global perturbations. 
Finally, the MI dynamics for the LC attack remain consistent with those observed in ResNet-18.

\subsection{Ablation Study on Noise Variance $\gamma$}
\label{sec:gamma}
\begin{figure*}[h]
    \centering
    \renewcommand{\arraystretch}{1.5} 
    \setlength{\tabcolsep}{3pt} 
    \begin{tabular}{c c c c} 
        \multicolumn{2}{c}{\colorbox{gray!20}{BadNets-10\%}} &
        \multicolumn{2}{c}{\colorbox{gray!20}{Blend-10\%}} \\ 
        \includegraphics[width=0.2\linewidth]{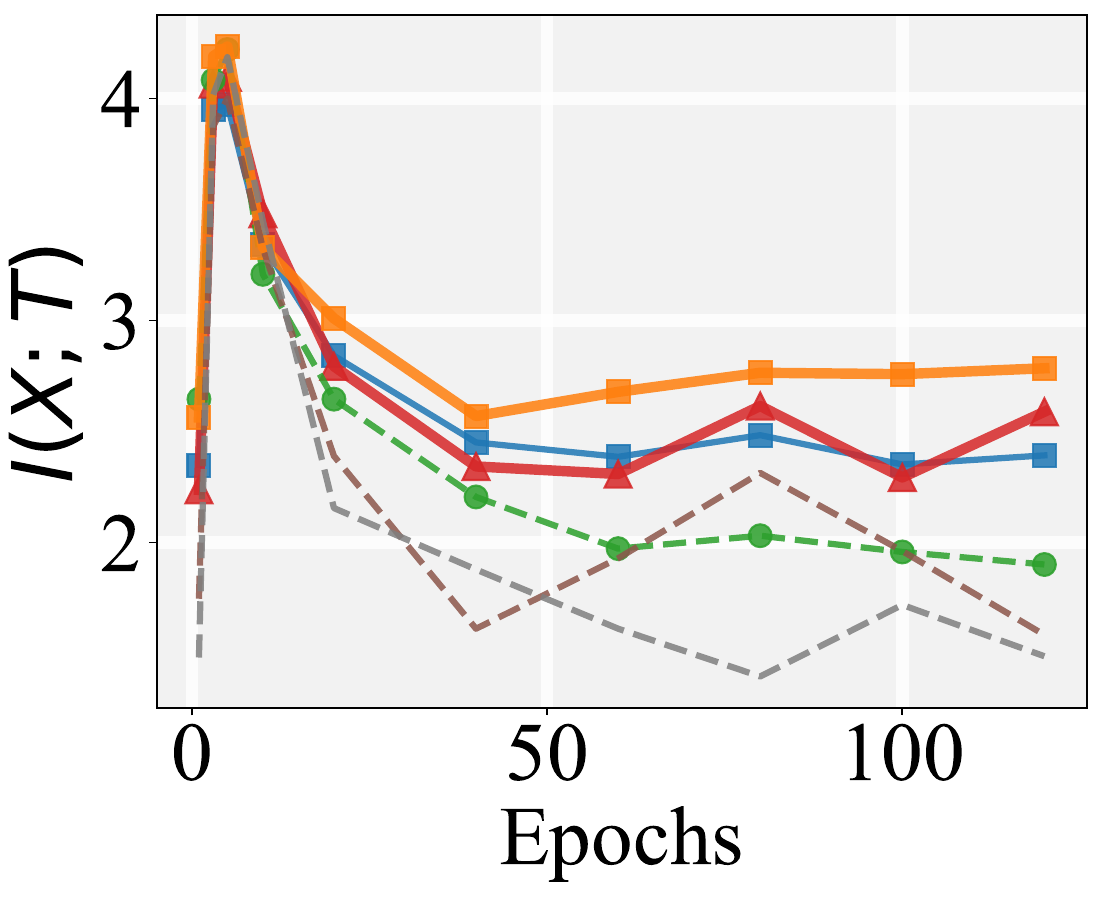} &
        \includegraphics[width=0.2\linewidth]{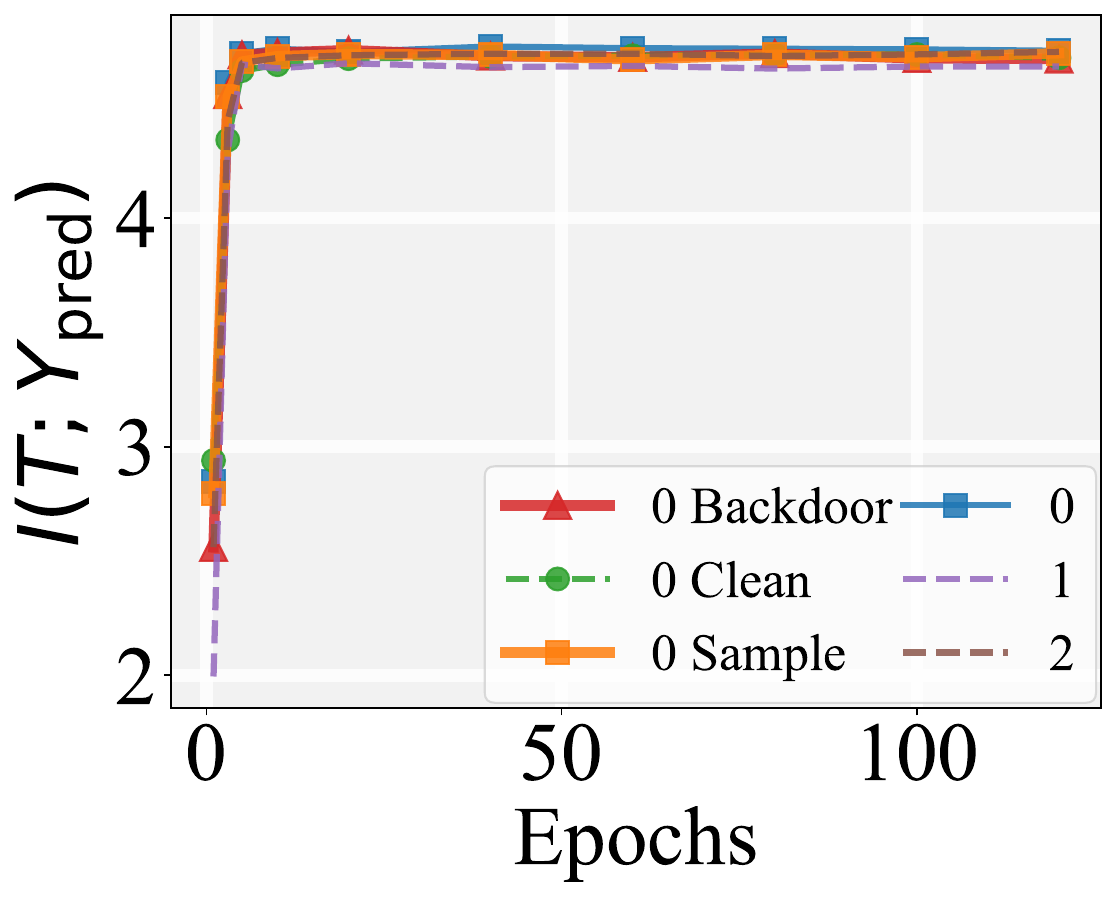} &
        \includegraphics[width=0.2\linewidth]{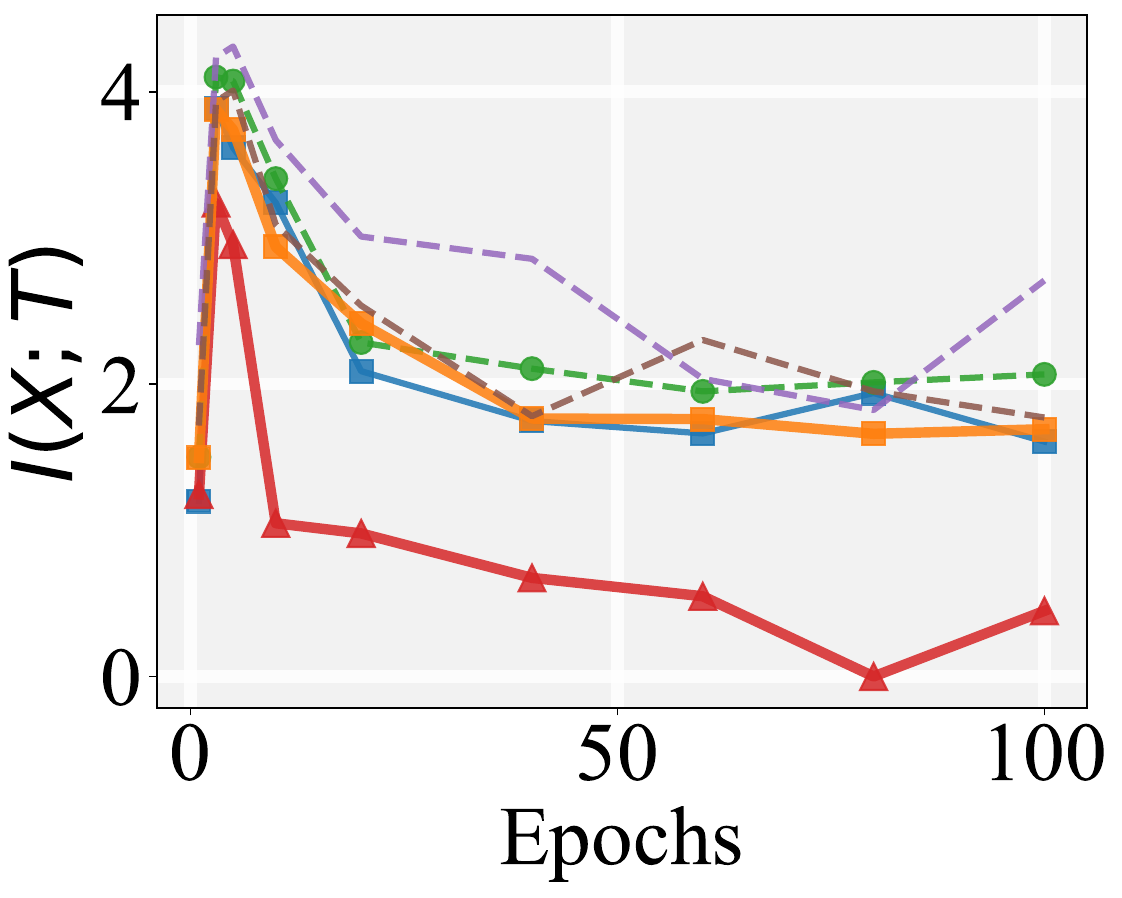} &
        \includegraphics[width=0.2\linewidth]{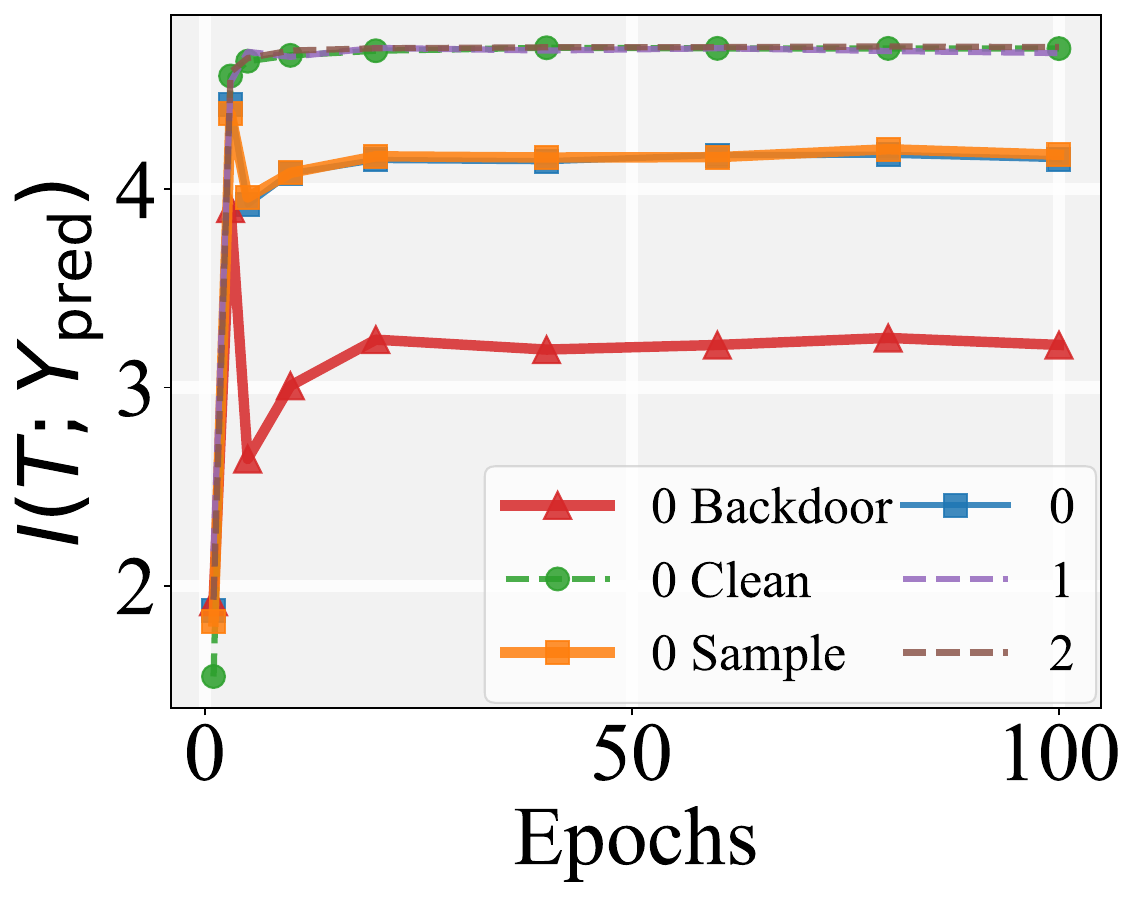} \\

        \multicolumn{2}{c}{\colorbox{gray!20}{WaNet-10\%}} &
        \multicolumn{2}{c}{\colorbox{gray!20}{LC-25\%}} \\
        \includegraphics[width=0.2\linewidth]{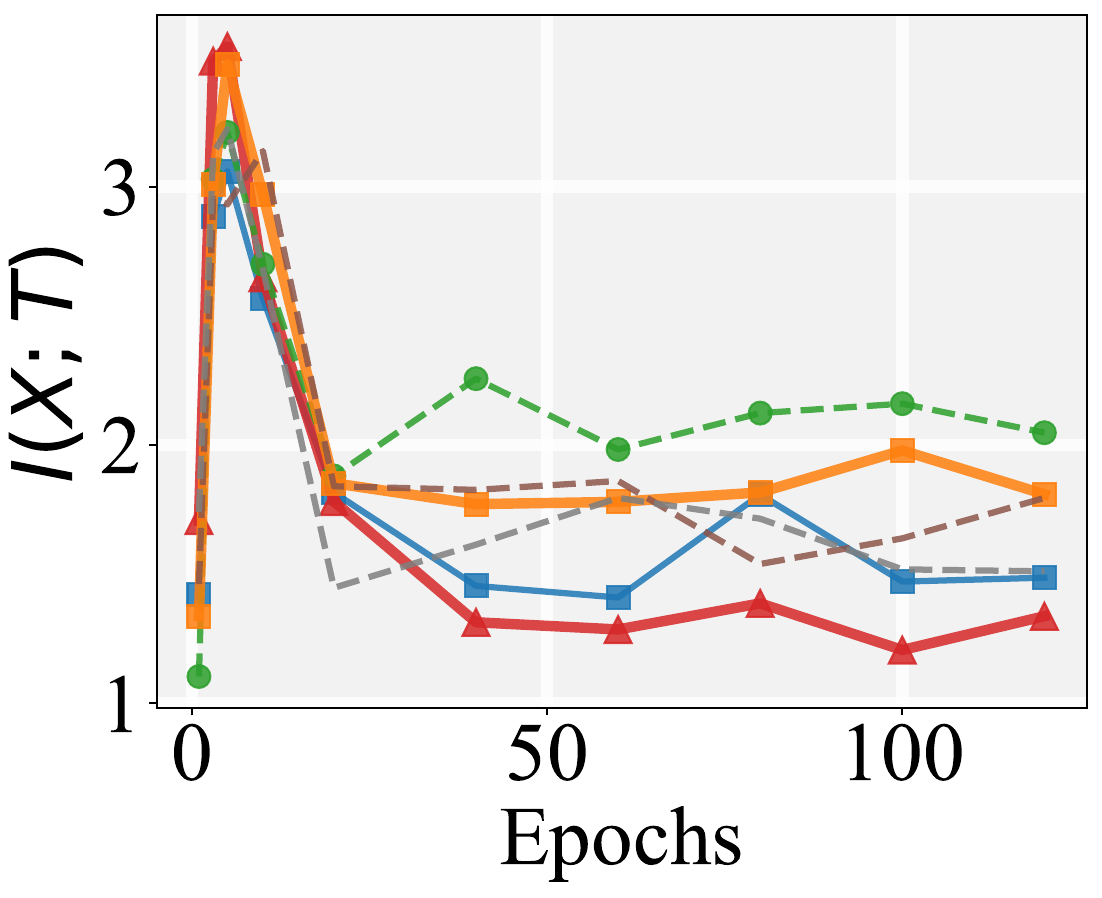} &
        \includegraphics[width=0.2\linewidth]{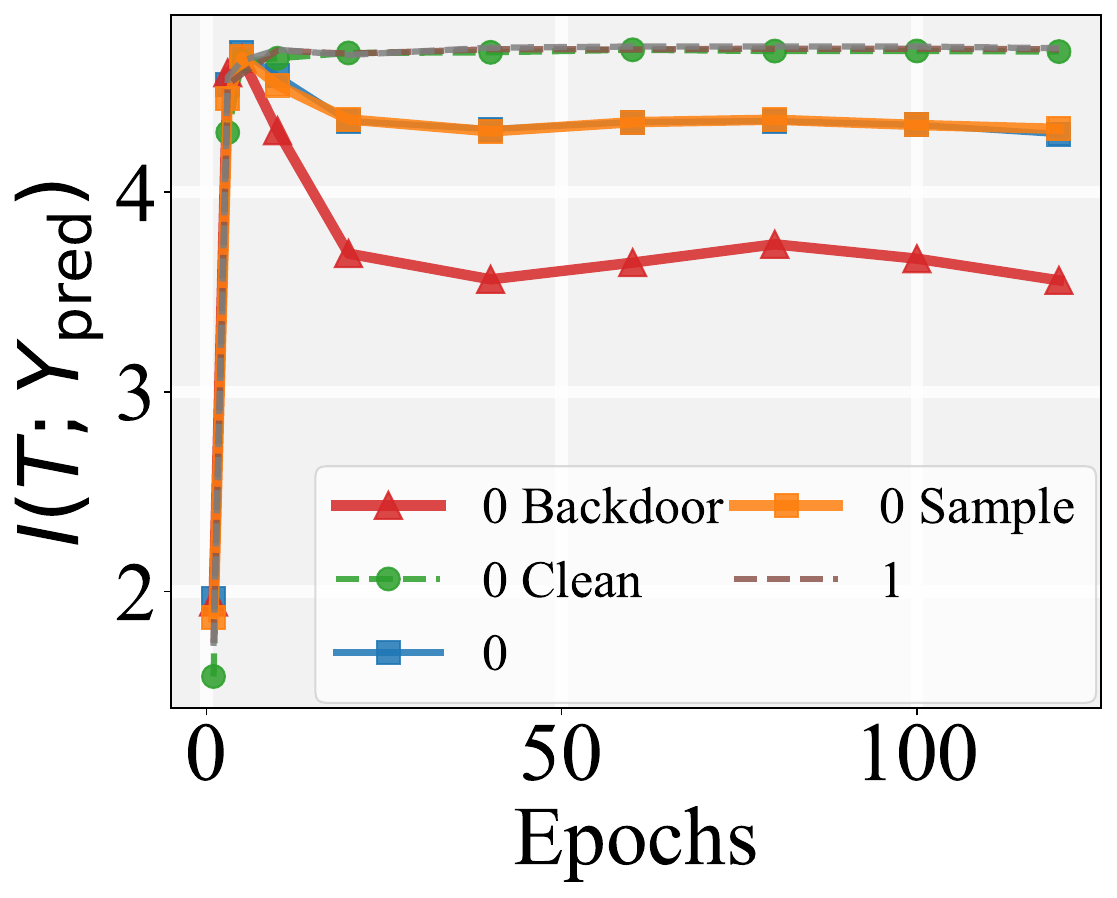} &
        \includegraphics[width=0.2\linewidth]{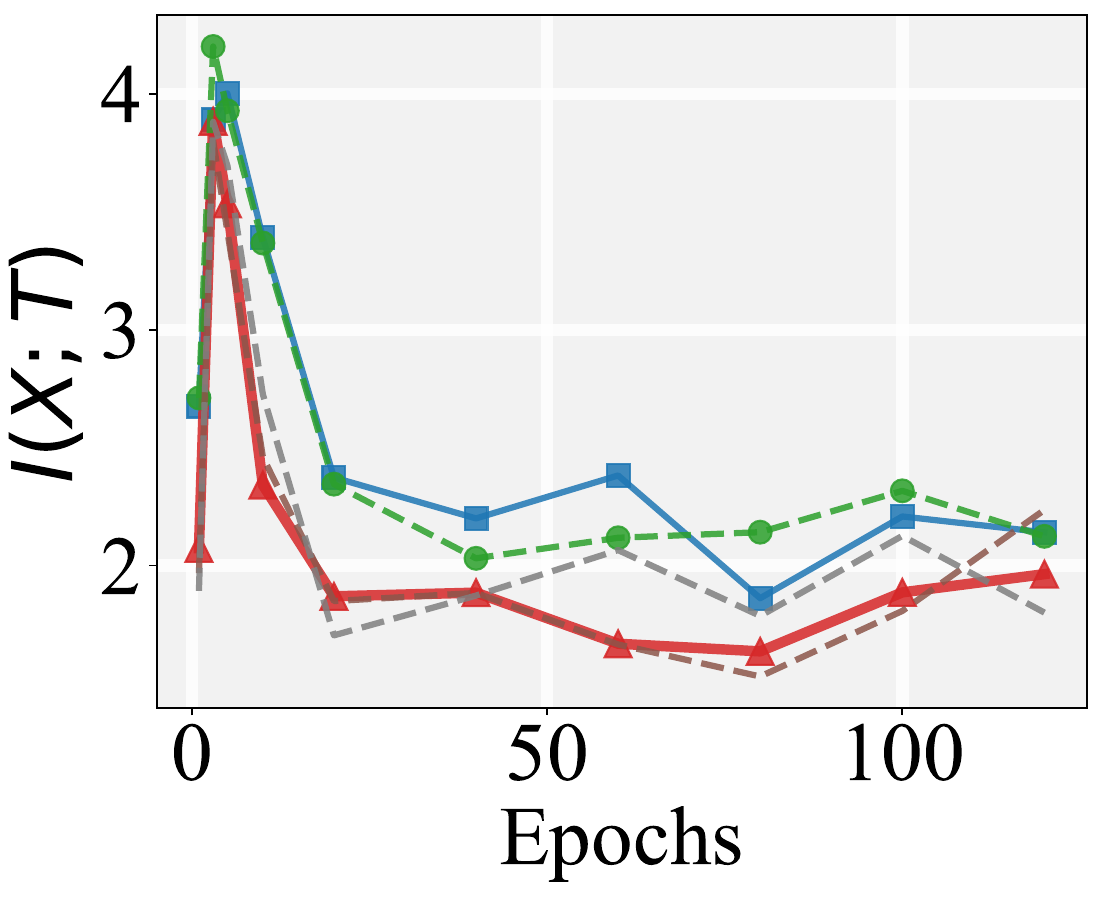} &
        \includegraphics[width=0.2\linewidth]{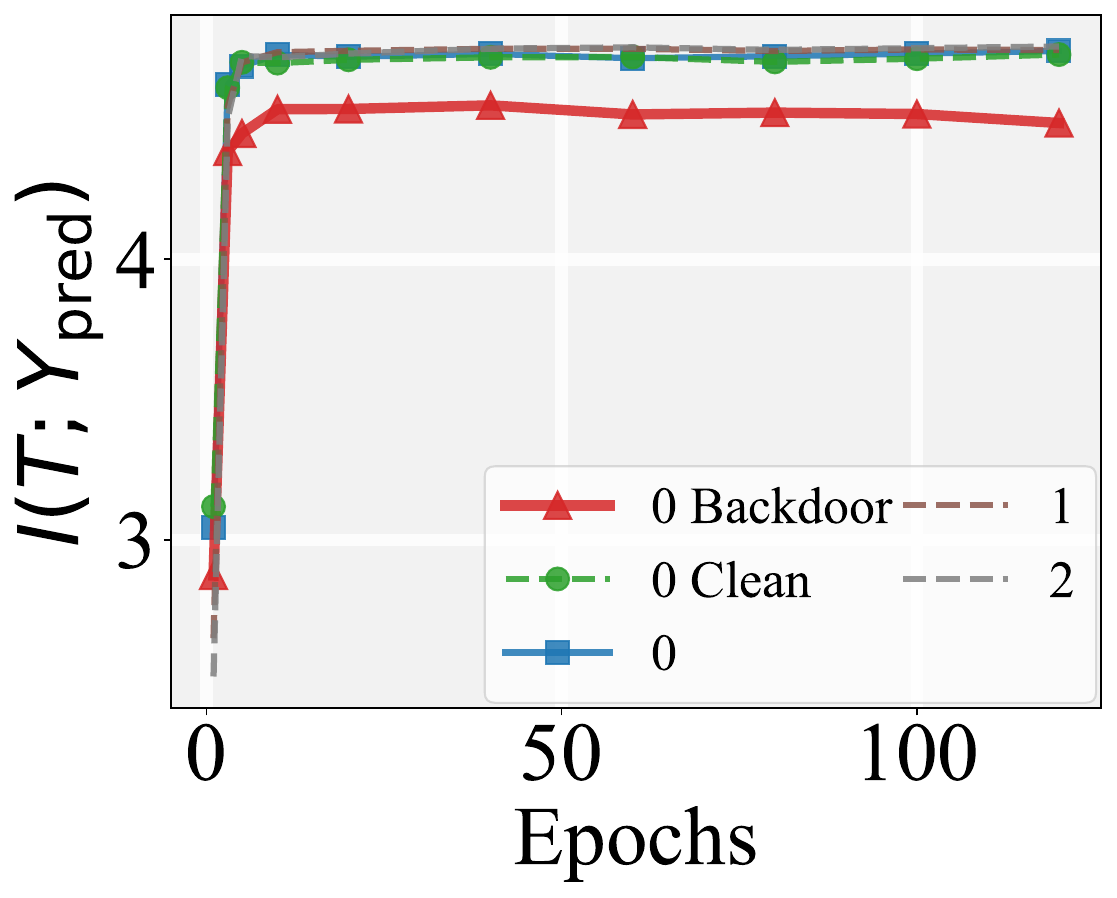} \\
    \end{tabular}
    \vspace{-2mm}
    \caption{\footnotesize{MI dynamics under a lower noise variance of $\gamma=0.2$. The rows represent different backdoor attacks, while the columns display the MI terms $I(X;T)$ and $I(T;Y_{pred})$.}}
    \label{fig:noise0.2_mi}
\end{figure*}

To validate robustness, we analyze MI dynamics under a reduced noise variance ($\gamma=0.2$) using the same attack configurations on CIFAR-10. As illustrated in Figure~\ref{fig:noise0.2_mi}, the characteristic MI patterns—such as the two-phase dynamic of BadNets and the rapid compression of Blend—remain highly consistent with the baseline ($\gamma=0.4$). The primary difference is uniformly higher absolute MI values, an expected consequence of the increased channel capacity associated with lower noise. This confirms that our findings capture fundamental dynamic behaviors and are not artifacts of specific hyperparameter choices.

\bibliographystyle{IEEEtran}
\bibliography{ref/ref, ref/ref_backdoor, ref/ref_IB_DP, ref/ref_backdoor_new}

@String(NeurIPS = {Adv. Neural Inform. Process. Syst.})

@String(ICASSP=	{ICASSP})

@String(NeurIPS = {NeurIPS})

@inproceedings{gu2017badnets,
	Author = {Gu, Tianyu and Dolan-Gavitt, Brendan and Garg, Siddharth},
	Booktitle = {Proc. of Machine Learning and Computer Security Workshop},
	Title = {Badnets: Identifying vulnerabilities in the machine learning model supply chain},
	Year = {2017}}

@inproceedings{wang2020practical,
  title={Practical detection of trojan neural networks: Data-limited and data-free cases},
  author={Wang, Ren and Zhang, Gaoyuan and Liu, Sijia and Chen, Pin-Yu and Xiong, Jinjun and Wang, Meng},
  booktitle={Computer Vision--ECCV 2020: 16th European Conference, Glasgow, UK, August 23--28, 2020, Proceedings, Part XXIII 16},
  pages={222--238},
  year={2020},
  organization={Springer}
}

@article{pal2023towards,
  title={Towards Understanding How Self-training Tolerates Data Backdoor Poisoning},
  author={Pal, Soumyadeep and Wang, Ren and Yao, Yuguang and Liu, Sijia},
  journal={arXiv preprint arXiv:2301.08751},
  year={2023}
}

@article{chen2017targeted,
	Author = {Chen, Xinyun and Liu, Chang and Li, Bo and Lu, Kimberly and Song, Dawn},
	Journal = {arXiv},
	Title = {Targeted backdoor attacks on deep learning systems using data poisoning},
	Year = {2017}}

@article{guo2019tabor,
	Author = {Guo, Wenbo and Wang, Lun and Xing, Xinyu and Du, Min and Song, Dawn},
	Journal = {arXiv preprint arXiv:1908.01763},
	Title = {Tabor: A highly accurate approach to inspecting and restoring trojan backdoors in ai systems},
	Year = {2019}}

@inproceedings{wang2019neural,
	Author = {Wang, Bolun and Yao, Yuanshun and Shan, Shawn and Li, Huiying and Viswanath, Bimal and Zheng, Haitao and Zhao, Ben Y},
	Booktitle = {IEEE S\&P},
	Title = {Neural cleanse: Identifying and mitigating backdoor attacks in neural networks},
	Year = {2019}}

@inproceedings{zhao2020clean,
  title={Clean-label backdoor attacks on video recognition models},
  author={Zhao, Shihao and Ma, Xingjun and Zheng, Xiang and Bailey, James and Chen, Jingjing and Jiang, Yu-Gang},
  booktitle={Proceedings of the IEEE/CVF conference on computer vision and pattern recognition},
  pages={14443--14452},
  year={2020}
}

@inproceedings{he2016deep,
  title={Deep residual learning for image recognition},
  author={He, Kaiming and Zhang, Xiangyu and Ren, Shaoqing and Sun, Jian},
  booktitle={Proceedings of the IEEE conference on computer vision and pattern recognition},
  pages={770--778},
  year={2016}
}

@article{krizhevsky2012imagenet,
  title={Imagenet classification with deep convolutional neural networks},
  author={Krizhevsky, Alex and Sutskever, Ilya and Hinton, Geoffrey E},
  journal={Advances in neural information processing systems (NeurIPS)},
  year={2012}
}

@article{krizhevsky2009learning,
  title={Learning multiple layers of features from tiny images},
  author={Krizhevsky, Alex and Hinton, Geoffrey},
  year={2009},
  journal={Technical Report TR-2009},
  publisher={Citeseer}
}

@article{wen2022deep,
  title={Deep learning-based perception systems for autonomous driving: A comprehensive survey},
  author={Wen, Li-Hua and Jo, Kang-Hyun},
  journal={Neurocomputing},
  year={2022},
  publisher={Elsevier}
}

@article{goldblum2022dataset,
  title={Dataset security for machine learning: Data poisoning, backdoor attacks, and defenses},
  author={Goldblum, Micah and Tsipras, Dimitris and Xie, Chulin and Chen, Xinyun and Schwarzschild, Avi and Song, Dawn and Madry, Aleksander and Li, Bo and Goldstein, Tom},
  journal={IEEE Transactions on Pattern Analysis and Machine Intelligence},
  year={2022},
  publisher={IEEE}
}

@article{tishby2000information,
  title={The information bottleneck method},
  author={Tishby, Naftali and Pereira, Fernando C and Bialek, William},
  journal={arXiv preprint physics/0004057},
  year={2000}
}

@inproceedings{tishby2015deep,
  title={Deep learning and the information bottleneck principle},
  author={Tishby, Naftali and Zaslavsky, Noga},
  booktitle={2015 ieee information theory workshop (itw)},
  pages={1--5},
  year={2015},
  organization={IEEE}
}

@article{amjad2019learning,
  title={Learning representations for neural network-based classification using the information bottleneck principle},
  author={Amjad, Rana Ali and Geiger, Bernhard C},
  journal={IEEE transactions on pattern analysis and machine intelligence},
  volume={42},
  number={9},
  pages={2225--2239},
  year={2019},
  publisher={IEEE}
}

@article{federici2020learning,
  title={Learning robust representations via multi-view information bottleneck},
  author={Federici, Marco and Dutta, Anjan and Forr{\'e}, Patrick and Kushman, Nate and Akata, Zeynep},
  journal={arXiv preprint arXiv:2002.07017},
  year={2020}
}

@article{wang2021revisiting,
  title={Revisiting hilbert-schmidt information bottleneck for adversarial robustness},
  author={Wang, Zifeng and Jian, Tong and Masoomi, Aria and Ioannidis, Stratis and Dy, Jennifer},
  journal={Advances in Neural Information Processing Systems},
  volume={34},
  pages={586--597},
  year={2021}
}

@inproceedings{simon2022towards,
  title={Towards a Robust Differentiable Architecture Search under Label Noise},
  author={Simon, Christian and Koniusz, Piotr and Petersson, Lars and Han, Yan and Harandi, Mehrtash},
  booktitle={Proceedings of the IEEE/CVF Winter Conference on Applications of Computer Vision},
  pages={3256--3266},
  year={2022}
}

@inproceedings{namekawa2021evolutionary,
  title={Evolutionary neural architecture search by mutual information analysis},
  author={Namekawa, Shizuma and Tezuka, Taro},
  booktitle={2021 IEEE Congress on Evolutionary Computation (CEC)},
  pages={966--972},
  year={2021},
  organization={IEEE}
}

@article{belghazi2018mine,
  title={Mine: mutual information neural estimation},
  author={Belghazi, Mohamed Ishmael and Baratin, Aristide and Rajeswar, Sai and Ozair, Sherjil and Bengio, Yoshua and Courville, Aaron and Hjelm, R Devon},
  journal={arXiv preprint arXiv:1801.04062},
  year={2018}
}

@article{shwartz2017opening,
  title={Opening the black box of deep neural networks via information},
  author={Shwartz-Ziv, Ravid and Tishby, Naftali},
  journal={arXiv preprint arXiv:1703.00810},
  year={2017}
}

@inproceedings{vera2018role,
  title={The role of the information bottleneck in representation learning},
  author={Vera, Matias and Piantanida, Pablo and Vega, Leonardo Rey},
  booktitle={2018 IEEE international symposium on information theory (ISIT)},
  pages={1580--1584},
  year={2018},
  organization={IEEE}
}

@article{goldfeld2018estimating,
  title={Estimating information flow in deep neural networks},
  author={Goldfeld, Ziv and Berg, Ewout van den and Greenewald, Kristjan and Melnyk, Igor and Nguyen, Nam and Kingsbury, Brian and Polyanskiy, Yury},
  journal={arXiv preprint arXiv:1810.05728},
  year={2018}
}

@inproceedings{liu2018trojaning,
  title={Trojaning attack on neural networks},
  author={Liu, Yingqi and Ma, Shiqing and Aafer, Yousra and Lee, Wen-Chuan and Zhai, Juan and Wang, Weihang and Zhang, Xiangyu},
  booktitle={25th Annual Network And Distributed System Security Symposium (NDSS 2018)},
  year={2018},
  organization={Internet Soc}
}

@inproceedings{borgnia2021strong,
  title={Strong data augmentation sanitizes poisoning and backdoor attacks without an accuracy tradeoff},
  author={Borgnia, Eitan and Cherepanova, Valeriia and Fowl, Liam and Ghiasi, Amin and Geiping, Jonas and Goldblum, Micah and Goldstein, Tom and Gupta, Arjun},
  booktitle={ICASSP 2021-2021 IEEE International Conference on Acoustics, Speech and Signal Processing (ICASSP)},
  pages={3855--3859},
  year={2021},
  organization={IEEE}
}

@article{huang2022backdoor,
  title={Backdoor defense via decoupling the training process},
  author={Huang, Kunzhe and Li, Yiming and Wu, Baoyuan and Qin, Zhan and Ren, Kui},
  journal={arXiv preprint arXiv:2202.03423},
  year={2022}
}

@article{xiang2022post,
  title={Post-training detection of backdoor attacks for two-class and multi-attack scenarios},
  author={Xiang, Zhen and Miller, David J and Kesidis, George},
  journal={arXiv preprint arXiv:2201.08474},
  year={2022}
}

@article{hu2021trigger,
  title={Trigger hunting with a topological prior for trojan detection},
  author={Hu, Xiaoling and Lin, Xiao and Cogswell, Michael and Yao, Yi and Jha, Susmit and Chen, Chao},
  journal={arXiv preprint arXiv:2110.08335},
  year={2021}
}

@inproceedings{kolouri2020universal,
  title={Universal litmus patterns: Revealing backdoor attacks in cnns},
  author={Kolouri, Soheil and Saha, Aniruddha and Pirsiavash, Hamed and Hoffmann, Heiko},
  booktitle={Proceedings of the IEEE/CVF Conference on Computer Vision and Pattern Recognition},
  pages={301--310},
  year={2020}
}

@inproceedings{shen2021backdoor,
  title={Backdoor scanning for deep neural networks through k-arm optimization},
  author={Shen, Guangyu and Liu, Yingqi and Tao, Guanhong and An, Shengwei and Xu, Qiuling and Cheng, Siyuan and Ma, Shiqing and Zhang, Xiangyu},
  booktitle={International Conference on Machine Learning},
  pages={9525--9536},
  year={2021},
  organization={PMLR}
}

@inproceedings{xu2021detecting,
  title={Detecting ai trojans using meta neural analysis},
  author={Xu, Xiaojun and Wang, Qi and Li, Huichen and Borisov, Nikita and Gunter, Carl A and Li, Bo},
  booktitle={2021 IEEE Symposium on Security and Privacy (SP)},
  pages={103--120},
  year={2021},
  organization={IEEE}
}

@inproceedings{liu2022complex,
  title={Complex backdoor detection by symmetric feature differencing},
  author={Liu, Yingqi and Shen, Guangyu and Tao, Guanhong and Wang, Zhenting and Ma, Shiqing and Zhang, Xiangyu},
  booktitle={Proceedings of the IEEE/CVF Conference on Computer Vision and Pattern Recognition},
  pages={15003--15013},
  year={2022}
}

@article{li2021anti,
  title={Anti-backdoor learning: Training clean models on poisoned data},
  author={Li, Yige and Lyu, Xixiang and Koren, Nodens and Lyu, Lingjuan and Li, Bo and Ma, Xingjun},
  journal={Advances in Neural Information Processing Systems},
  volume={34},
  pages={14900--14912},
  year={2021}
}

@article{turner2019label,
  title={Label-consistent backdoor attacks},
  author={Turner, Alexander and Tsipras, Dimitris and Madry, Aleksander},
  journal={arXiv preprint arXiv:1912.02771},
  year={2019}
}

@inproceedings{nguyen2021wanet,
  title={WaNet-Imperceptible Warping-based Backdoor Attack},
  author={Nguyen, Tuan Anh and Tran, Anh Tuan},
  booktitle={International Conference on Learning Representations},
  year={2021}
}

@inproceedings{bai2023physics,
  title={Physics-constrained backdoor attacks on power system fault localization},
  author={Bai, Jianing and Wang, Ren and Li, Zuyi},
  booktitle={2023 IEEE Power \& Energy Society General Meeting (PESGM)},
  pages={1--5},
  year={2023},
  organization={IEEE}
}

@inproceedings{mo2024robust,
  title={Robust backdoor detection for deep learning via topological evolution dynamics},
  author={Mo, Xiaoxing and Zhang, Yechao and Zhang, Leo Yu and Luo, Wei and Sun, Nan and Hu, Shengshan and Gao, Shang and Xiang, Yang},
  booktitle={2024 IEEE Symposium on Security and Privacy (SP)},
  pages={2048--2066},
  year={2024},
  organization={IEEE}
}

@inproceedings{netzer2011reading,
  title={Reading digits in natural images with unsupervised feature learning},
  author={Netzer, Yuval and Wang, Tao and Coates, Adam and Bissacco, Alessandro and Wu, Baolin and Ng, Andrew Y and others},
  booktitle={NIPS workshop on deep learning and unsupervised feature learning},
  volume={2011},
  number={2},
  pages={4},
  year={2011},
  organization={Granada}
}

@article{li2022backdoor,
  title={Backdoor learning: A survey},
  author={Li, Yiming and Jiang, Yong and Li, Zhifeng and Xia, Shu-Tao},
  journal={IEEE Transactions on Neural Networks and Learning Systems},
  volume={35},
  number={1},
  pages={5--22},
  year={2022},
  publisher={IEEE}
}

@article{torfi2020natural,
  title={Natural language processing advancements by deep learning: A survey},
  author={Torfi, Amirsina and Shirvani, Rouzbeh A and Keneshloo, Yaser and Tavaf, Nader and Fox, Edward A},
  journal={arXiv preprint arXiv:2003.01200},
  year={2020}
}

@article{downer2024securing,
  title={Securing GNNs: Explanation-Based Identification of Backdoored Training Graphs},
  author={Downer, Jane and Wang, Ren and Wang, Binghui},
  journal={arXiv preprint arXiv:2403.18136},
  year={2024}
}

@misc{goldfeldEstimatingInformationFlow2019,
  title = {Estimating {{Information Flow}} in {{Deep Neural Networks}}},
  author = {Goldfeld, Ziv and van den Berg, Ewout and Greenewald, Kristjan and Melnyk, Igor and Nguyen, Nam and Kingsbury, Brian and Polyanskiy, Yury},
  year = {2019},
  month = may,
  number = {arXiv:1810.05728},
  eprint = {1810.05728},
  publisher = {arXiv},
  urldate = {2024-10-27},
  archiveprefix = {arXiv}
}

@misc{oordRepresentationLearningContrastive2019,
  title = {Representation {{Learning}} with {{Contrastive Predictive Coding}}},
  author = {van den Oord, Aaron and Li, Yazhe and Vinyals, Oriol},
  year = {2019},
  month = jan,
  number = {arXiv:1807.03748},
  eprint = {1807.03748},
  primaryclass = {cs, stat},
  publisher = {arXiv},
  urldate = {2024-08-19},
  archiveprefix = {arXiv},
  langid = {english}
}

@misc{palBackdoorSecretsUnveiled2024,
  title = {Backdoor {{Secrets Unveiled}}: {{Identifying Backdoor Data}} with {{Optimized Scaled Prediction Consistency}}},
  shorttitle = {Backdoor {{Secrets Unveiled}}},
  author = {Pal, Soumyadeep and Yao, Yuguang and Wang, Ren and Shen, Bingquan and Liu, Sijia},
  year = {2024},
  month = mar,
  number = {arXiv:2403.10717},
  eprint = {2403.10717},
  primaryclass = {cs},
  publisher = {arXiv},
  urldate = {2024-08-13},
  archiveprefix = {arXiv},
  langid = {english}
}

@article{saxe2019information,
  title={On the information bottleneck theory of deep learning},
  author={Saxe, Andrew M and Bansal, Yamini and Dapello, Joel and Advani, Madhu and Kolchinsky, Artemy and Tracey, Brendan D and Cox, David D},
  journal={Journal of Statistical Mechanics: Theory and Experiment},
  volume={2019},
  number={12},
  pages={124020},
  year={2019},
  publisher={IOP Publishing}
}

@article{simonyan2014very,
  title={Very deep convolutional networks for large-scale image recognition},
  author={Simonyan, Karen},
  journal={arXiv preprint arXiv:1409.1556},
  year={2014}
}

@inproceedings{wang2022invisible,
  title={An invisible black-box backdoor attack through frequency domain},
  author={Wang, Tong and Yao, Yuan and Xu, Feng and An, Shengwei and Tong, Hanghang and Wang, Ting},
  booktitle={European Conference on Computer Vision},
  pages={396--413},
  year={2022},
  organization={Springer}
}

@inproceedings{qi2023revisiting,
  title={Revisiting the assumption of latent separability for backdoor defenses},
  author={Qi, Xiangyu and Xie, Tinghao and Li, Yiming and Mahloujifar, Saeed and Mittal, Prateek},
  booktitle={The eleventh international conference on learning representations},
  year={2023}
}

@inproceedings{liu2020reflection,
  title={Reflection backdoor: A natural backdoor attack on deep neural networks},
  author={Liu, Yunfei and Ma, Xingjun and Bailey, James and Lu, Feng},
  booktitle={European Conference on Computer Vision},
  pages={182--199},
  year={2020},
  organization={Springer}
}

@article{ghazanfari2023r,
  title={R-LPIPS: An adversarially robust perceptual similarity metric},
  author={Ghazanfari, Sara and Garg, Siddharth and Krishnamurthy, Prashanth and Khorrami, Farshad and Araujo, Alexandre},
  journal={arXiv preprint arXiv:2307.15157},
  year={2023}
}

@article{wang2004image,
  title={Image quality assessment: from error visibility to structural similarity},
  author={Wang, Zhou and Bovik, Alan C and Sheikh, Hamid R and Simoncelli, Eero P},
  journal={IEEE transactions on image processing},
  volume={13},
  number={4},
  pages={600--612},
  year={2004},
  publisher={IEEE}
}
\newpage


 




\vfill

\end{document}